\documentclass[preprint,3p,authoryear,11pt]{elsarticle}

\usepackage{times}
\usepackage{epsfig}
\usepackage{natbib}
\usepackage{graphicx,color}
\usepackage{amsmath}
\usepackage{amssymb}
\usepackage{booktabs}
\usepackage{bm}

\newcommand{\real}{\ensuremath{\mathbb{R}}}

\newcommand{\ltwo}{\ensuremath{\mathbb{L}^2}}

\newtheorem{defn}{Definition}

\newtheorem{algorithm}{Algorithm}

\newcommand{\abs}[1]{\left\vert#1\right\vert}

\newcommand \bbE{\mathbb{E}}

\begin{document}

\begin{frontmatter}

\title{Bayesian Clustering of Shapes of Curves}



\author[fsu]{Zhengwu Zhang \corref{corl}}
\ead{zhengwu@stat.fsu.edu}

\author[fsu]{Debdeep Pati}
\ead{debdeep@stat.fsu.edu}

\author[fsu]{Anuj Srivastava}
\ead{anuj@stat.fsu.edu}

\cortext[corl]{Corresponding author. Tel. : +1 850-644-7412}
\address[fsu]{Department of Statistics, Florida State University, FL, 32306}

\begin{abstract}
Unsupervised clustering of curves according to their shapes is an important problem 
with broad scientific applications. The existing model-based clustering techniques either rely on simple probability models 
(e.g., Gaussian) that are not generally valid for shape analysis or assume the number of clusters.  
 We develop an efficient  Bayesian method to cluster curve data using
an elastic shape metric that is based on joint registration
and comparison of shapes of curves. The elastic-inner product matrix obtained from the data is modeled using a Wishart distribution whose parameters are assigned carefully chosen prior distributions to allow for automatic inference on the number of clusters. 
Posterior is sampled through an efficient Markov chain Monte Carlo procedure based on the Chinese restaurant process to infer (1) the posterior distribution on the number of clusters, 
and (2) clustering configuration of shapes. This method is demonstrated on a variety of synthetic data and real data examples on 
protein structure analysis, cell shape analysis in microscopy images, and clustering of shaped from MPEG7 database.
\end{abstract}



\end{frontmatter}

\section{Introduction}
The automated clustering of objects is an important area of research  in unsupervised classification of large object databases. The general goal here is to choose
groups (clusters) of objects so as to maximize homogeneity within clusters and minimize homogeneity across clusters.   
The clustering problem has been  addressed by researchers in many disciplines. A few well-known methods are  metric based e.g. K-means \citep{macqueen1967some}, hierarchical clustering \citep{Ward63}, clustering based on principal components,  spectral clustering \citep{ng2002spectral} and so on \citep{Jain1988,Ozawa1985279}. 
Traditional clustering methods are complemented by methods based on a probability model where one assumes a data generating distribution (e.g., Gaussian) and infers clustering configurations that maximize certain objective function  \citep{banfield1993model,fraley1998many,fraley2002model,fraley2006mclust, MacCullagh2008}. 
A model-based clustering can be useful in addressing challenges posed by traditional clustering methods.  This is because a probability model allows  the number of clusters to be treated as a parameter in the model, and can be embedded in a Bayesian framework providing quantification of uncertainty  in the number of clusters and clustering configurations.
 
 A popular probability model  is obtained by considering that the population of interest consists of $K$ different sub-populations and the density of the observation $y$ from the 
 $k^{th}$ sub-population is $f_k$. Given observations $y_1, \ldots, y_n$, we introduce indicator random variables $(c_1, \ldots, c_n)$ such that $c_i = k$ if $y_i$ comes from the $k$th sub-population.  The maximum likelihood inference is based on finding the value of $(c, f_1, \ldots, f_k)$ that maximizes the likelihood  $\prod_{i=1}^n f_{c_i}(y_i)$.  Typically $K$ is assumed to be known or a suitable upper bound is assumed for convenience.    When $y_i  \in \mathbb{R}^p$, $f_k$ is commonly parametrized by a multivariate Gaussian density with mean vector $\mu_k$ and covariance matrix $\Sigma_k$.   An alternative is to use a nonparametric Bayesian approach which has an appealing advantage of allowing $K$ to be unknown and inferring it from the data.  An advantage of such an approach is that it not only provides an estimate of the number of clusters, but also the entire posterior distribution. 
  
The vast majority of the literature on model-based clustering is almost exclusively focused on Euclidean data.  This is primarily due to the easy availability of 
parametric distributions on the Euclidean space as well as computational tractability of estimating the cluster centers. For clustering functional data, 
e.g. shapes of curves,  one encounters several challenges.
Unlike Euclidean data, where the notions of cluster centers and cluster variance are standard, these quantities and 
the resulting quantification of homogeneity within clusters are not obvious for shape spaces. 
Moreover,  it is important to use representations and metrics for clustering objects 
that are invariant to shape-preserving transformations (rigid motions, scaling, and re-parametrization).  
For example,  \cite{Kurtek2012}  takes a model-based approach for clustering of curves using an elastic metric that has proper invariances.
However, under the  chosen representations and metrics, even simple summary statistics of the observed data are difficult to compute.  
Other existing shapes clustering methods \citep{Belongie2002,Meizhu2012} either extract finite-dimensional features to represent the shapes or project the high-dimensional shape space to a low-dimensional space \citep{Yankov2006,Auder2012}, and then apply clustering methods for Euclidean data; these approaches are not necessarily invariant to shape preserving transformations. 
 Also, several methods \citep{Srivastava05shapte,Gaffney05}  have been proposed  to 
cluster non-Euclidean data based on a distance-based notion of dispersion, thus, avoiding the computation of shape means (e.g. Karcher means), 
but they all assume a given number of clusters. 




In this paper we develop a model-based clustering method for curve data that does not require the knowledge of cluster number $K$ apriori.  
This approach is based on modeling a summary statistic that encodes the clustering information, namely the inner product matrix.  
The salient points of this approach are: 
(1)  The comparison of curves is based on the inner product matrix under 
elastic shape analysis, so that the analysis is invariant to all desired shape-preserving transformations. 
(2) The inner product matrix is modeled using a Wishart distribution with prior on the clustering configurations induced by the Chinese restaurant process \citep{translationinavariantwd}.   A model directly on the inner-product matrix has an appealing advantage of reducing computational cost substantially by avoiding computation of the Karcher means.  
(3) We formulate and sample from a posterior on the number of clusters, and use the mode of this distribution for final clustering. 
We illustrate our ideas through several synthetic and real data examples. The results show that our model on the inner product matrix leads to a more accurate estimate of the number of clusters as well as the clustering configurations compared to a Bayesian nonparametric model directly on the data, even in the Euclidean case.  

This paper is organized as follows. We start by introducing two case studies in Section \ref{sec:data}.   The mathematical details of the metric used for computing the inner product and and the model specifications are presented in Section \ref{sec:model}.  In Section \ref{sec:sim}, we illustrate our methodology on several synthetic data examples and the case studies on clustering cell shapes and protein structures.  Section \ref{sec:disc} closes the paper with some conclusions. 

\section{Case studies} \label{sec:data}
We propose to undertake two specific case studies involving clustering of curve data. 
\subsection{Clustering of protein sequences}
\begin{figure}
\begin{center}
\begin{tabular}{cc}
\includegraphics[height=1.8in]{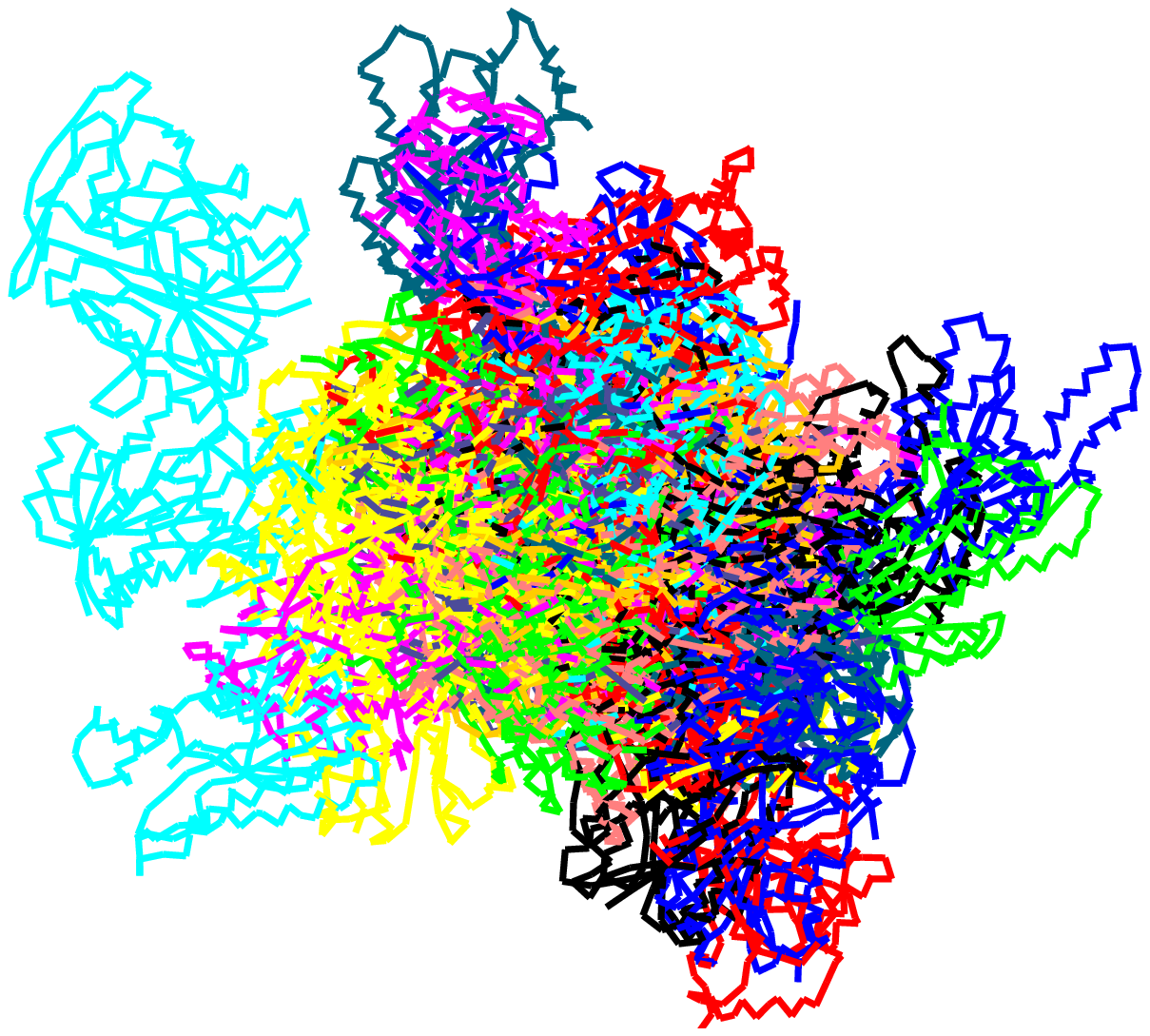}&
\includegraphics[height=1.8in]{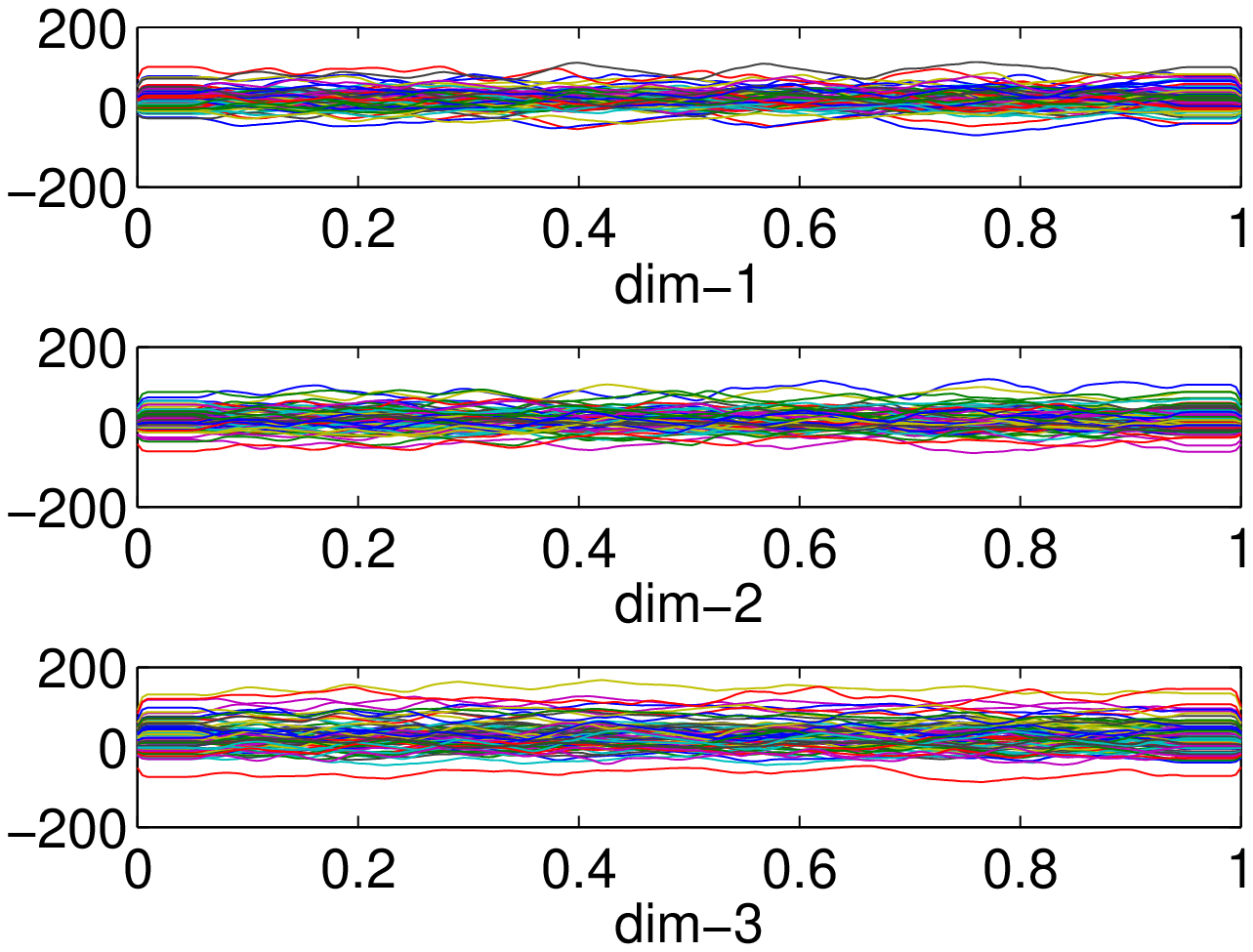}\\
(a) & (b)\\
\end{tabular}
\caption{Protein sequences. (a) raw protein structure data in $\real^3$. (b) $3$-dimensional components of the protein sequences, where the $x$-axis indicates the length of each sequence.}
\label{fig:proteinexample}
\end{center}
\end{figure}

Protein structure analysis is an outstanding scientific problem in structural biology. A large number of new proteins  are regularly discovered 
and scientists are interested in learning about their functions in larger biological systems. Since protein functions are closely 
related to their folding patterns and structures in native states, the task of structural analysis of proteins becomes important. 
In terms of evolutionary origins, proteins with similar structures are considered to have common evolutionary origin.  The Structural Classification of Proteins (SCOP) database \citep{Murzin1995536} provides a manual classification of protein structural domains based on similarities of their structures and amino acid sequences.  Refer to Fig. \ref{fig:proteinexample} for a snapshot of the proteins in $\real^3$ and the $3$-coordinates  of the protein sequences. Clustering protein sequences is extremely important  to trace the evolutionary relationship between proteins and  for detecting conserved structural motifs.  
 In this article, we focus on an automated clustering of protein sequences based on their global structures.

\subsection{Clustering of cell shapes}
\begin{figure}
\begin{center}
\begin{tabular}{cc}
\includegraphics[height=1.8in]{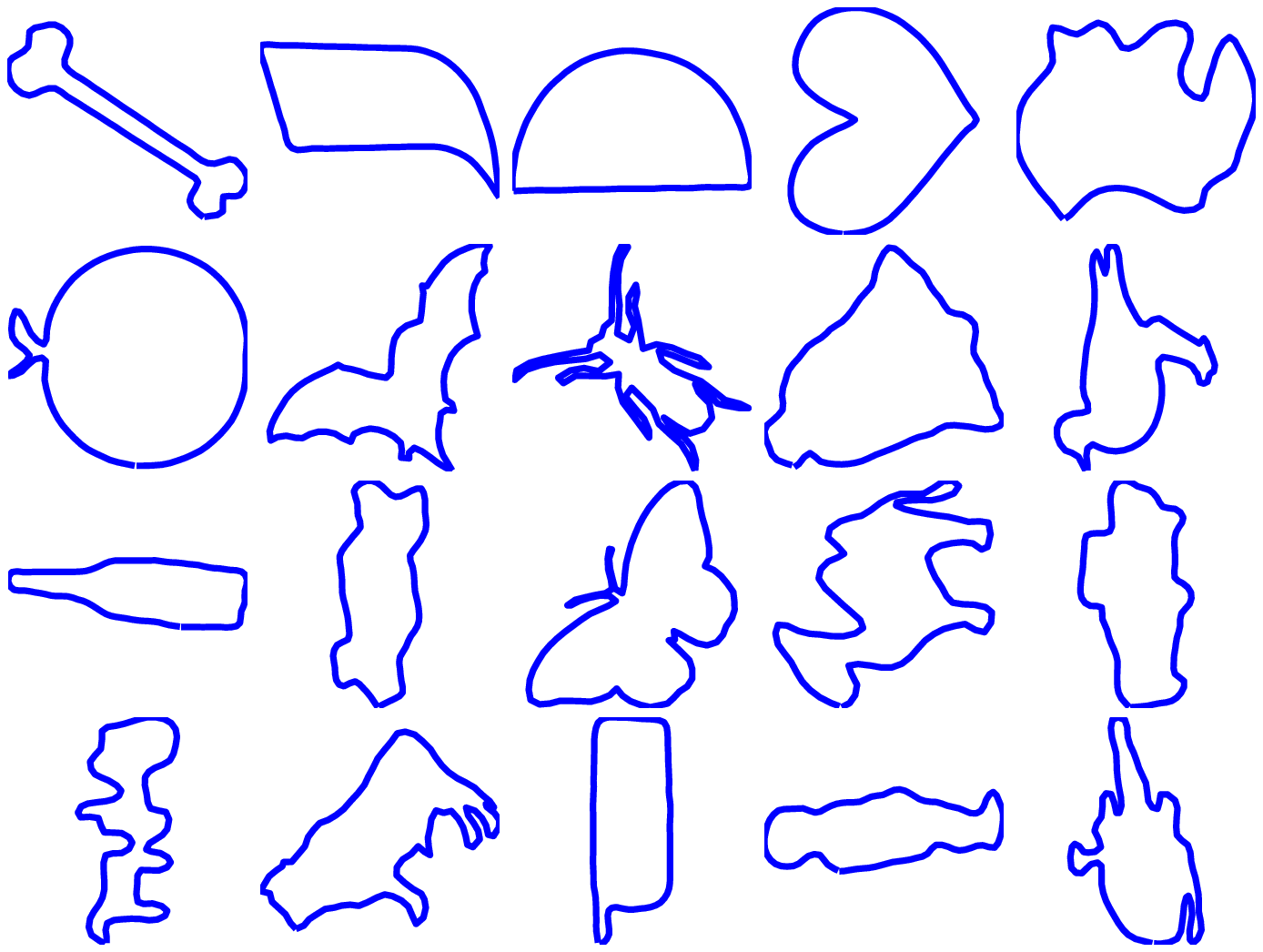}&
\includegraphics[height=1.8in]{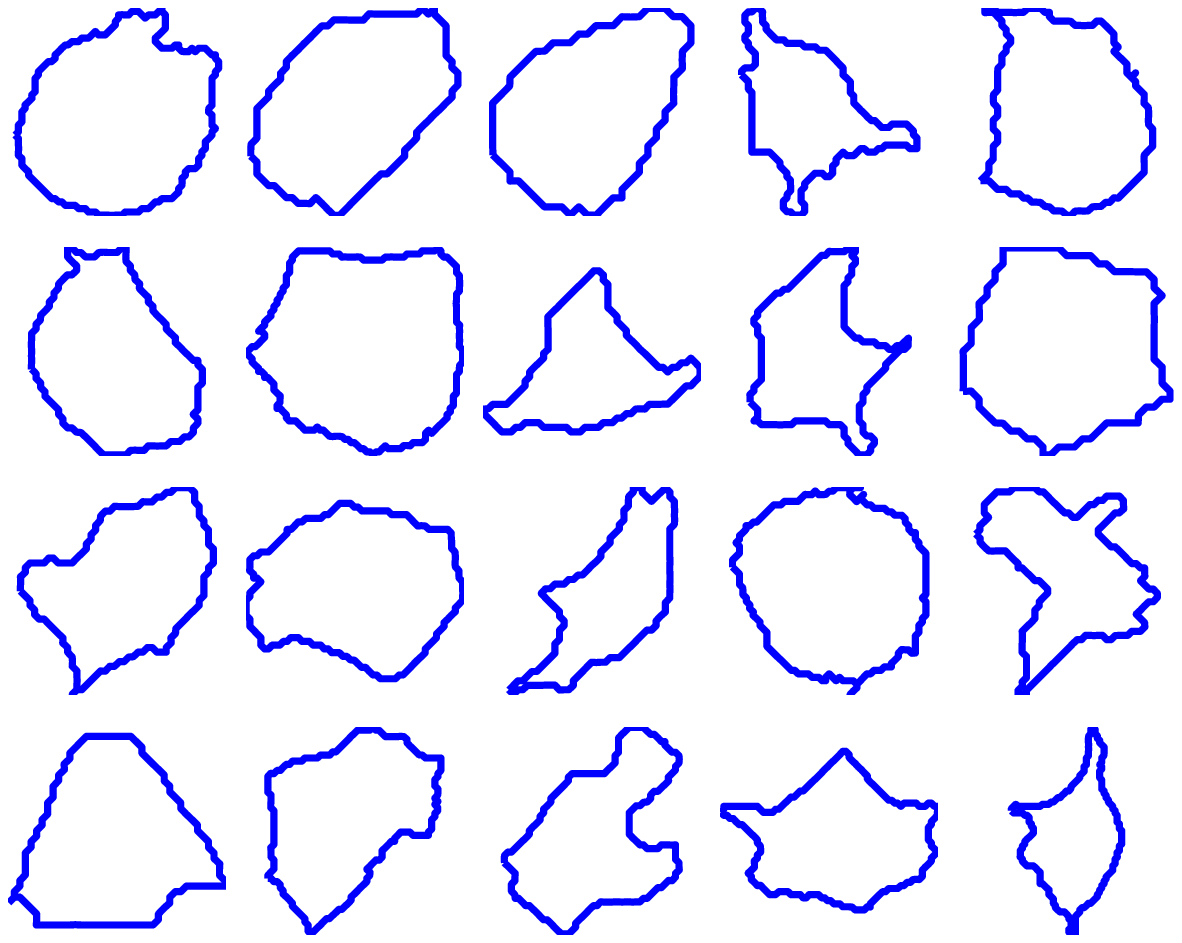}\\
(a) & (b)
\end{tabular}
\caption{Example shapes. (a) example shapes in the MPEG-7 dataset; (b) example cell shapes.}
\label{fig:shape_mpeg7}
\end{center}
\end{figure}

The problem of studying shapes of cellular structures using microscopic image data is very important medical diagnosis \citep{CYTO:CYTO20506} and 
genetic engineering \citep{thomas-etal-PNAS:2012}. This research involves extracting cell contours from images using segmentation techniques 
\citep{hagwood2012evaluation}
and then studying shapes of these extracted contours for medical diagnosis.   We will focus on the problem of clustering of cells according to their shapes; these 
clusters can be further used for statistical modeling and hypothesis testing although these steps are not pursued in the current paper. 
The specific database used here was obtained by segmenting the 2D microscopy images, as described in
\cite{hagwood2012evaluation,6587104}. Fig. \ref{fig:shape_mpeg7} (b) shows some examples of the cell contours used in this paper. 
In this article, we consider two types of cell shapes: DLEX-p46 cell shapes and NIH-3T3 cell shapes. Visually, DLEX-P46 cells are round, denoting normal 
cell shapes whereas, NIH-3T3 cells have an elongated, spindly appearance, denoting progression of some pathological conditions.



\section{Methodology}\label{sec:model}

Previous methods of clustering non-Euclidean objects can be broadly categorized into two parts: (1) clustering based on representation of the data in 
an infinite-dimensional quotient space under a chosen 
Riemannian metric and (2) clustering based on suitable summary statistic of the data e.g. distance matrices.  Any representation in the infinite-dimensional space involves the calculation of the mean and the covariance matrix  \citep{Kurtek2012,Tucker:2013} which is computationally expensive.   To avoid calculating the mean and the covariance matrix,   \cite{Srivastava05shapte} developed a method for clustering functional data based on  pairwise distance matrix and resorted to  stochastic simulated annealing for fast implementation.  Although the method is quite efficient, one requires the knowledge of the number of clusters 
apriori. 


In this paper,  we develop a model based on the Wishart distribution for the inner product matrices to cluster shapes of curves.  Since we model a summary statistic of the data as in \citep{BayesianPt,translationinavariantwd} instead of the infinite dimensional data points,  our method is computationally efficient.  However, unlike \citep{BayesianPt,translationinavariantwd} which consider a standard $\ltwo$ metric to calculate the distance matrices,  the inner product matrix is calculated  using a specific representation of curves called {\it square-root velocity function} (SRVF) \citep{AnujShape}. This along with some registration techniques make the inner product invariant to the shape preserving transformations, thus eliminating the drawback of \citep{BayesianPt,translationinavariantwd}.   Moreover, a Bayesian nonparametric approach allows us to do automatic inference on the number of clusters. 

Below, we describe the mathematical framework for computing the inner product matrix. 

\subsection{Inner product matrix using elastic shape analysis}

We adapt the elastic shape analysis introduced in \cite{AnujShape} to calculate the inner product matrix in the  
{\it square-root velocity function} (SRVF) space for the non-Euclidean functional data. 
Let $\beta: D \to \real^p$ be a parameterized curve in $\real^p$ with domain $D$. We restrict our attention to those $\beta$ which are absolutely continuous on $D$. Usually $D = [0,1]$ for open curves and  $D$ = $\mathbb{S}^1$ for closed curves. Define 
$\cal F =$ $ \{ \beta : D \rightarrow \real^p:   \beta \text{ is absolutely continuous on $D$} \}$ and a continuous mapping: $Q:\real^p \rightarrow \real^p$ as
$$
Q(x) \equiv  
 \begin{cases}
x/\sqrt{|x|} &\quad \text{if } |x| \neq 0\\
0 &\quad \text{otherwise.}
\end{cases} $$
Here, $|\cdot|$ is the Euclidean $2$-norm in $\real^p$. 
For the purpose of studying the shape of a curve $\beta$, we will represent it as: $q:D \rightarrow \real^p$, where $q(t) \equiv Q(\dot{\beta}(t))$. 
The function $q:D \to \real^p$ is called {\it square-root velocity function} (SRVF).  
It can be shown that for any $\beta \in \cal F$, the resulting SRVF is square integrable. Hence, we will define $\ltwo(D,\real^p)$ to be the set of all SRVFs. For every $q \in \ltwo(D,\real^p)$ there exists a curve $\beta$ (unique up to a constant, or a translation) such that the given $q$ is the SRVF of that $\beta$.

There are several motivations for using SRVF for functional data analysis.
First, an elastic metric becomes the standard $\ltwo$ metric under the SRVF representation \citep{AnujShape}. This elastic metric  is invariant to the re-parameterization of curves and provides nice physical interpretations. Although the original elastic
metric has a complicated expression, the SRVF transforms it into the $\ltwo$ metric, thus providing a substantial simplification in terms of computing the metric.

By representing a parameterized curve $\beta$ by its SRVF $q$,  we have taken care  of the translation variability, but the scaling, rotation and the re-parameterization variabilities still remain. In some applications like clustering of protein sequences, it is not advisable to remove the scaling variabilities as the length can be a predictor of its biological functions.  On the contrary, in applications like clustering images with the camera placed at variable distances,  it is necessary to remove the scales by  rescaling all curves to be of unit length, i.e., $\int_D|\dot{\beta}(t)| dt = \int_D|q(t)|^2 dt=1$. The set of all  
SRVFs representing unit-length curves is a unit hypersphere in the Hilbert manifold $\ltwo(D,\real^p)$.  
We will use $\mathcal {C}^o$ to denote this hypersphere, i.e.,  $\mathcal {C}^o = \{ q \in \ltwo(D,\real^p)|\int_D |q(t)|^2 dt = 1 \} $.
A rigid rotation in $\real^p$ is represented as an element of $SO(p)$, the special orthogonal group of $p \times p$ matrices. The rotation action is defined to be $SO(p) \times \mathcal {C}^o \rightarrow \mathcal {C}^o$ as follows. If a  curve is rotated by a rotation matrix $O \in SO(p)$, then its SRVF is also rotated by the same matrix, i.e. the SRVF of $O\beta(t)$ is $Oq(t)$, where $q$ is the SRVF of $\beta$. A re-parameterization function is an element of $\Gamma$, the set of all orientation-preserving diffeomorphisms of $D$. For any $\beta \in \cal F$ and $\gamma \in \Gamma$, the composition $\beta \circ \gamma$ denotes the re-parameterization of $\beta$ by $\gamma$.
The SRVF of $\beta \circ \gamma$ is given by: $\tilde{q}(t) = q(\gamma(t))\sqrt{\dot{\gamma}(t)}$. We will use $(q,\gamma)$ to denote $q(\gamma(t))\sqrt{\dot{\gamma}}$ in the following.

It is easy to show that the actions of $SO(p)$ and $\Gamma$ on ${C}^o$ commute each other, thus we can form a join action of the product group $SO(p) \times \Gamma$ on $\mathcal {C}^o$ according to $((O,\gamma),q) = O(q\circ\gamma)\sqrt{\dot{\gamma}}$. The action of the product group $\Gamma \times SO(p)$ is by isometries under the chosen Riemannian metric. The orbit of an SRVF $q \in \mathcal {C}^o$ is the set of SRVFs associated with all the reparameterizations and rotations of a given curve and is given by: $[q] = \text{closure}\{(q,(O,\gamma))|(O,\gamma) \in SO(p) \times \Gamma\}$. 
The specification of orbits is important because each orbit uniquely represents a shape and, therefore, analyzing the shapes is equivalent to the analysis of orbits. 
The set of all such orbits is denoted by $\cal S$ and termed the {\it shape space}. 
${\cal S}$ is actually a quotient space given by  $\mathcal{S} = \mathcal{C}^o/(SO(p) \times \Gamma )$. 
Now we can define an inner product on the space $\mathcal{S}$ which is invariant to translation, scaling, rotation and reparameterization of curves.   
\begin{defn}
\label{def:innerpm} 
(inner product on shape space of curves).  For given curves $\beta_1,\beta_2 \in \cal F$ and the corresponding SRVFs, $q_1,q_2$, we define the inner product, $s_{\beta_1,\beta_2}$ or $\left< [q_1],[q_2] \right >_{\mathbb{H}}$, to be:
$$s_{\beta_1,\beta_2} = \sup_{\gamma \in \Gamma, O \in SO(d)} \left <q_1,(q_2,(O,\gamma)) \right>.$$
\end{defn}
Note that this inner product is well-defined because the action on $SO(p) \times \Gamma$  is by isometries.  
 
{\bf Optimization over $SO(p)$ and $\Gamma$:} The maximization over $SO(p)$ and $\Gamma$ can be performed iteratively as in \citep{AnujShape}. In our case, we use Dynamic Programming algorithm  to solve for an optimal $\gamma$ first. Then we fix $\gamma$, and search for the optimal rotation $O$ in $SO(p)$ using a rotational Procrustes algorithm \citep{Kurtek2012}. 

\subsection{Likelihood specification for the inner product matrix} \label{ssec:DMW}
Let $S_{+ n}(\mathbb{R})$ denote the set of all $n \times n$ symmetric non-negative definite matrices over $\mathbb{R}$.  Depending on whether we rescale the curves to have unit length or not,  we define two classes of inner product matrices:  
i) $U_{+ n}(\mathbb{R})  = \{A \in  S_{+n}(\mathbb{R}): a_{ii} = 1, \abs{a_{ij}} \leq 1, 1\leq i \neq j \leq n \}$ and ii) 
$S_{+ n}(\mathbb{R})$.  In this article,  we do not make a distinction between these two cases and specify our model for the larger subspace $S_{+ n}(\mathbb{R})$ irrespective of whether we rescale the curves or not. As illustrated using the  experimental results in Section \ref{sec:sim},  having a probability model on a slightly larger space does not pose any practical issues when we actually rescale the curves. 

For a scaled  inner product matrix $S \in S_{+n}(\mathbb{R})$, let $S \sim W_n(\Sigma,d)$, the Wishart distribution with degrees of freedom $d$ and parameter $\Sigma = \bbE(S)$ of rank $n$ ($d>n$).  To allow rank-deficient $S$, a {\it generalized Wishart distribution} with degrees of freedom $d$ ($d < n$)  can be defined as
\begin{eqnarray} \label{eq:wishartgen}
p(S \mid \Sigma,d) \propto \abs{S}^{(d-n-1)/2} \abs{\Sigma^{-1}}^{d/2}\exp\bigg\{-\frac{d}{2}\mbox{tr}(\Sigma^{-1}S)\bigg\}\ ,
\end{eqnarray}
where $\abs{\cdot}$ implies the product of non-zero eigenvalues and $\text{tr}(\cdot)$ is the sum of the diagonal elements.
For an observation of $S$, the log-likelihood function is 
\begin{eqnarray} \label{eq:wishartsigma}
  l(\Sigma;S,d) \propto -\frac{d}{2}\log(\abs{\Sigma}) -\frac{d}{2}\mbox{tr}(\Sigma^{-1}S)\  
\end{eqnarray}
for $\Sigma \in S_{+n}(\mathbb{R})$.   One can easily identify  this as an exponential family distribution with canonical parameter $W = \Sigma^{-1}$, and the deviance is minimized at $\Sigma = S$ \citep{mcculagh2009}. Therefore, $\Sigma$ encodes the similarity between the observed shapes measured by the inner product matrix $S$. For instance,  $\Sigma_{jk}$ encodes  the similarity between $y_i$ and $y_j$ as measured by the inner product $\langle [q_i], [q_j] \rangle_{\mathbb{H}}$, where $q_i$ and $q_j$ are the SRVFs of $y_i$ and $y_j$, respectively. 

Clustering is equivalent to finding an optimal partition of the data.  We use $P=\{P_1, P_2, \ldots, P_K\} \in \mathcal{P}$ to denote a partition of set $\{1,2,...,n\}$ into $K$ classes, where $\mathcal{P}$ denotes the set of all partitions of $\{1, 2, \ldots, n\}$. A partition $P$ can also be represented by {\it membership indicators} $\{c_i, i=1, \ldots, n\}$, where $c_i = j$ if $i \in P_j, j=1,\ldots,K$, or a {\it membership matrix} $B \in \real^{n \times n}$, defined as
$
 B_{ij} = 
 \begin{cases}
1 & \text{if } c_i = c_j \\ 
0 &\text{otherwise\ .} 
\end{cases}
$
If we assume: (1) observed shapes $\{y_i \in \mathcal{F}, i=1, \ldots, n\}$ come from several sub-populations, and (2) observed shapes from the same population are placed next to each other; one would expect to observe a block pattern in the inner product matrix $S$ because the observations from the same cluster will have similar inner product.  Fig. \ref{fig:pmatrix} on the left panel shows one example of such inner product matrix, which is calculated from simulated Euclidean data with three clusters. One can observe three large-value-blocks along the diagonal. 

To perform Bayesian inference on the clustering configurations, we  define the  following prior on $\Sigma$ that enables clustering of the observations.  
Motivated by  \citep{MacCullagh2008,BayesianPt,translationinavariantwd},  consider the following decomposition of $\Sigma$.
Let 
\begin{eqnarray}\label{eq:decomp} 
\Sigma = \alpha I + \beta B \ ,
\end{eqnarray}
where $\alpha,\beta \in \real$, $I$ is the identity matrix and $B\in \real^{n\times n}$ is the membership matrix.  Equation \eqref{eq:decomp} decomposes the scalar matrix $\Sigma$ into a sparse matrix $\alpha I$ and a low-rank matrix $\beta B$, where $B$ encodes the clustering information. For convenience of introducing a conjugate prior for $\alpha$ \citep{translationinavariantwd} , we re-parameterize this model into $\Sigma = \alpha (I + \theta B)$, where $\theta = \beta/\alpha$. Intuitively, the parameter $\theta$ controls the strength of similarity between two observations measured by their inner product - a large $\theta$ indicates a strong association, and vise versa. Refer to Fig. \ref{fig:pmatrix}  for an illustration of the membership matrix $B$ and the corresponding $\Sigma$ matrix.

Our primary goal is to develop a Bayesian approach to infer the posterior distribution on the membership matrix $B$.  To that end,  
we denote the likelihood  by $p(S|\Sigma, d)$, and we place priors on $\Sigma$ and $d$, the unknown parameters in the likelihood. The prior on $\Sigma$ is induced by first letting $\Sigma = \alpha(I + \theta B)$ and then  placing priors on $\alpha$, $\theta$ and $B$.  These constitute our Bayesian model for $S$, the inner-product matrix.   Below, we discuss the specification of prior distributions for $\alpha,\theta, B$ and $d$. 

\subsection{Priors and hyperpriors}
 A popular method of inducing a prior distribution on the space of partitions $\mathcal{P}$ is the Chinese restaurant process (CRP) \citep{pitman2006combinatorial} induced by a Dirichlet process \citep{ferguson1973bayesian,ferguson1974prior}.  Since a prior on $\{c_i,i=1,...,n\}$ induces a prior on $\mathcal{P}$ and, hence, on the space of membership matrices $B$, it is enough to specify a prior on $\{c_i, i=1,...,n\}$.
We assume 
\begin{eqnarray}\label{eq:CRP}
P(c_n = j \mid  c_1, \ldots,c_{n-1})  =
\begin{cases}
\frac{n_j}{n-1+\xi} \quad \text{if}\,  c_n = j \, \text{for some } \,1\leq j \leq K\ \\ 
\frac{\xi}{n-1+\xi}\quad \text{otherwise,} 
\end{cases}
\end{eqnarray}
where $n_j = \# \{i: 1\leq i  < n, c_i=j\}$ and $\xi > 0$ is the precision parameter which controls the prior probability of introducing new clusters.  The expected cluster size under CRP is given by $\sum_{i=1}^n\frac{\xi}{\xi+i-1} \sim \xi \log(\frac{\xi +n}{\xi})$.

\subsubsection{Hyperpriors}\label{sssec:hyper} 
We need to choose hyperpriors for parameters associated with the prior distributions. \\

\noindent {\bf Priors on $\alpha$ and $\theta$ }: 
$\alpha$ is assigned an inverse Gamma distribution, denoted  $\alpha \sim \text{Inv-Gamma}(r,s)$ for constants $r, s > 0$. An inverse Gamma distribution for $\alpha$ allows us to marginalize out $\alpha$ in the posterior distribution, thus obviating the need to sample from its conditional posterior distribution in the Gibbs sampler (refer to Section \ref{SS:post}). Recall that $\theta$ controls the strength of similarity within cluster. Thus a large $\theta$ will encourage tight clusters (elements in each cluster are very similar). We will explore the sensitivity of the final clustering to $\theta$ in Section \ref{sec:sim}. 
We assume a discrete uniform distribution for $\theta$ on the set $\{\theta_1,...,\theta_m\}$, with $P(\theta = \theta_i) = \frac{1}{m}$, $i = 1,...,m$. 

\begin{figure}
\begin{center}
\begin{tabular}{ccc}
\includegraphics[height=1.6in]{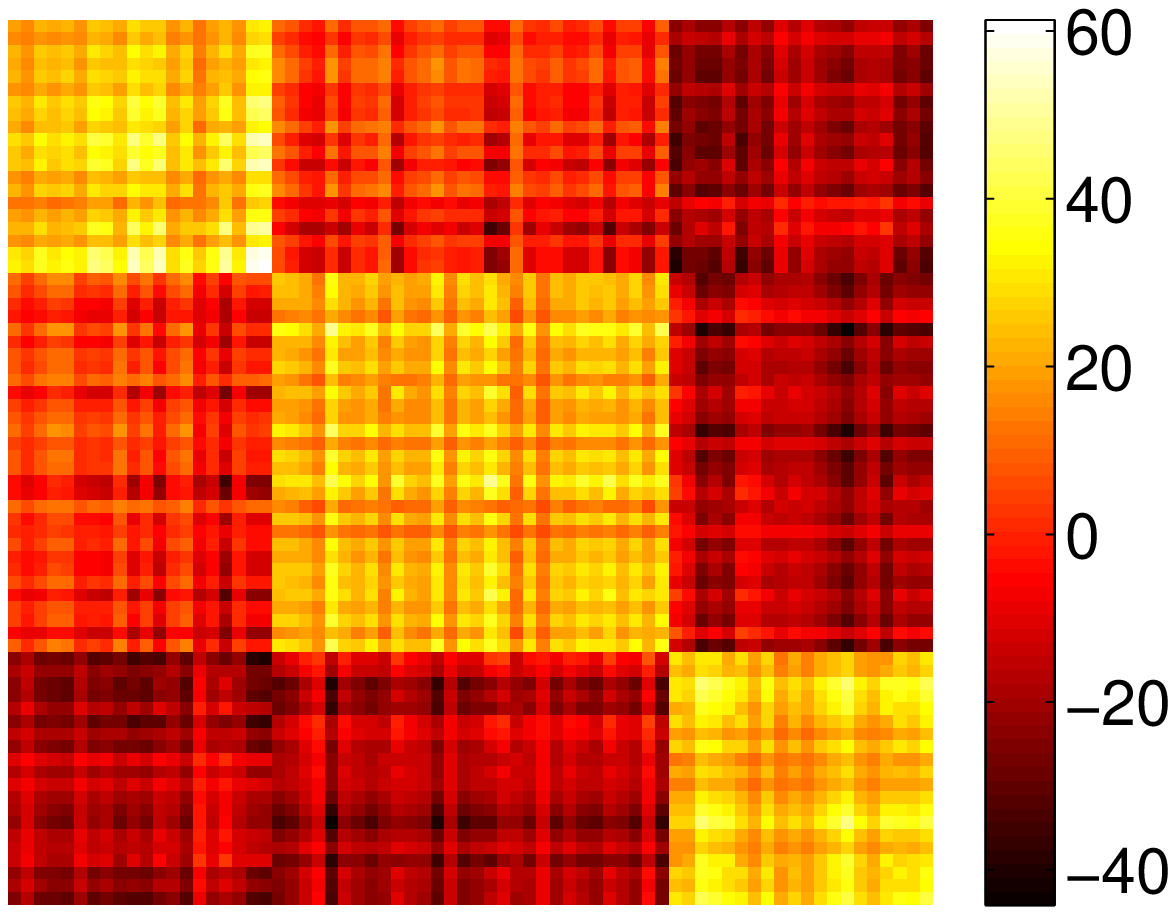}&
\includegraphics[height=1.4in]{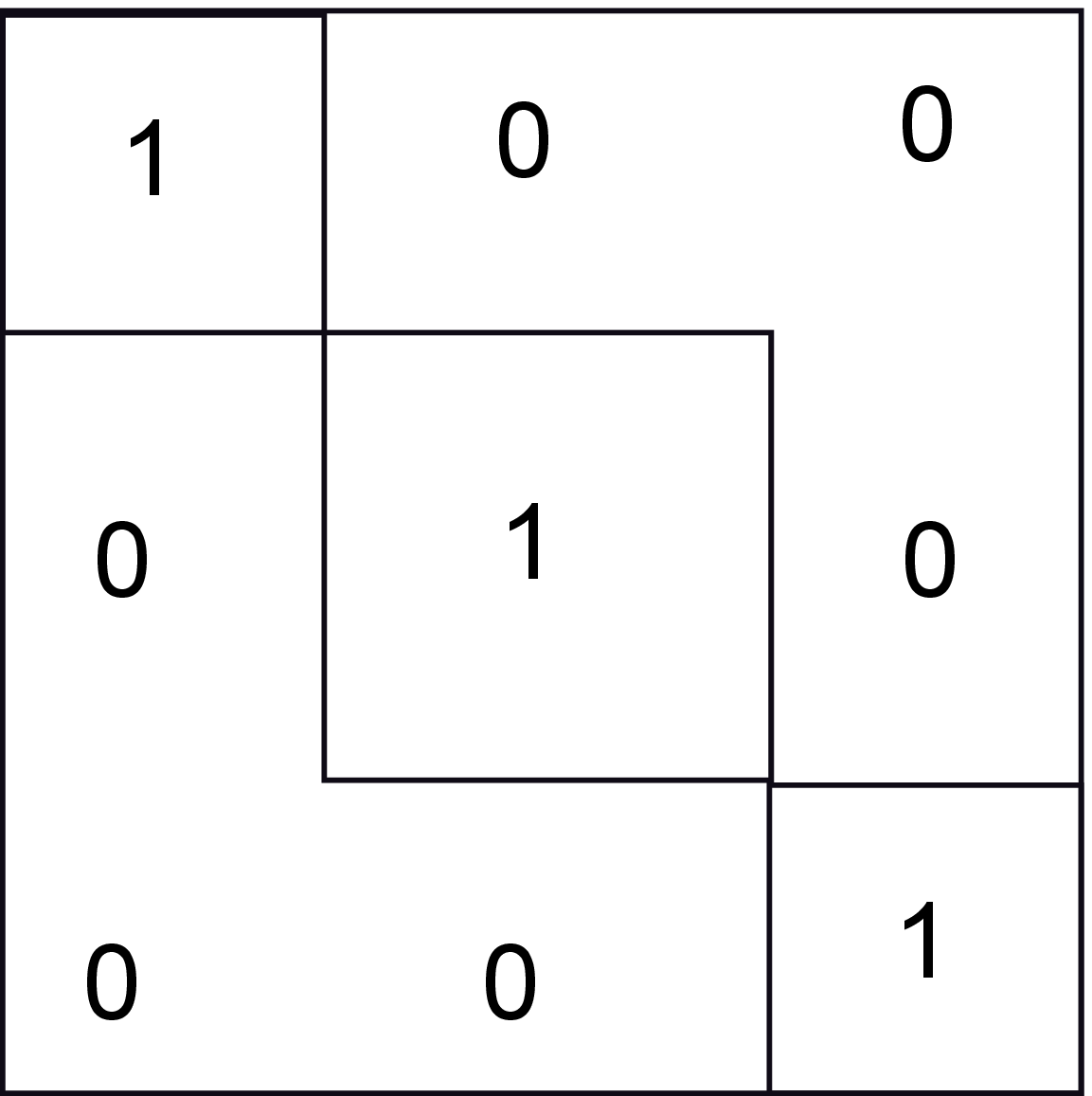}&
\includegraphics[height=1.4in]{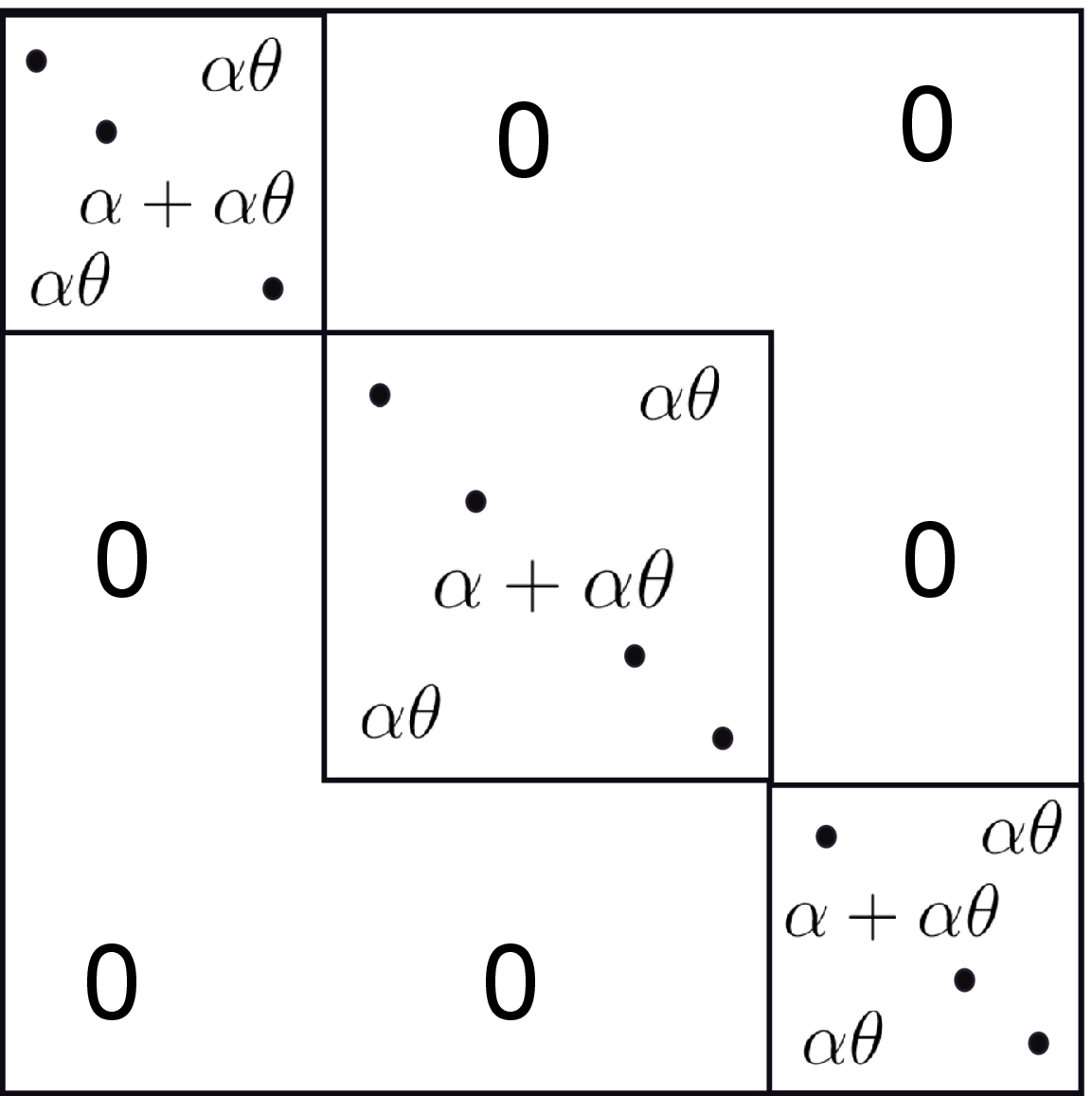}\\
S & B & $\Sigma$ \\
\end{tabular}
\caption{ From the left to right: an inner product matrix $S$,  a partition matrix $B$ and  a scale matrix $\Sigma$.}
\label{fig:pmatrix}
\end{center}
\end{figure}

\noindent {\bf Choice of $\xi$ and $d$:}
Recall that the $\xi$ controls the prior probability of introduction of new clusters in the CRP (\ref{eq:CRP}).  We start with an initial guess of the number of clusters $C_0$ using standard algorithms for shape clustering \citep{Yankov2006,Auder2012}.   In our experience, $C_0/\log n$ provides reasonable choice for $\xi$.  

Also, recall that $d$ is the degrees of freedom for the Wishart distribution.   Since $d$ represents the rank of the inner product matrix $S$, it is natural to estimate $d$ using the number of largest eigenvalues of $S$ which explains $95\%$ of the total variation. This forms an empirical Bayes estimate of $d$, denoted $d_{EB}$.  Let the eigenvalues of $S$ to be 
$\{ \lambda_1,\ldots, \lambda_m\}$, where $m\leq n$ and $\lambda_1 \geq \lambda_2  \geq  \ldots \geq \lambda_m$. $d_{EB}$ is taken to be the smallest integer such that $\frac{\sum_{j=1}^{d_{EB}} \lambda_j}{\sum_{i=1}^m \lambda_i} \geq 0.95.$

\subsubsection{Posterior computation and final selection of clusters}\label{SS:post}
Next, we develop a Gibbs sampling algorithm to sample from the posterior distribution of the unknown  parameters.  To that end, we propose the following simplifications to the likelihood.  The trace and determinant that involve the $\alpha$ and $\theta$ in  equation (\ref{eq:wishartgen}) can be computed analytically \citep{translationinavariantwd,BayesianPt}.  Observe that 
\begin{eqnarray}\label{eq:det}
\abs{\Sigma^{-1}} = \alpha^{-n}\prod_{j=1}^{J}(1+\theta n_j)^{-1} \ ,
\end{eqnarray}
where $n_j$ is the number of elements in $j^{th}$ cluster.  Clearly, the $j^{th}$ cluster corresponds to the $j^{th}$ diagonal block in $B$; refer to Fig. \ref{fig:pmatrix}(a).  Let $S_{jj}, j=1, \ldots J$ be a  {\it sub-square-matrix} in $S$ corresponding to the $j^{th}$ diagonal block in $B$, and $\bar{S}_{jj} = \mathbf{I}_j^t S \mathbf {I}_j$.  Let $\mathbf{I}_j \in \real^{n \times 1}$ be such that the $i^{th}$ element is $\bm{1}(c_i =j)$ for $i=1, \ldots, n$.  Then
\small
\begin{eqnarray}\label{eq:tr}
\mbox{tr}(\Sigma^{-1}S) = \sum_{j = 1}^{J} \frac{1}{\alpha}\bigg\{\mbox{tr}(S_{jj}) - \frac{\theta}{1+n_j \theta}\bar{S}_{jj} \bigg\} = \frac{1}{\alpha}\bigg\{\mbox{tr}(S) - \sum_{j = 1}^{J}\frac{\theta}{1+n_j \theta}\bar{S}_{jj}\bigg\}. 
\end{eqnarray}
\normalsize
Substituting (\ref{eq:det}) and (\ref{eq:tr}) in  (\ref{eq:wishartgen}) with (\ref{eq:decomp}), we obtain, 
\begin{eqnarray} \label{eq:likelihood}
&&P(S\mid B,\alpha,\theta,d) \propto \notag \\ &&\alpha^{-nd/2}\prod_{j=1}^J(1+\theta n_j)^{-d/2} \exp\bigg[-\frac{d}{2 \alpha} \bigg\{\mbox{tr}(S) - \sum_{j=1}^J\frac{\theta}{1+n_j \theta}\bar{S}_{jj}\bigg\}\bigg].
\end{eqnarray} 
If $\alpha \sim \text{Inv-Gamma}(r_0 d/2,s_0 d/2)$ it is possible to integrate out $\alpha$ analytically in (\ref{eq:likelihood})  as $P(S\mid B,\theta,d) = \int P(\alpha) P(S\mid B,\alpha,\theta,d)d\alpha$ yielding
\begin{equation}
P(S\mid B,\theta,d) \propto \prod_{j=1}^{J}(1+\theta n_j)^{-d/2}\bigg[\frac{d}{2} \bigg\{\mbox{tr}(S) - \sum_{j=1}^J\frac{\theta}{1+n_b \theta}\bar{S}_{jj} + s_0 \bigg\}\bigg]^{-(n+r_0)d/2} \ .
\label{equ:LF}
\end{equation}

\begin{figure}
\begin{center}
\begin{tabular}{c}
\includegraphics[height=2.6in]{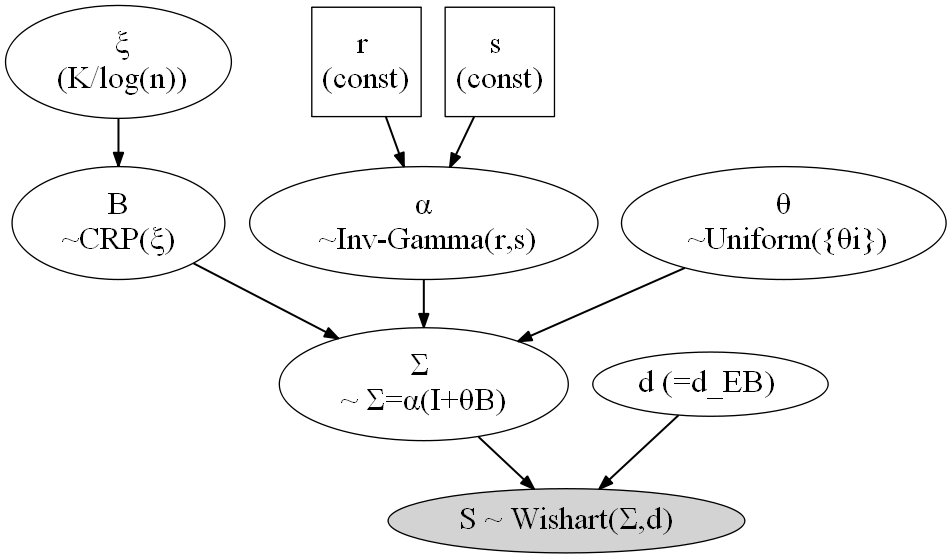}
\end{tabular}
\caption{Graphical model representation of our Bayesian model. Squares indicate fix parameters and circles indicate random variables.}
\label{fig:graphmodel}
\end{center}
\end{figure}

Using the prior distributions for $\theta,  B$ with the $d_{EB}$ plugged in the likelihood $(\ref{eq:likelihood})$, we get the posterior distribution of the membership matrix $B$:
\begin{equation}
P(B|S,\theta,d,\xi) \propto P(S|B,\theta,d_{EB})P(B|\xi)P(\theta).
\label{equ:FM}
\end{equation}
Fig. \ref{fig:graphmodel} shows the graphical model representation of our Bayesian model. Some suggestions of specifying hyper-priors are summarized in Table \ref{tab:hyperpriors}. We use Markov chain Monte Carlo (MCMC) algorithm to obtain posterior samples $B^{(1)},...,B^{(M)}$
 for a suitable large integer $M > 0$ using (\ref{equ:FM}). The detailed algorithm is described in the following.

\begin{table}
\caption{Suggestions for specifying hyper-priors in our model. }
\label{tab:hyperpriors}
\begin{tabular}{c|l|l}
\hline
   Hyper-parameters& Description & Suggested values \\
	\hline
		 $\theta$&  \begin{tabular}{@{}c@{}}Parameter for $\Sigma$, \\  $\Sigma = \alpha(I + \theta B)$ \end{tabular} & \begin{tabular}{@{}c@{}} $\theta \sim \text{Uniform}(\theta_1,...,\theta_n)$ \\ large $\theta$ - tight clusters, small $\theta$ - loose clusters  \end{tabular} \\ 
	\hline
	$\alpha$& Parameter for $\Sigma$ & $\alpha \sim \text{Inv-Gamma}(r,s)$, where $r,s$ are constants \\
	\hline
		$d$&  Degrees of freedom of Wishart & Estimated using the rank of $S$ \\
	\hline
	$\xi$& Parameter for CRP & $\xi = K/\log(n)$, $K$ is initial estimated \# of clusters  \\
\hline
\end{tabular}
\end{table}

\begin{algorithm} \label{algMCMC}
{\bf Posterior sampling using the MCMC} \\
Given the prior parameters $d_{EB}$, $r$, $s$, $\xi$, $\theta$, and the inner product matrix $S$ from the $n$ observations, and let $N_\theta = \text{length}(\theta)$, we want to sample $Iter$ number of posterior samples of membership matrices $B$:  
\begin{enumerate}
\item Initialize the cluster number $K$ (a large integer), the cluster indices $\{c _i, i=1,\ldots, n\}$ and obtain the initial membership matrix $B^{(0)}$;

\item \label{mcmc:mainiter} {\bf For} each sweep of the MCMC ($it = 0$ to $Iter$)

 \begin{enumerate}
 \item {\bf For} each $\theta_i, i=1, \ldots, N_{\theta}$ , obtain posteriors $P({\theta_{i}|\cdot}) \propto P(S|\theta_i, B^{(it)},d_{EB})p(\theta_i)$ using \eqref{equ:LF}. Normalize $\{P({\theta_i}|\cdot)$\} and sample $\theta^{it}$ from the discrete distribution on the support points $\theta_i, i=1, \ldots, N_{\theta}$ with probabilities $\{P({\theta_1}|\cdot)$ $,\ldots,P({\theta_{N_\theta}}|\cdot)\}$. The complexity for this step is $O(N_\theta*k_{B^{(it)}})$, where $k_{B^{(it)}}$ is the number of clusters obtained from $B^{(it)}$. 

 \item \label{mcmc:ob} {\bf For} each observation ($i = 1$ to $n$)
       \begin{enumerate}
			      \item {\bf For} each cluster ($j = 1$ to $k_{B^{(it)}} + 1$)
						\begin{enumerate} 
						\item Assign current observation ($y_i$) to the $j$-th cluster, update the membership matrix $B^{(it)}$ to $B^{\prime}_{it}$, and calculate the posterior $ \pi_{j} = P(B^{\prime}_{it} |S,\theta^{it},d_{EB}, \xi)$ using  \eqref{equ:FM}. The complexity for this step is $O(1)$ \footnote{Since only one observation changes the cluster index, one can explicitly calculate the difference between the old values of  \eqref{eq:det} and \eqref{eq:tr} and new values in O(1) steps.}.
				\end{enumerate}
			\item Normalize $\{\pi_1,...,\pi_{k_{B^{(it)}} + 1}\}$ and sample $c_i$ from a discrete distribution on support points  $\{1 ,2, \ldots, k_{B^{(it)}} + 1\}$ with probabilities $(\pi_1, \ldots, \pi_{k_{B^{(it)}} + 1})$. Update $B^{(it)}$. Complexity for this step is  $O\{\log(k_{B^{(it)}})\}$ \citep{Bringmann2011}.
		\end{enumerate}
\item After completing Step \ref{mcmc:ob}, we obtain one MCMC sample of $B^{(it)}$.  
 \end{enumerate} 
\item Repeat step \ref{mcmc:mainiter} so that we have $Iter$ many samples. Discard the first few  samples (burn-in), and relabel the  remaining $B^{(it)}$s as $B^{(1)}, \ldots, B^{(M)}$. 
\end{enumerate}

\end{algorithm} 

From algorithm \ref{algMCMC},  the complexity of each sweep of the MCMC is $O\{N_\theta K +nK\log K\}$. Usually $N_{\theta} \leq n$, leading to an overall complexity of $O(nK\log (K))$. 

Once we obtain the posterior samples $\{B^{(i)}, i=1, \ldots, M\}$, our goal is to estimate the clustering configuration. However, the space of membership matrices $B$ is huge,  and we would expect the posterior to explore only an insignificant fraction of the space based on a moderate values of $M$. 
Therefore, instead of using the mode of $\{B^{(i)}, i=1, \ldots, M\}$,  we devise the following alternate strategy to estimate the clustering configuration more accurately. We treat the set of the membership matrices, denoted as $\mathcal{F}_B$, as a subset of symmetric $n \times n$ matrices with restrictions:  (1) $ B(i,j) = \{0,1\} \text{ for all } i,j = 1, ..., n $; (2) $B(i,\cdot) = B(j,\cdot)$ and $B(\cdot,i) = B(\cdot,j)$  if $i^{th}$ observation and $j^{th}$ observation are in the same cluster. The final matrix $B^*$ is obtained by calculating the {\bf extrinsic mean} of the posterior samples defined as follows. 

\begin{figure}
\begin{center}
\begin{tabular}{ccc}
\includegraphics[height=1.2in]{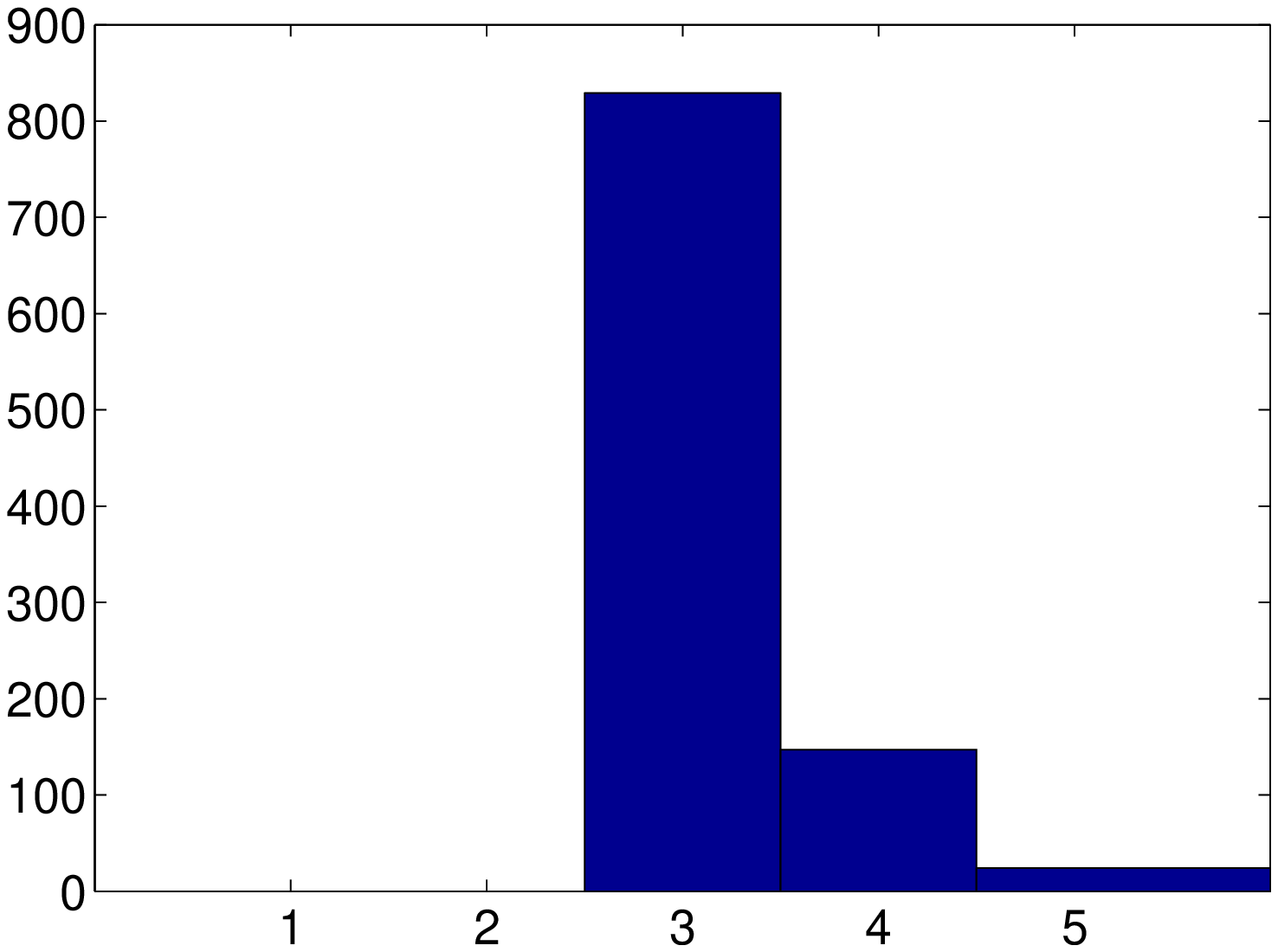}&
\includegraphics[height=1.2in]{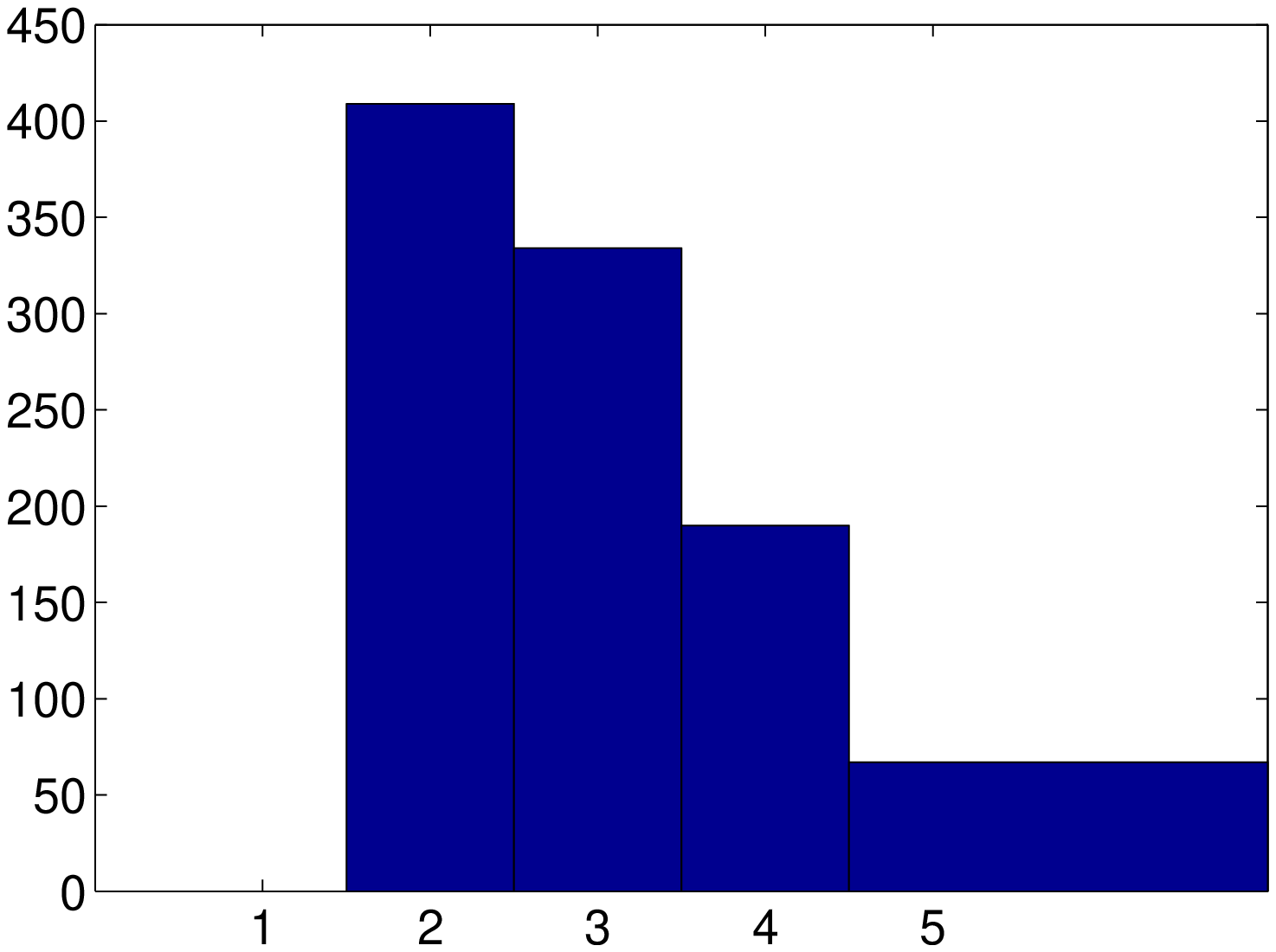}&
\includegraphics[height=1.2in]{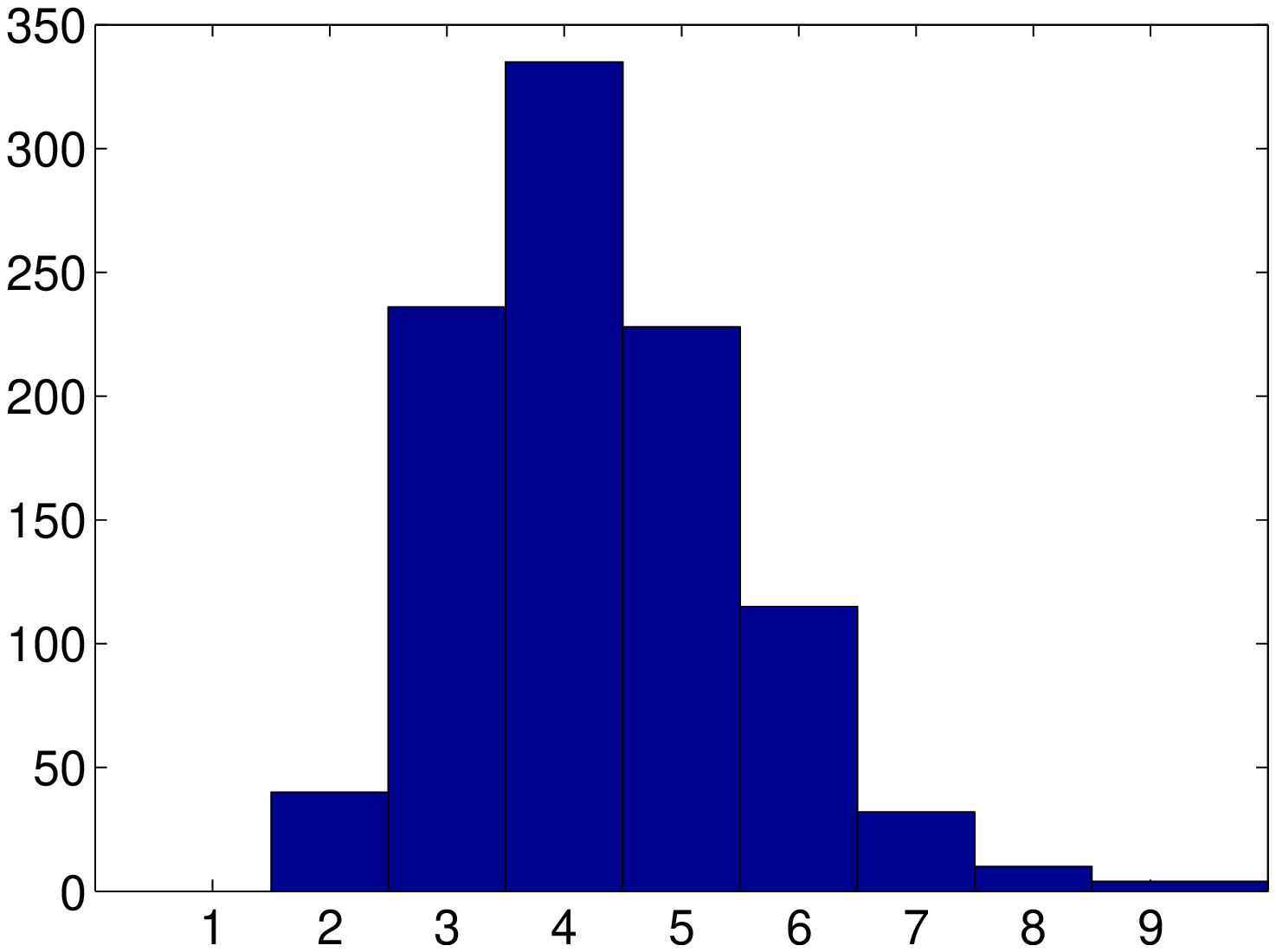}\\
(a)&(b)&(c)\\
\end{tabular}
\caption{Posterior distribution of clustering number $k$.}
\label{fig:histpost}
\end{center}
\end{figure}

\begin{algorithm} \label{extrisicmean} 
{\bf Calculating extrinsic mean of membership matrices}

Given the samples $B^{(1)}, \ldots, B^{(M)}$, the extrinsic mean $B^*$ is calculated as the following:
\begin{enumerate}
\item Find the mode of the number of clusters $k_0$ based on the samples $B^{(1)}, \ldots, B^{(M)}$.
\item Calculate the Euclidean mean and threshold it onto the set of membership matrices ($\mathcal{F}_B$): 
  \begin{enumerate} 
		\item {\bf Euclidean mean}: Let $\bar{B}= \frac{1}{M}\sum_{t=1}^M B^{(t)}$.
		\item {\bf Thresholding}: threshold the Euclidean mean onto $\mathcal{F}_B$: $B^* = \text{threshold} (\bar{B}, t^*)$, where $t^*$ is the largest threshold such that $B^*$ has $k_0$ clusters. 
		Setting $k = N$ and $iter = M$, the thresholding procedure is described below: \\
	 {\bf While} ($k \neq k_0$), do
	
	\begin{enumerate}
	   \item Set $J_{array} = \{1,...,N\}$, $B^* = \text{zeros}(N,N)$. Also set $iter = iter - 1$, let $t^* = iter/M$. 
		 \item {\bf For} $j$ in $J_{array}$, calculate
		     \begin{enumerate}
				  \item ${\bf v} = \bm{1} (\bar{B}(j,\cdot)>t^*)$; record the index of elements in ${\bf v}$ equal to $1$, denoted as set $C$. Let $J_{array} = J_{array} - C$, which means remove elements in $C$ from $J_{array}$.
					\item {\bf For} $i$ in set $C$, set $B^*(i,\cdot) = {\bf v}$, $B^*(\cdot,i) = {\bf v}^t$ and $\bar{B}(i,\cdot) = {\bf 0}, \bar{B}(\cdot,i) = {\bf 0}^t$. 
				 \end{enumerate}
			\item Set $k = \#B^*$, which is number of clusters in $B^*$. 
	\end{enumerate}
	
   \end{enumerate}
 
\end{enumerate}
\end{algorithm}
Fig. \ref{fig:histpost} shows some generic illustrations of the posterior distribution on the number of clusters obtained from $B^{(1)}, \ldots, B^{(M)}$. One may notice that the Euclidean mean $\bar{B} \notin \mathcal{F}_{B}$. Actually $\bar{B}(i,j)$ represents the posterior probability that the $i^{th}$ and the $j^{th}$ observations are clustered together. To  project $\bar{B}$ into $\mathcal{F}_{B}$, we find the largest value $t^*$ to threshold $\bar{B}$, such that the thresholded $\bar{B}$, denoted as $B^*$, has $k_0$ clusters, and $B^* \in \mathcal{F}_{B}$. 
In other words,  we assign two data points to the same cluster if the posterior probability of being clustered together is greater than or equal to  $t^*$.  It is rare that we can not find a $t^*$ to threshold $\bar{B}$ to obtain $B^* \in \mathcal{F}_B$ with $k_0$ clusters. In this circumstance, one can either sample more posteriors, or re-sample them.  

\section{Experimental Results} \label{sec:sim}

In this section, we demonstrate the performance of our model (Wishart-CRP, denoted by W-CRP) both on  synthetic data (in Section \ref{ssec:syn}) and the case studies (in Section \ref{ssec:real}) introduced earlier in Section \ref{sec:data}.    For the Euclidean datasets, we generated 8000 samples from the posterior distribution and discarded a burn-in of 1000, whereas those numbers for the non-Euclidean data are 4000 and 1000, respectively.   Convergence was monitored using trace plots of the deviance as well as several parameters.   The high effective sample size of the main parameters of interest shows good mixing of the Markov chain.   Also we get essentially identical posterior modes with different starting points and moderate changes to hyperparameters.   



\subsection{Synthetic Examples}\label{ssec:syn}

 We consider several simulation settings for both Euclidean and non-Euclidean datasets. 
\subsubsection { Euclidean Data}
\noindent {\bf Elicitation of hyperpriors:}

\begin{figure}
\begin{center}
\begin{tabular}{cc||cc}
\hline
\hline
 \multicolumn{2}{c}{Small $\xi$, small $\theta$} &  \multicolumn{2}{c}{Big $\xi$, small $\theta$} \\
\includegraphics[height=1.0in]{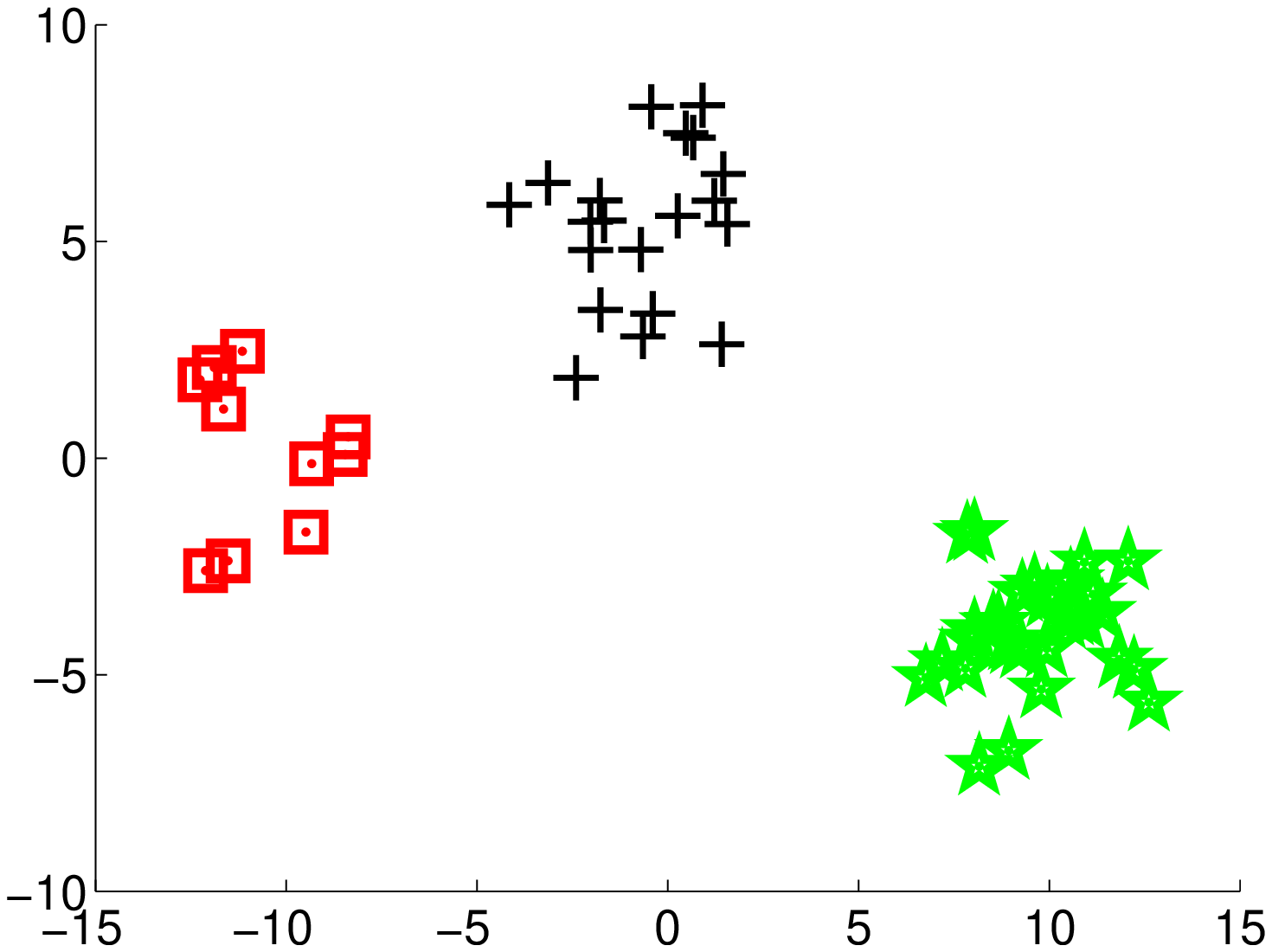}&
\includegraphics[height=1.0in]{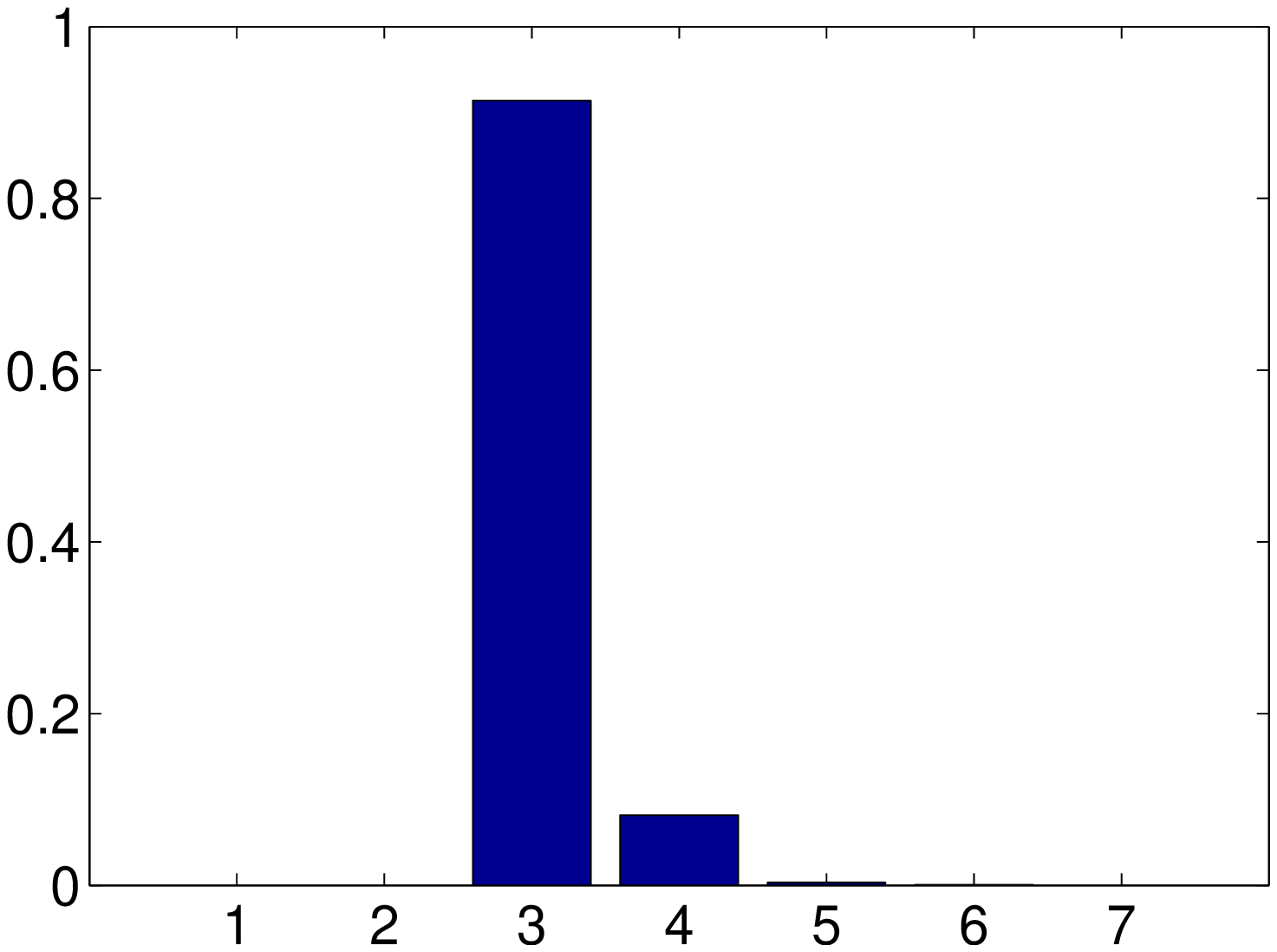}&
\includegraphics[height=1.0in]{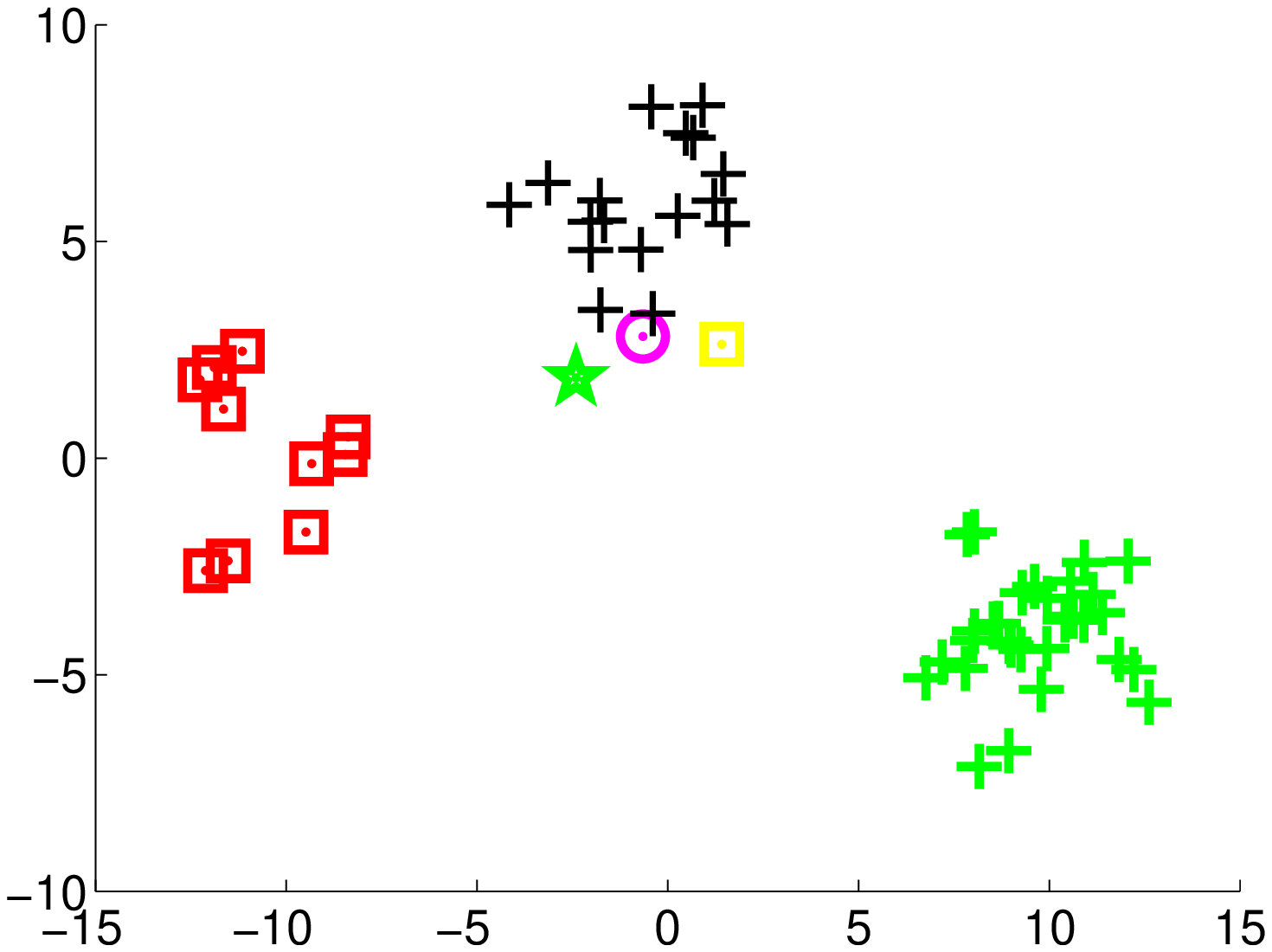}&
\includegraphics[height=1.0in]{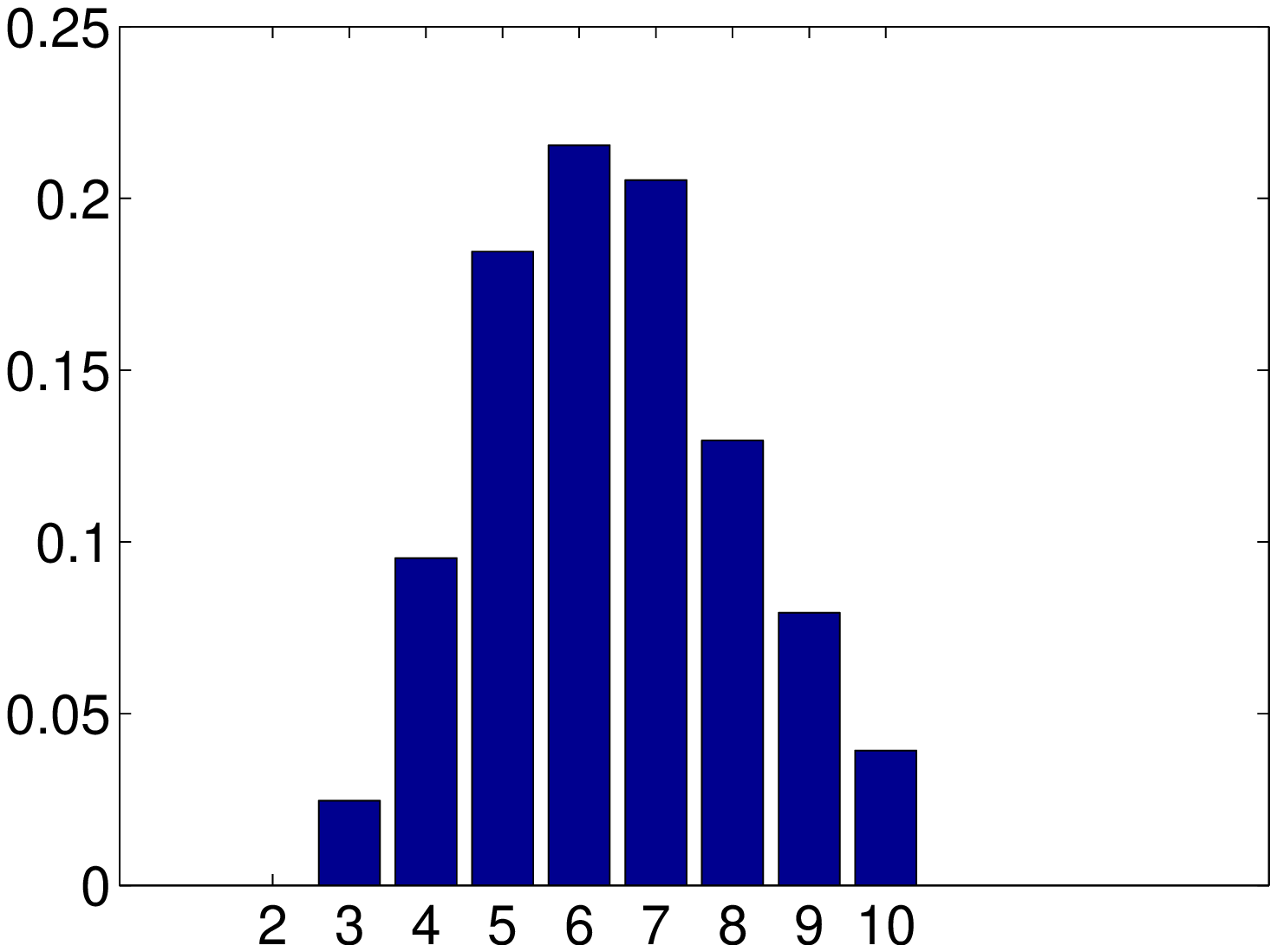}\\
\includegraphics[height=1.0in]{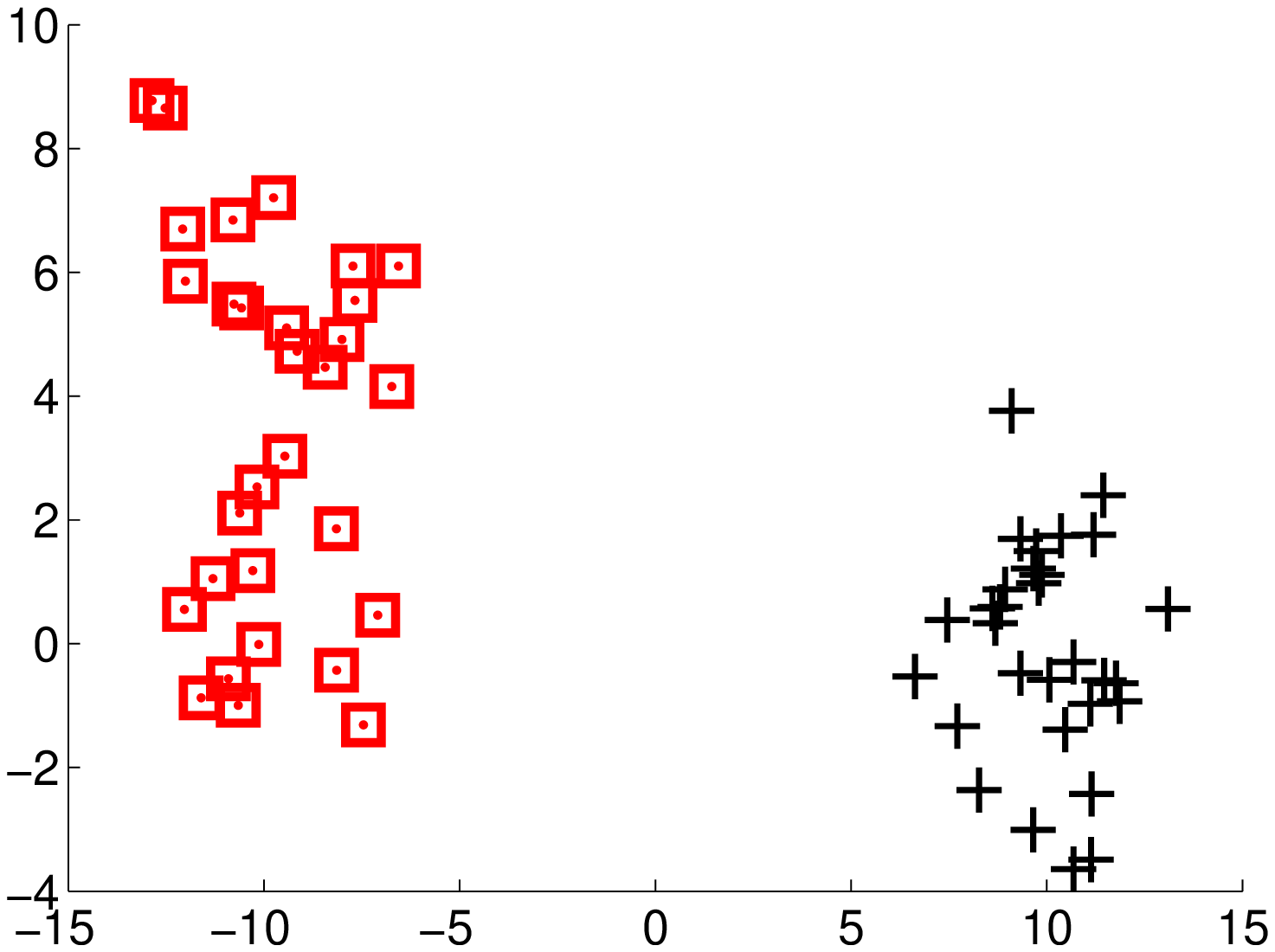}&
\includegraphics[height=1.0in]{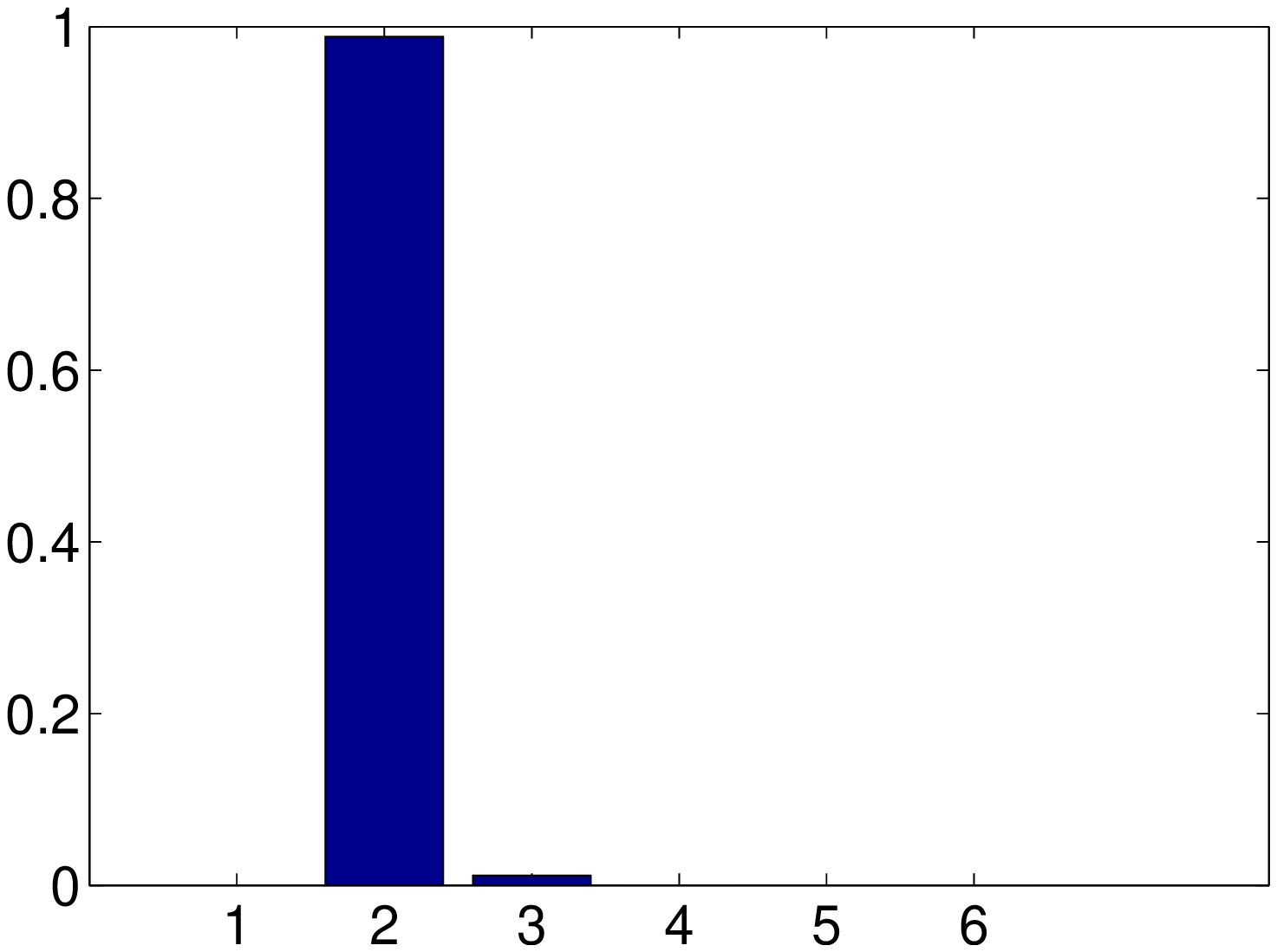}&
\includegraphics[height=1.0in]{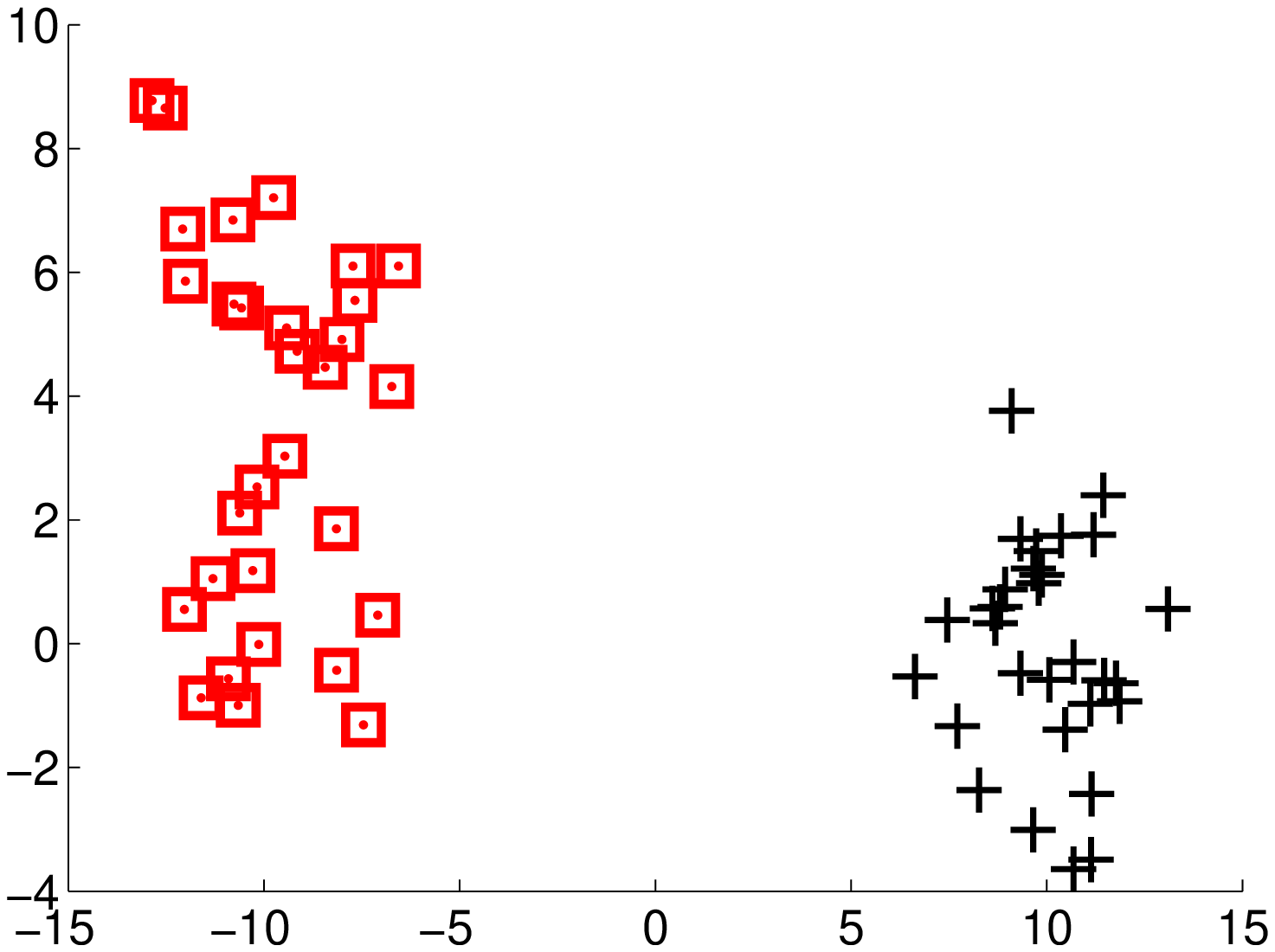}&
\includegraphics[height=1.0in]{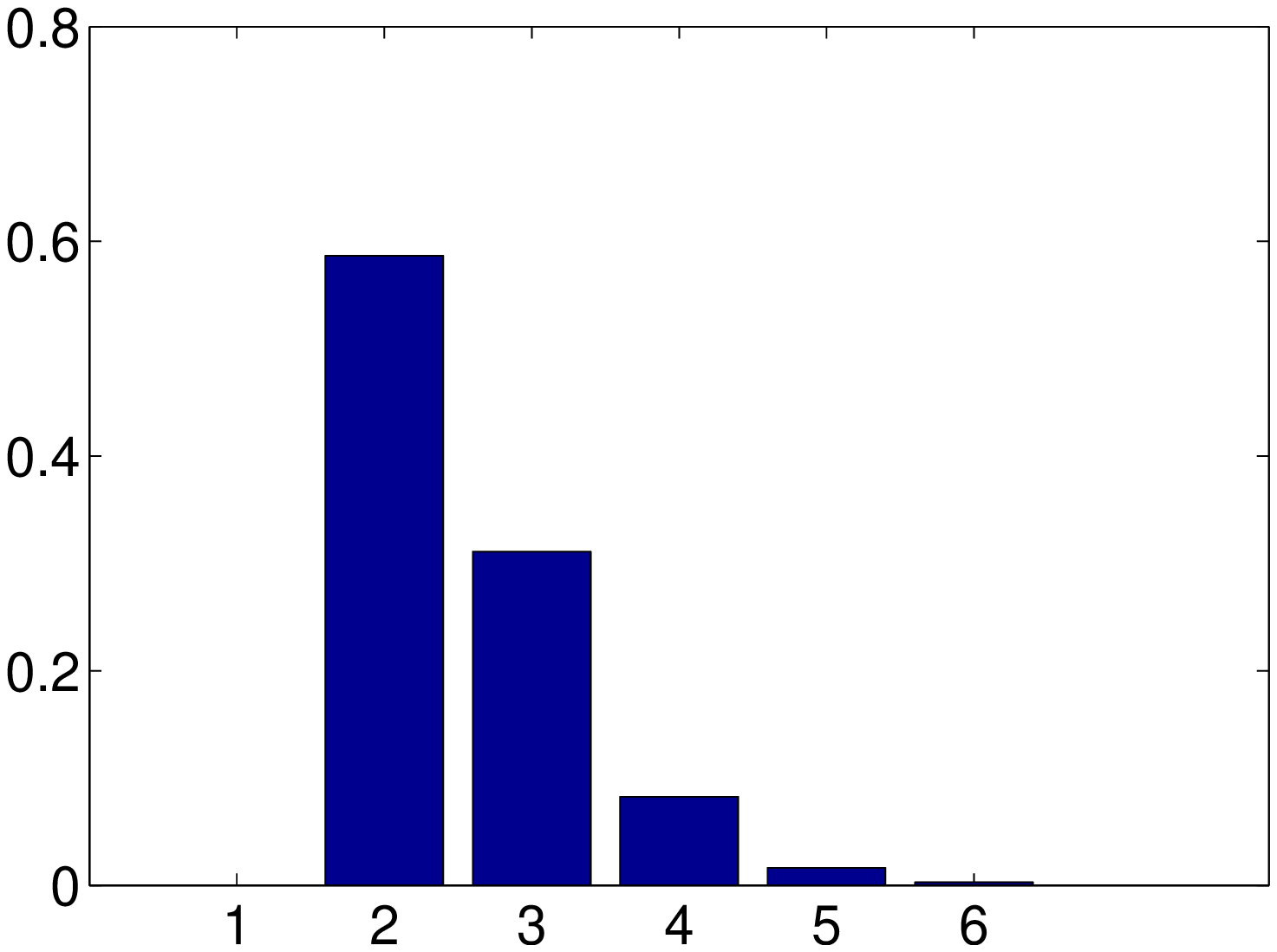}\\
\includegraphics[height=1.0in]{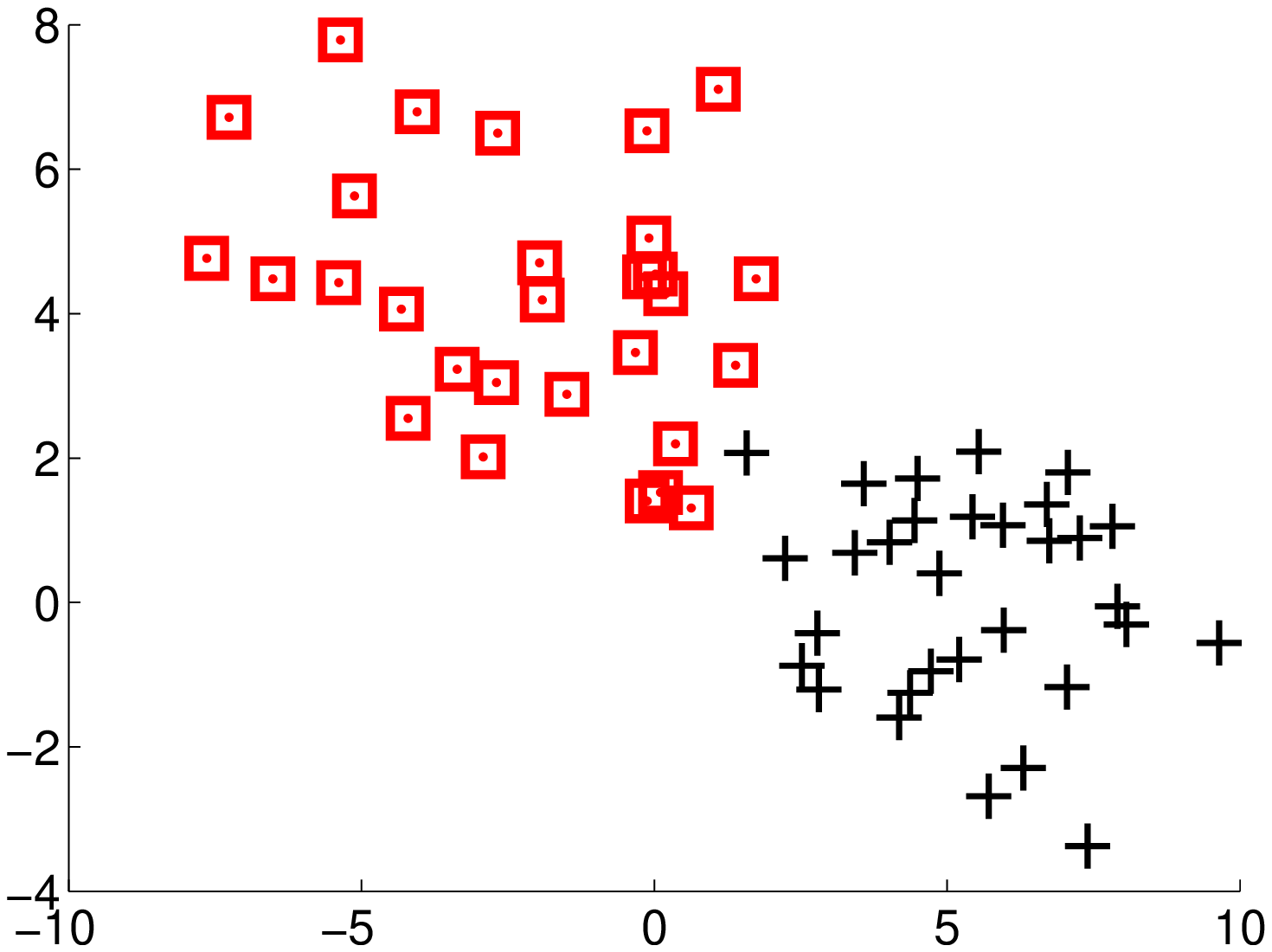}&
\includegraphics[height=1.0in]{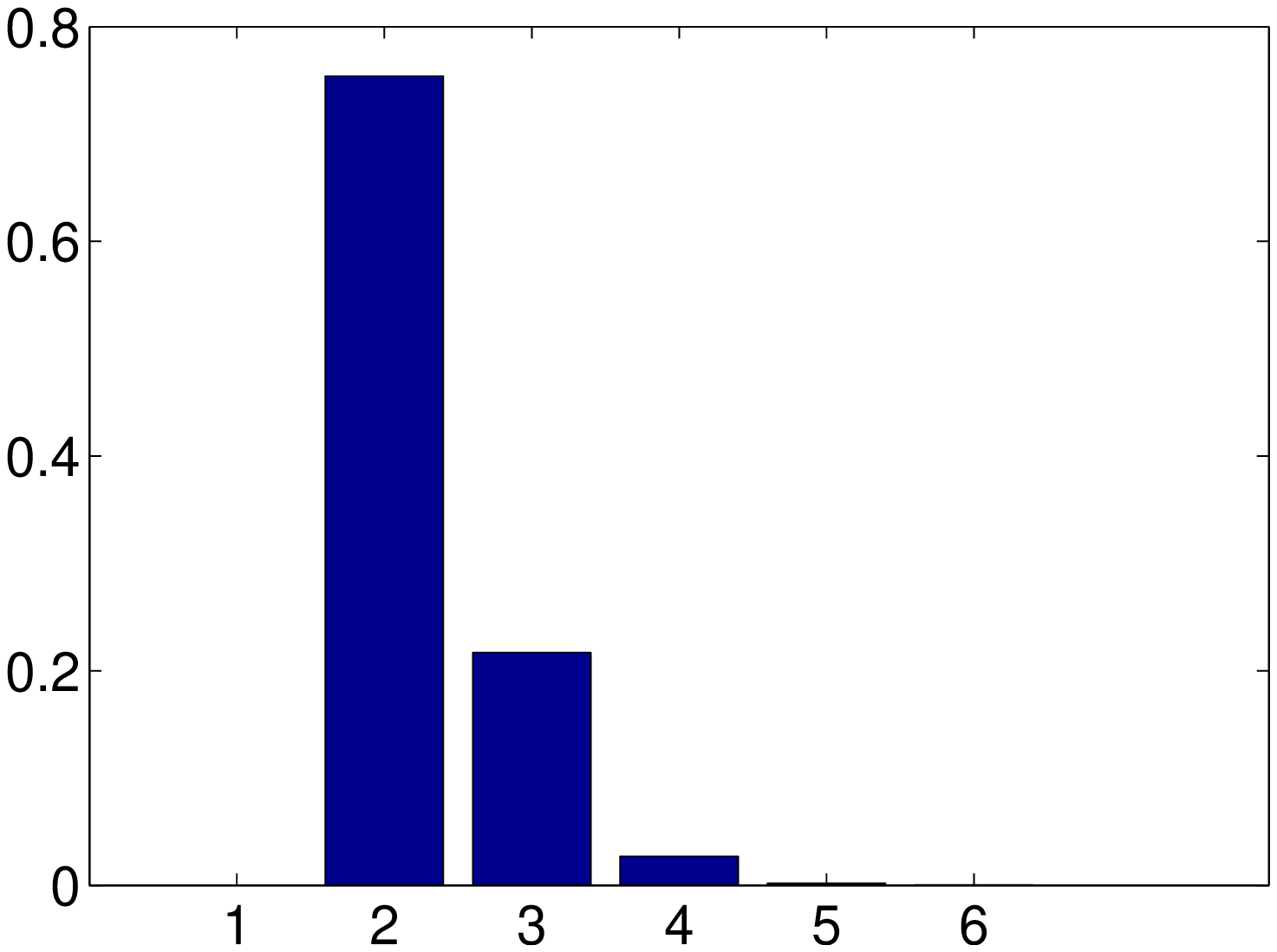}&
\includegraphics[height=1.0in]{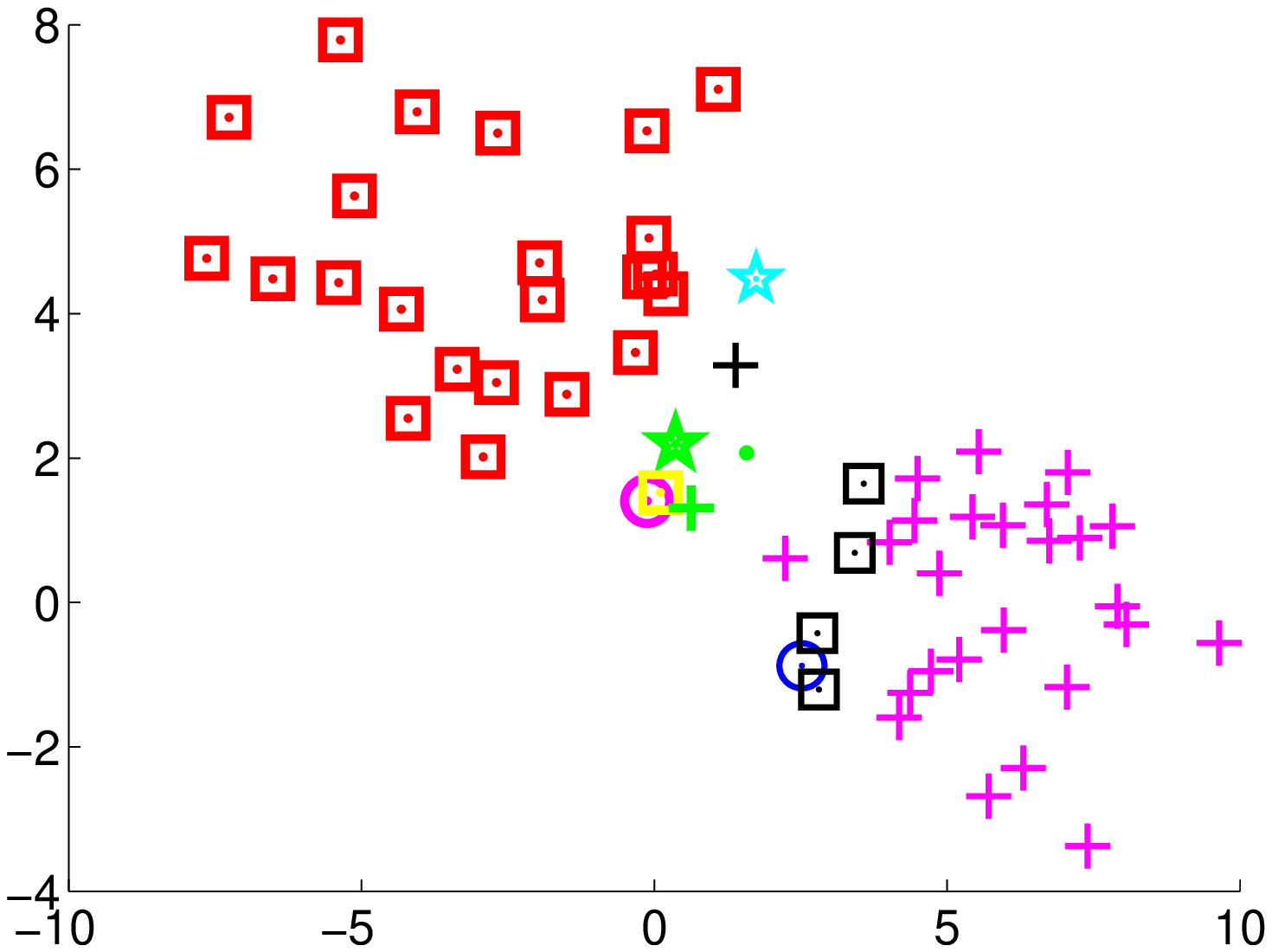}&
\includegraphics[height=1.0in]{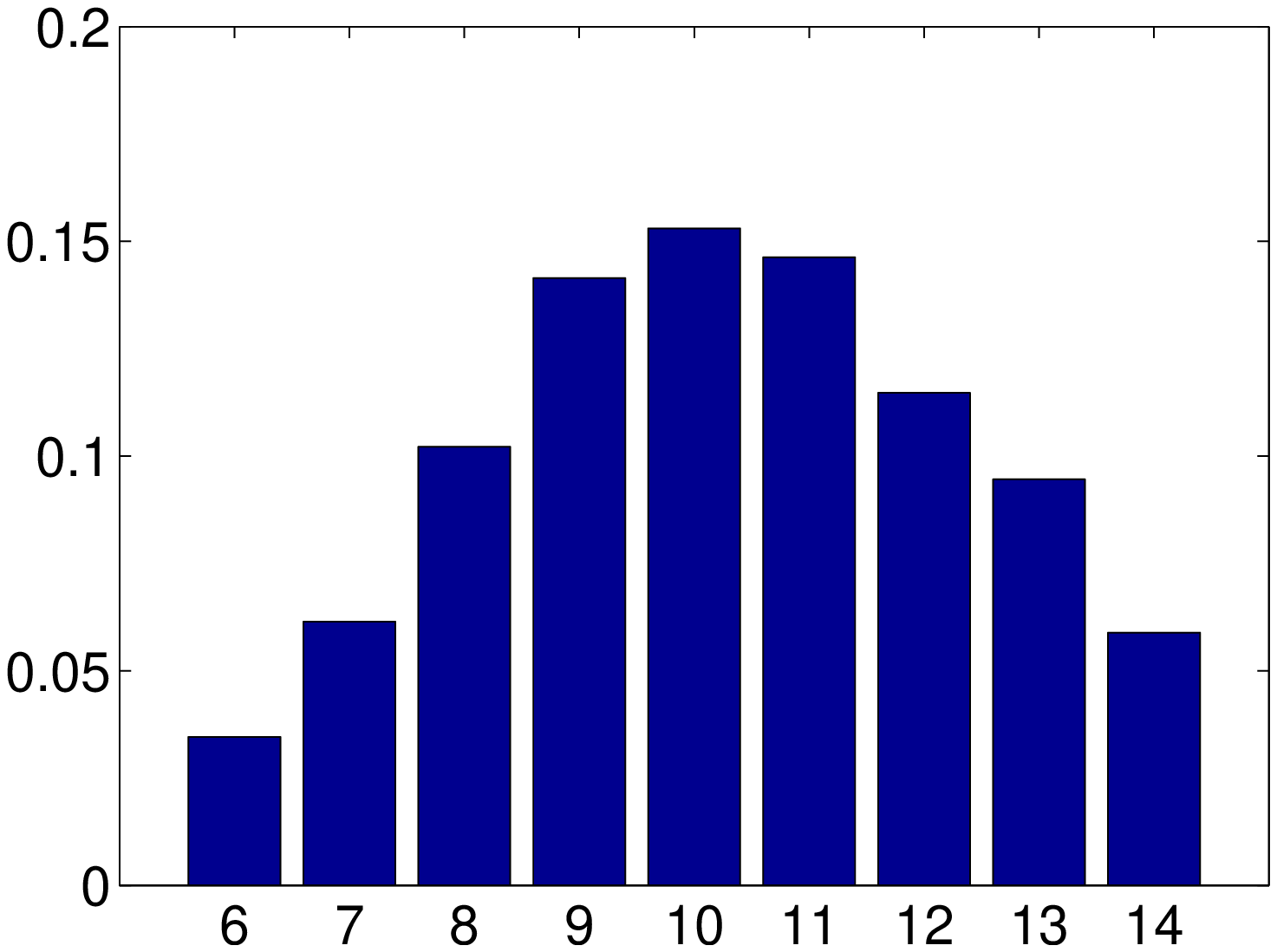}\\
 \multicolumn{2}{c||} {$\xi=0.2$, $\theta \in \{1,2,3,4,5\}\times 10^{-1}$} &  \multicolumn{2}{c} {$\xi = 10$, $\theta \in \{1,2,3,4,5\} \times10^{-1}$}\\
\hline
\hline
 \multicolumn{2}{c}{Small $\xi$, big $\theta$} &  \multicolumn{2}{c}{Big $\xi$, big $\theta$}\\
\includegraphics[height=1.0in]{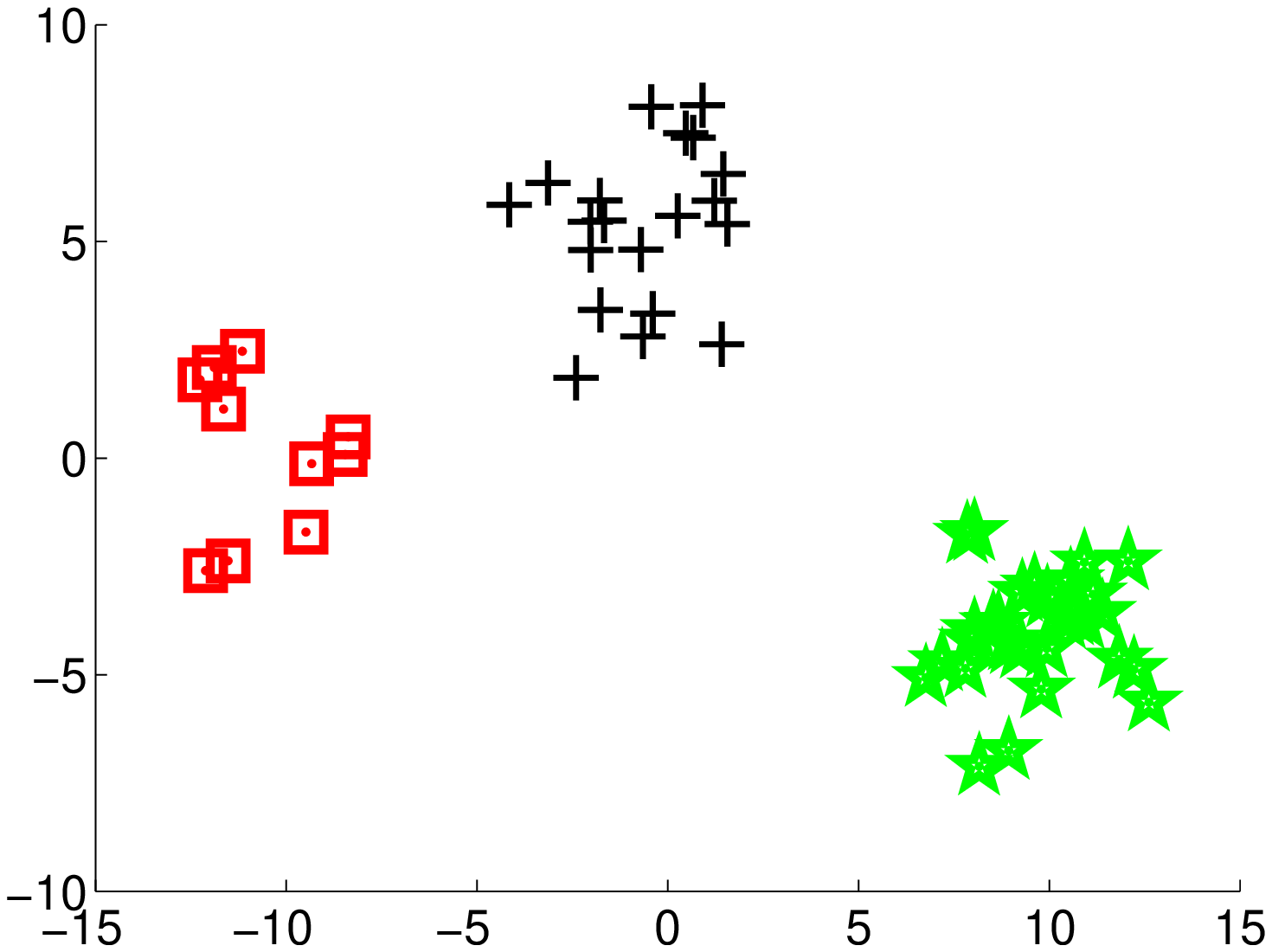}&
\includegraphics[height=1.0in]{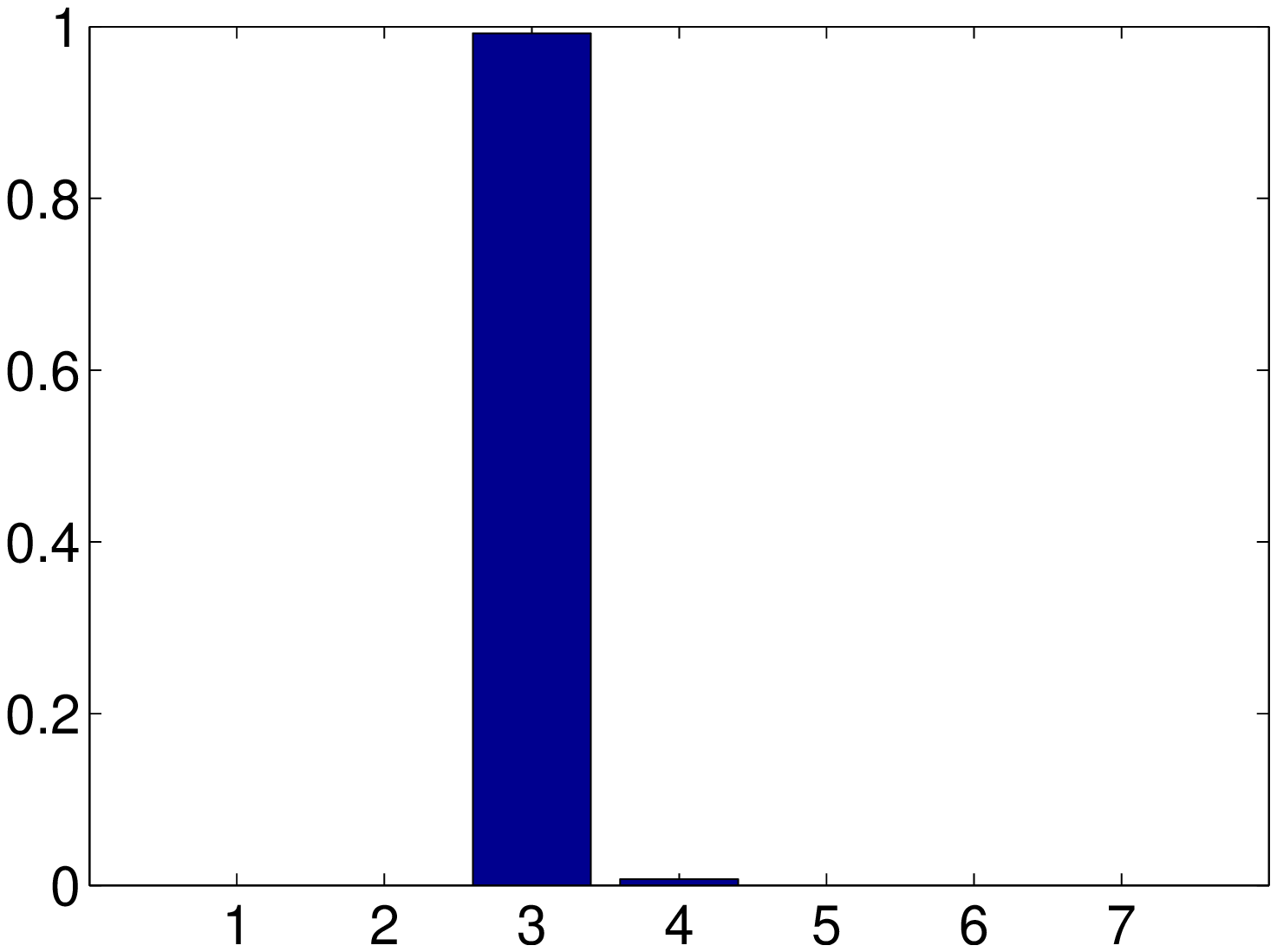}&
\includegraphics[height=1.0in]{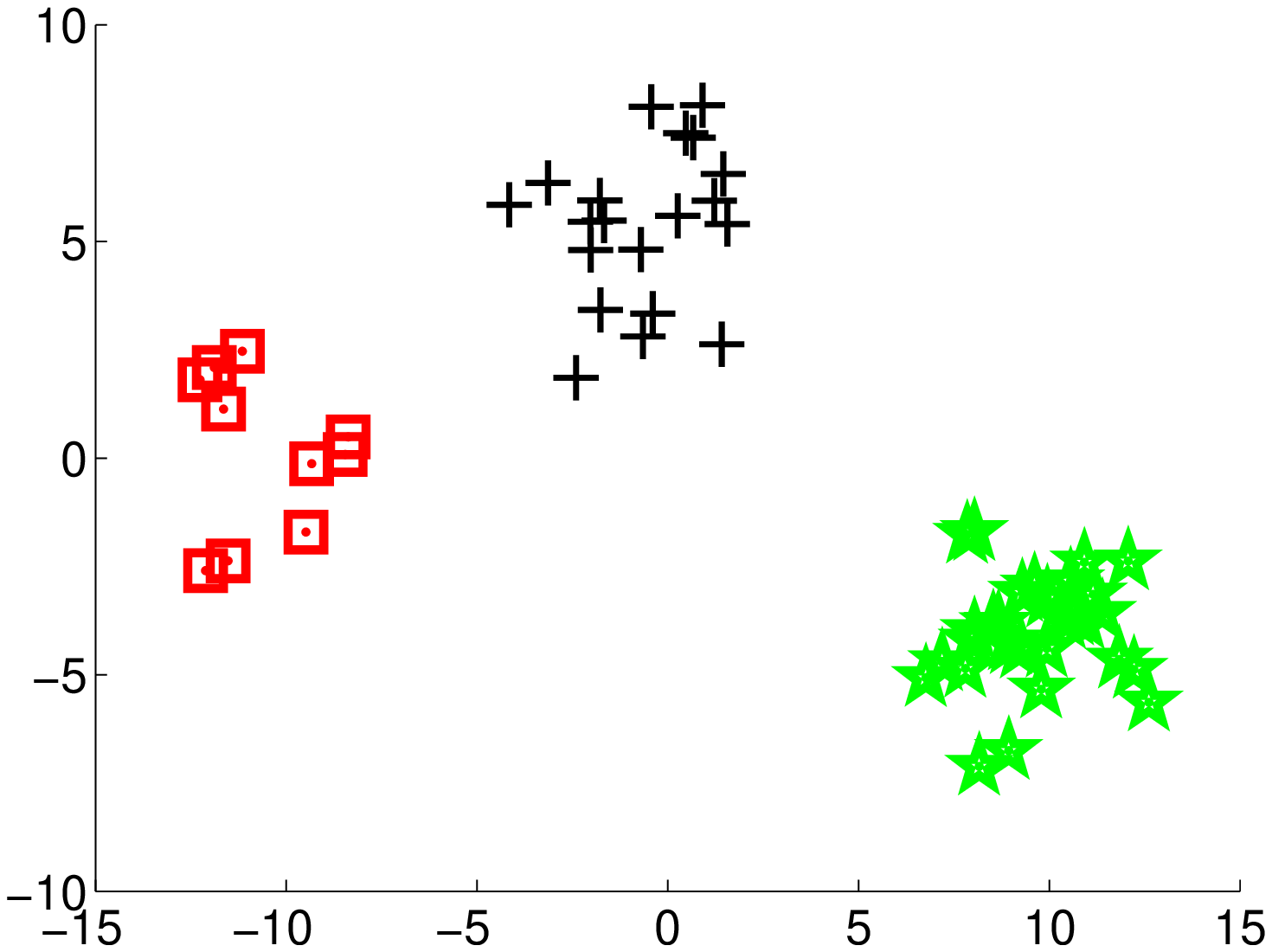}&
\includegraphics[height=1.0in]{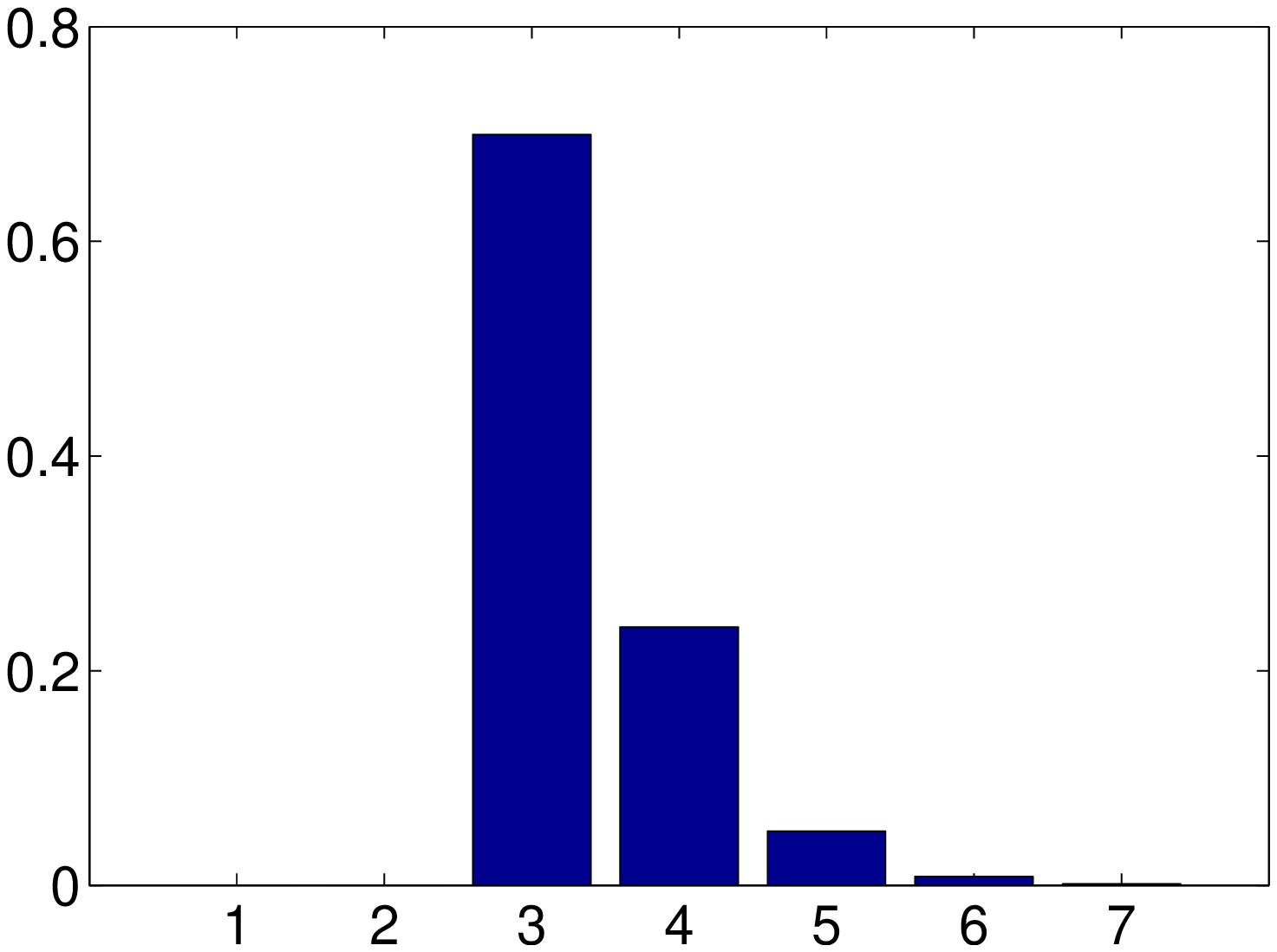}\\
\includegraphics[height=1.0in]{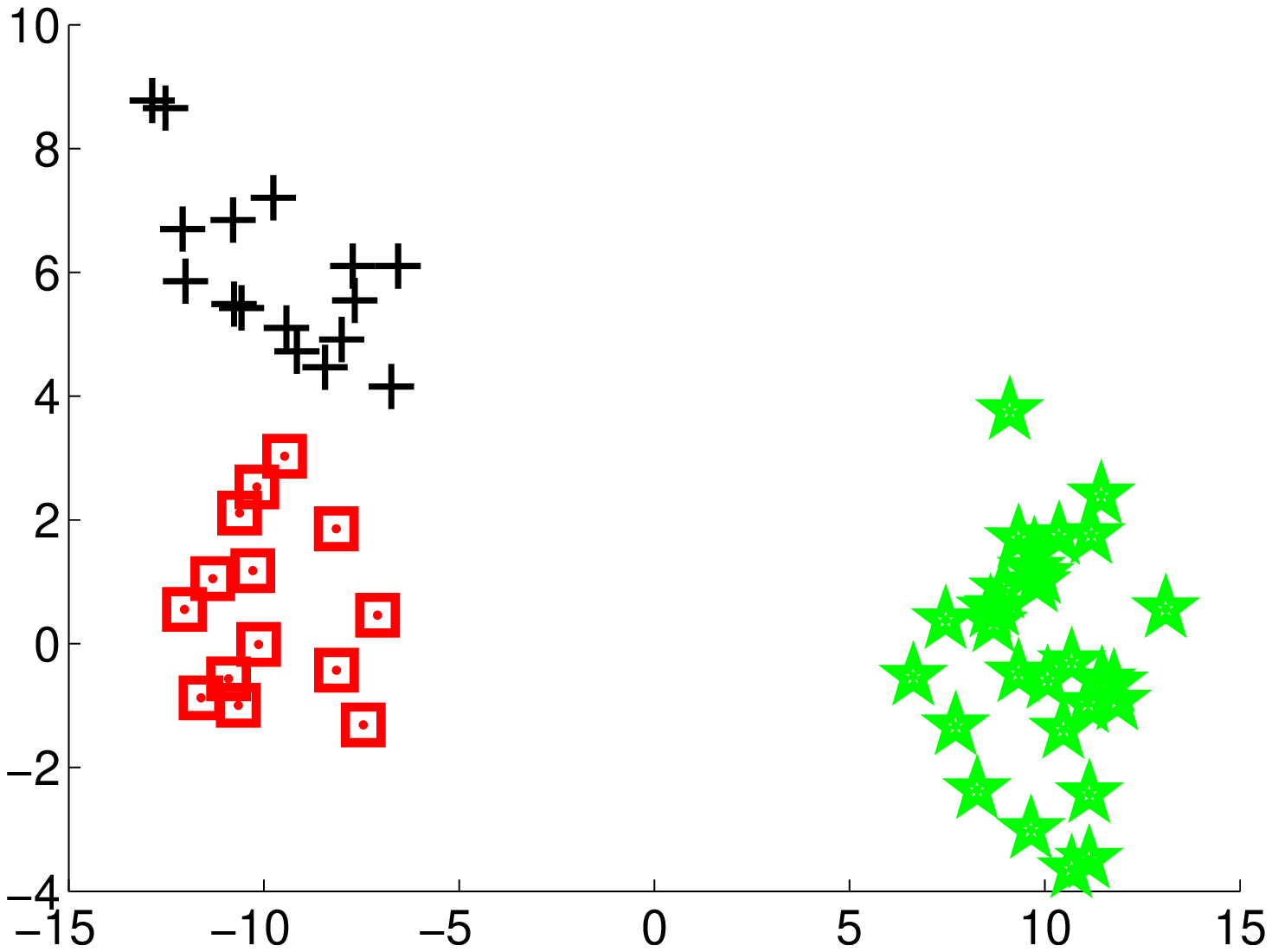}&
\includegraphics[height=1.0in]{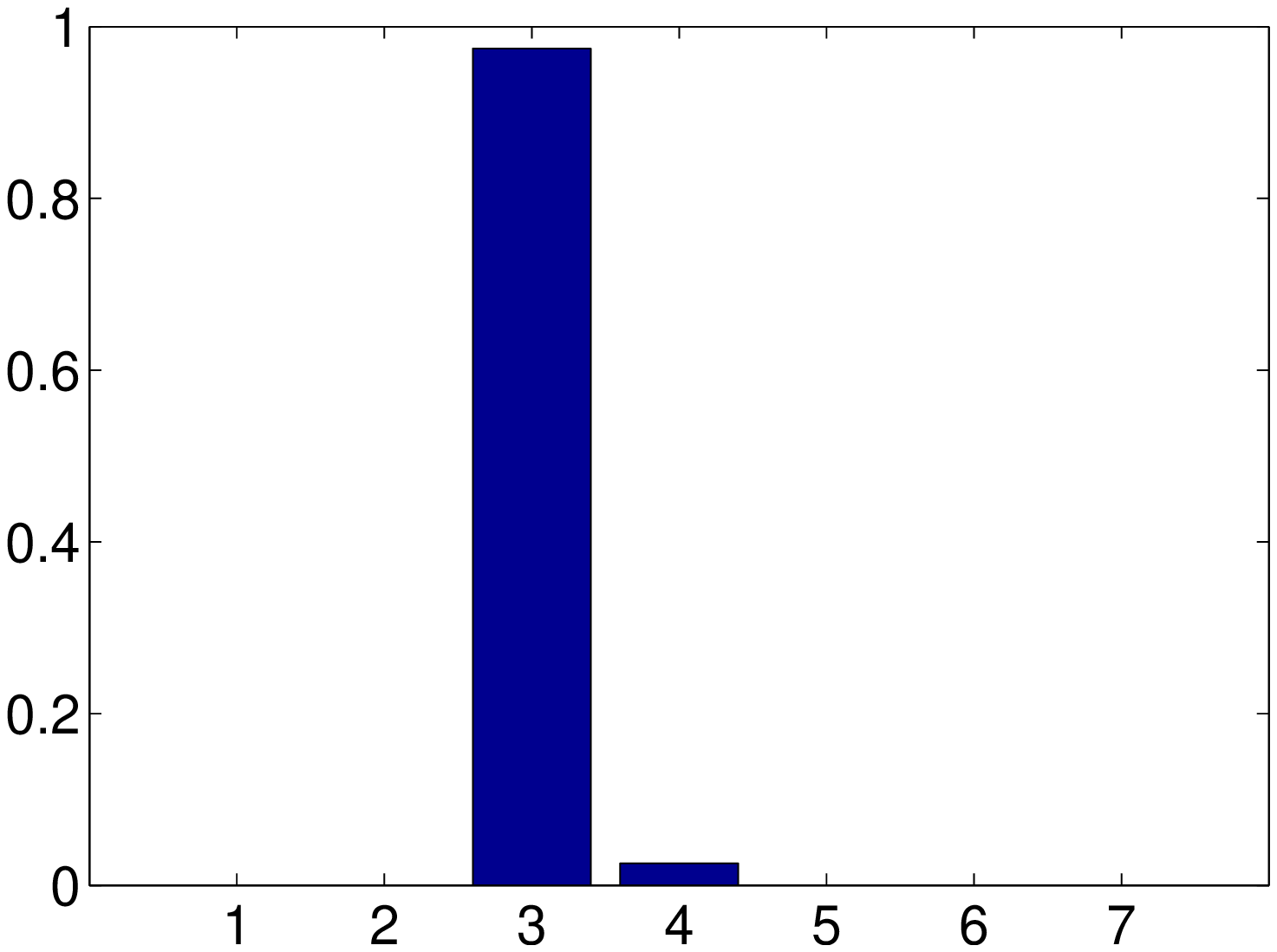}&
\includegraphics[height=1.0in]{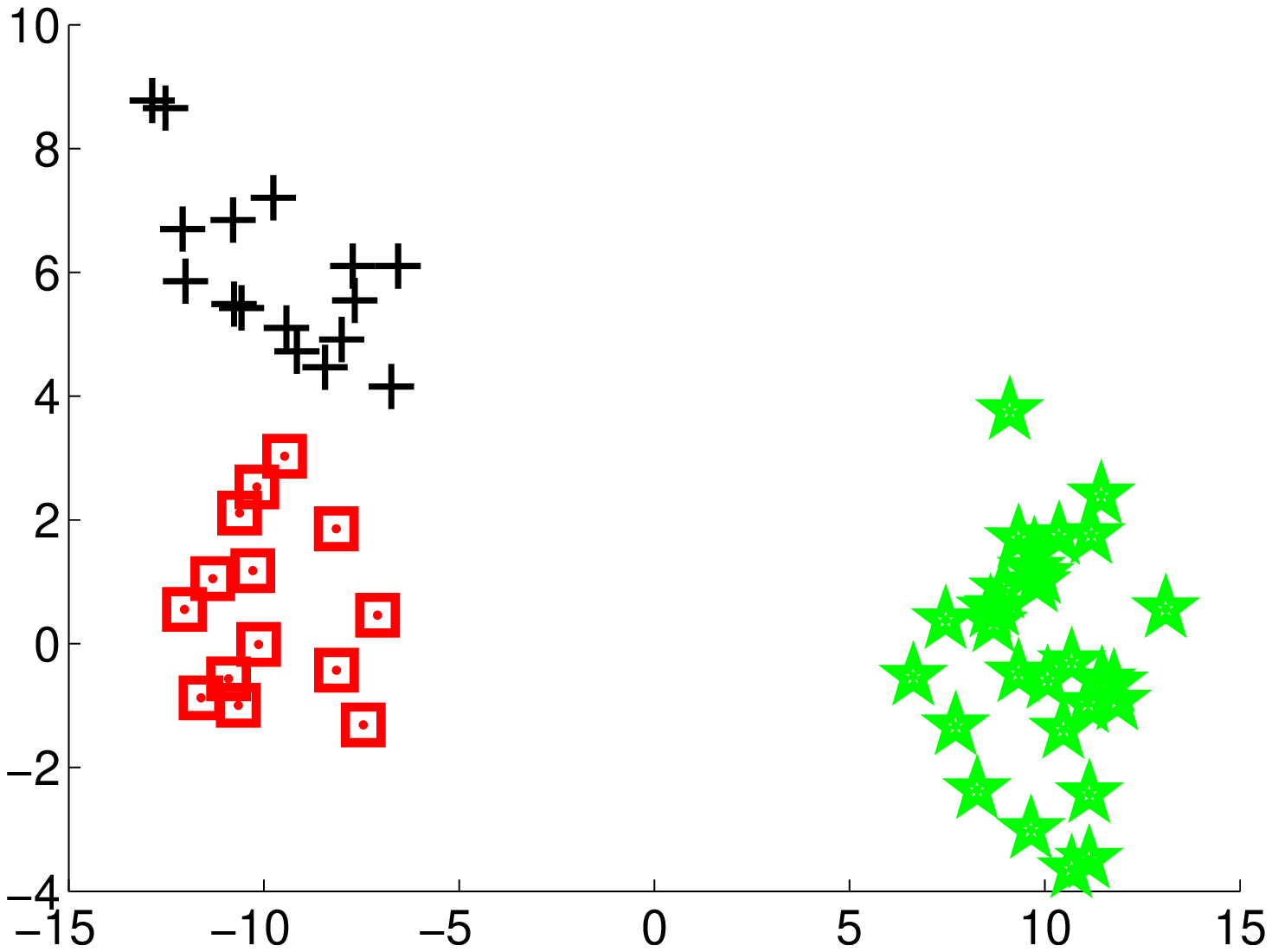}&
\includegraphics[height=1.0in]{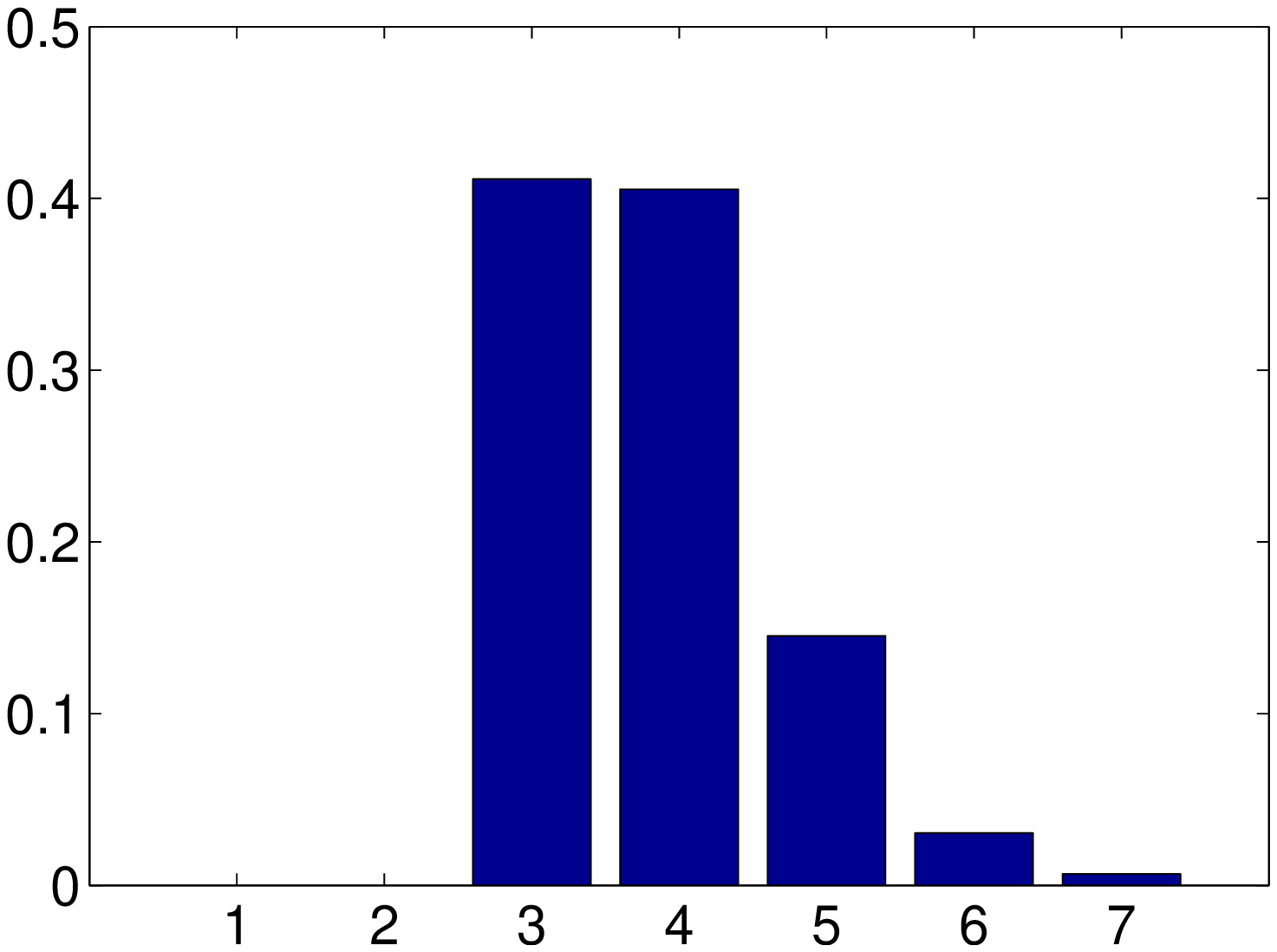}\\
\includegraphics[height=1.0in]{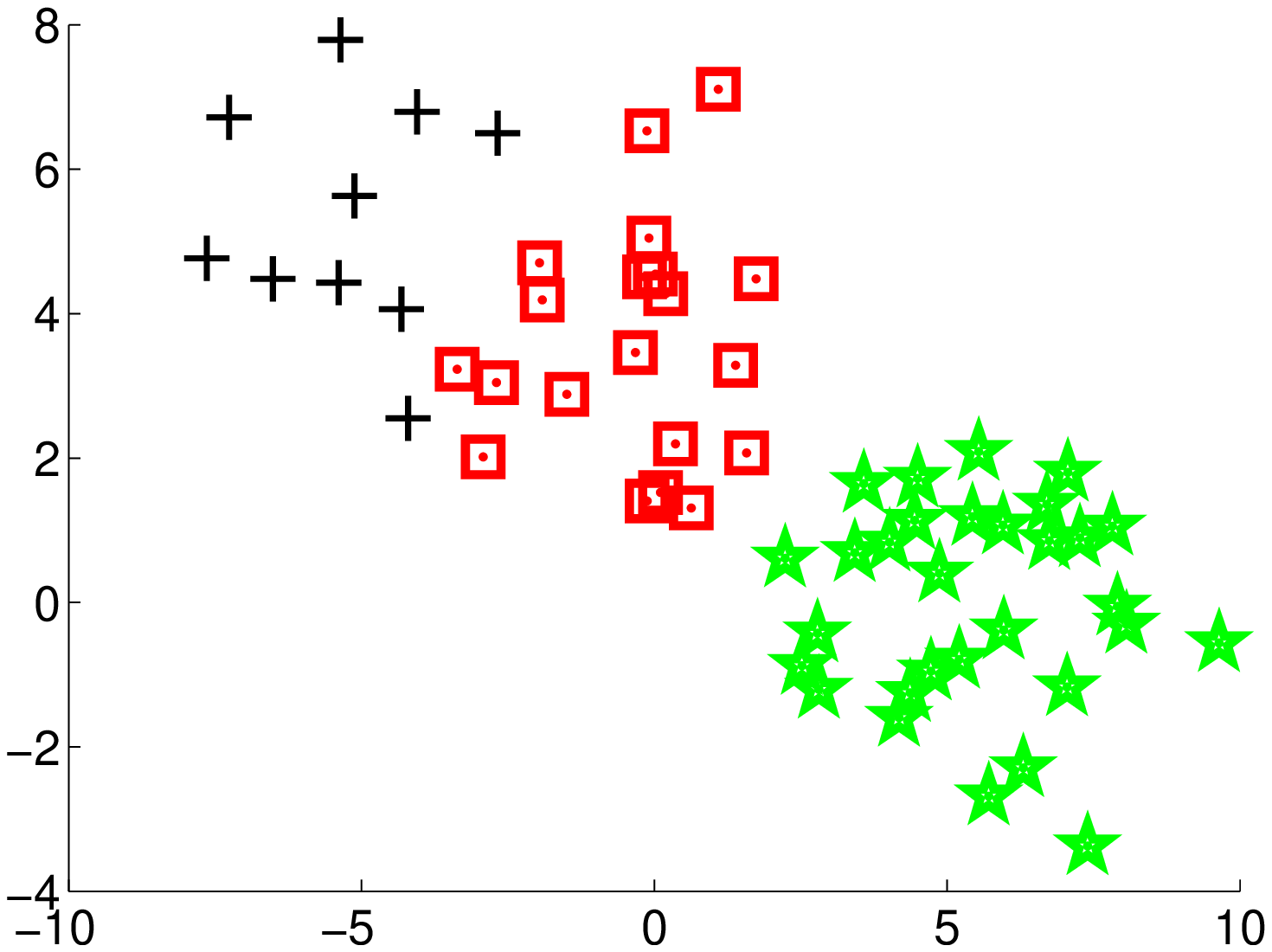}&
\includegraphics[height=1.0in]{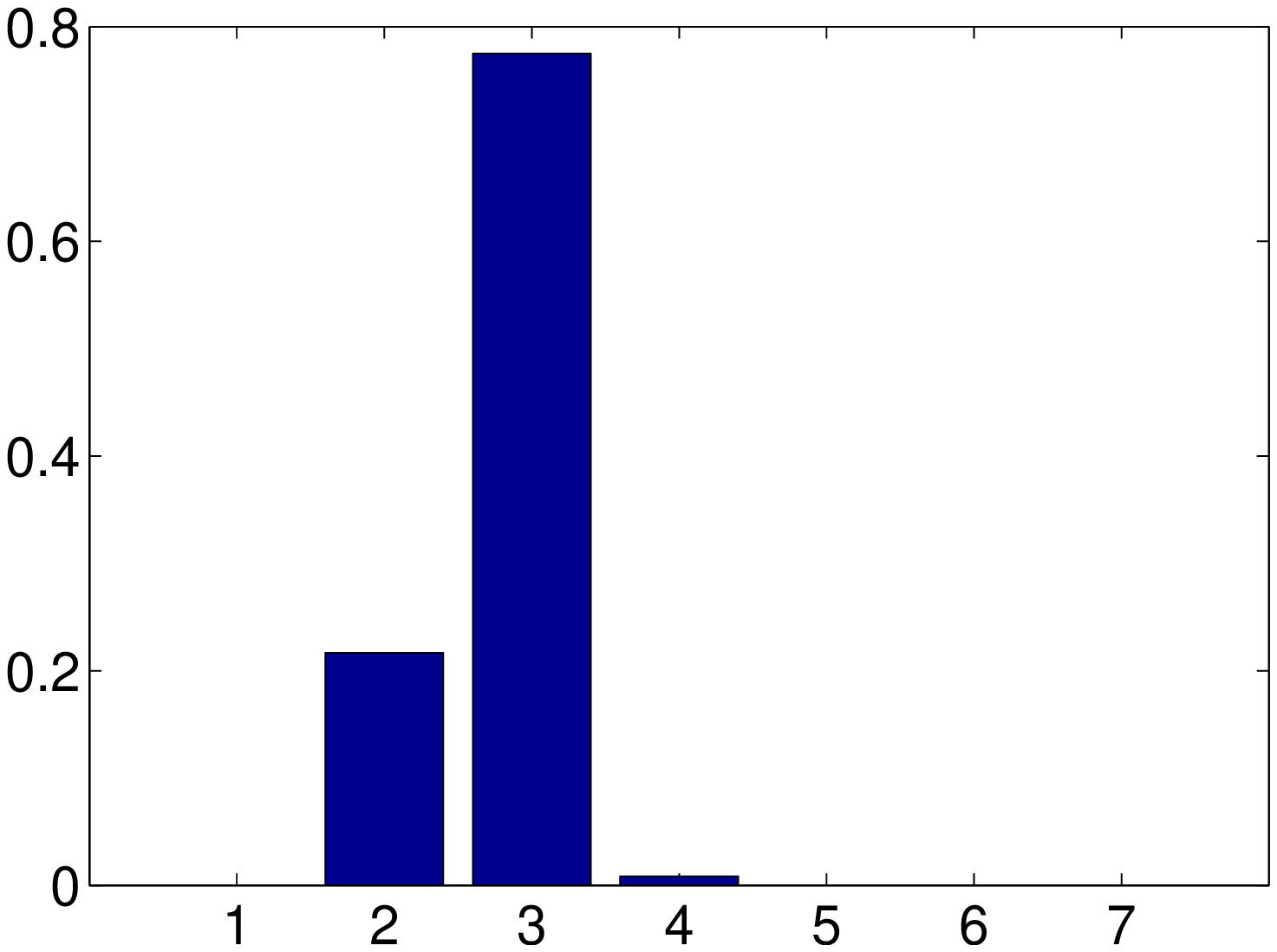}&
\includegraphics[height=1.0in]{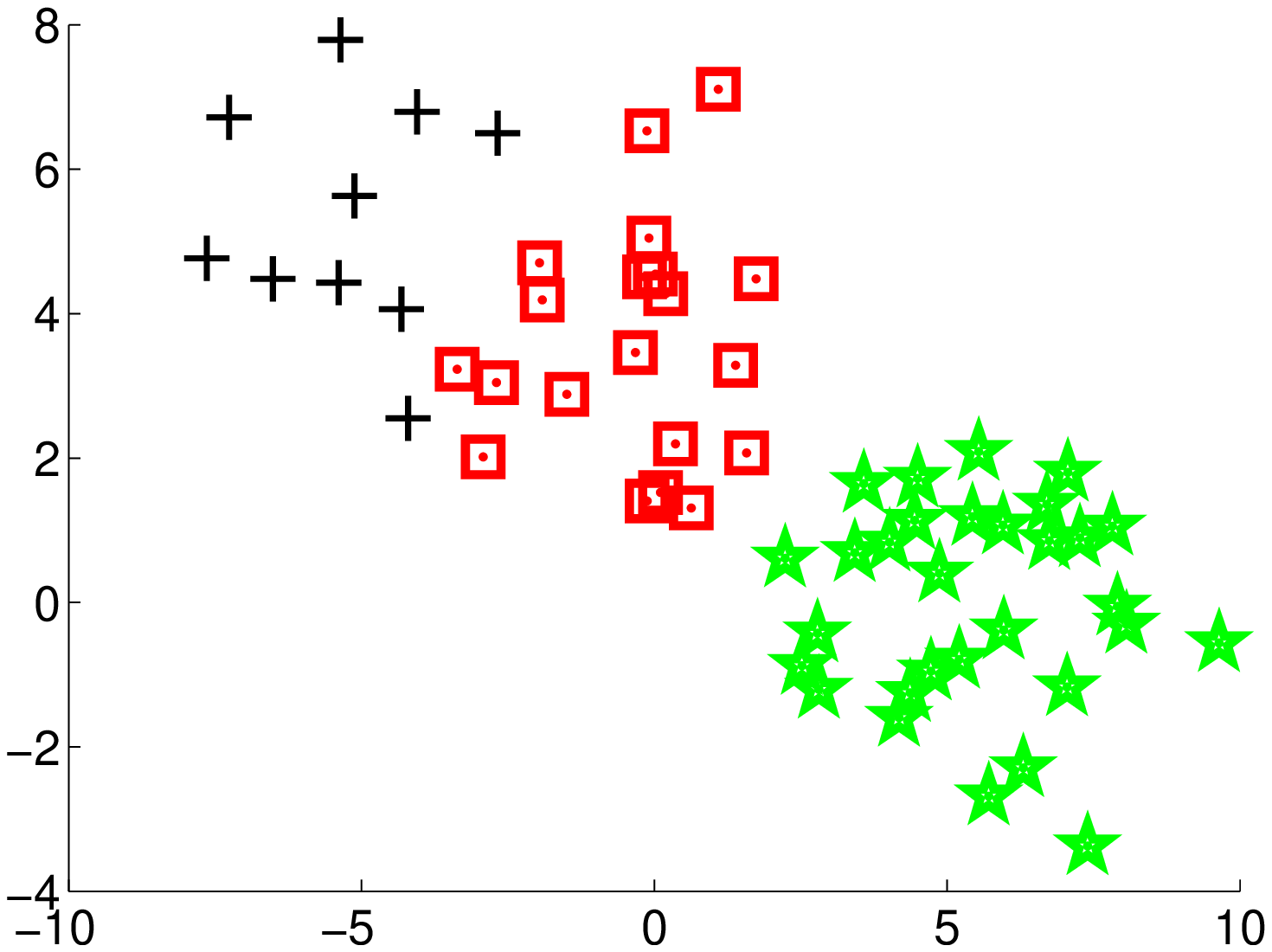}&
\includegraphics[height=1.0in]{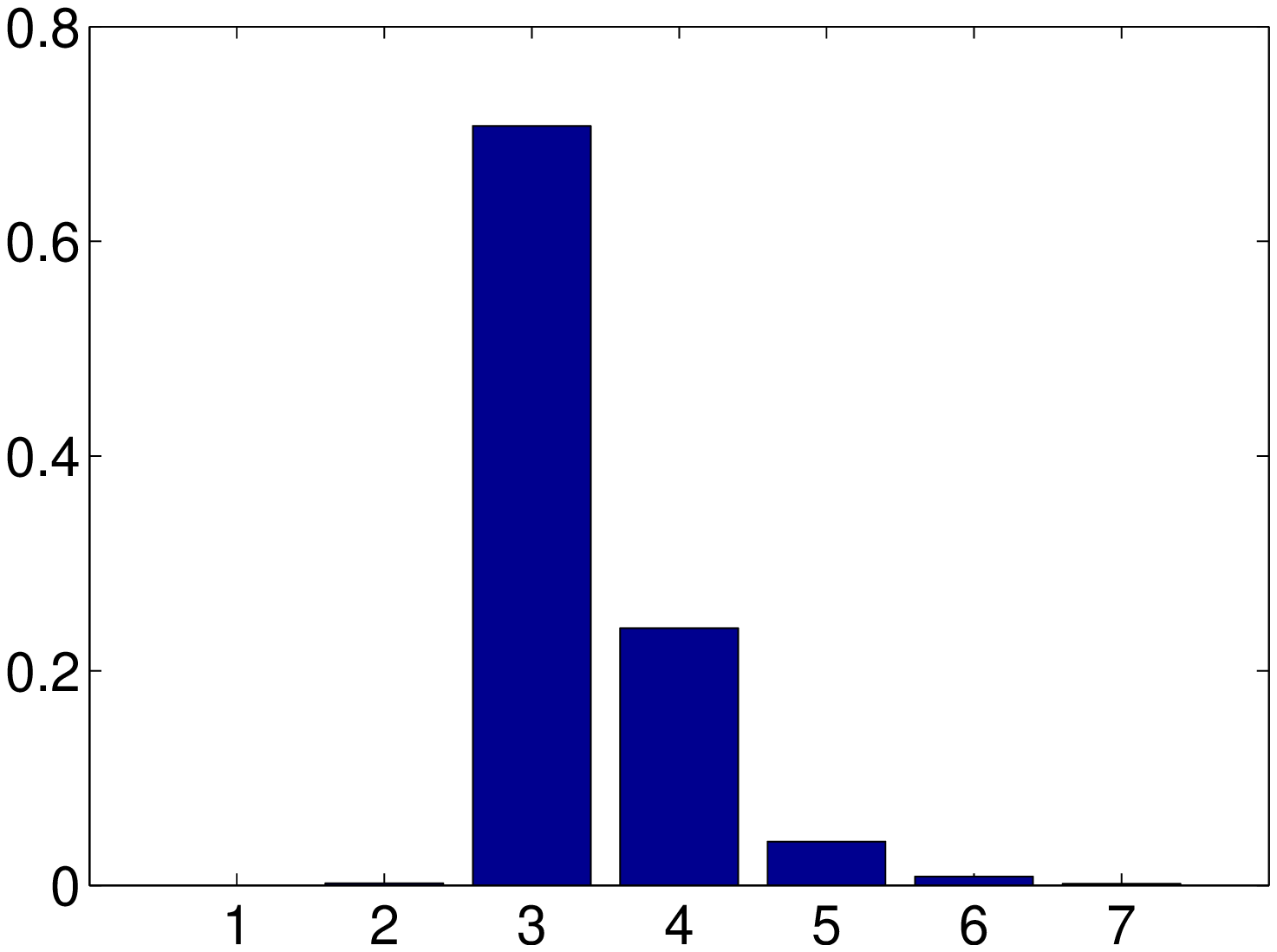}\\
 \multicolumn{2}{c||} {$\xi=0.2$, $\theta \in \{1,2,3,4,5\}\times 10^{3}$} &  \multicolumn{2}{c} {$\xi = 10$, $\theta \in \{1,2,3,4,5\}\times10^{3}$}\\
\hline
\hline
\end{tabular}
\caption{Euclidean data clustering results under different priors on $\theta$ and $\xi$.}
\label{fig:euclidpara}
\end{center}
\end{figure}

In the Euclidean case, the data are generated from a Gaussian mixture model $p(x\mid \lambda) = \sum_{i=1}^{k_0} w_i g(x \mid \mu_i,\Sigma_i), x\in \real^2$, with the mixing weights $w_i$, $i=1,2,3$ and the component-specific Gaussian densities  $g(x \mid \mu_i,\Sigma_i)$, $i=1,2,3$. 
In the first experiment, we perform a sensitivity  analysis to the choices of hyperpriors $\theta$ and $\xi$ in the W-CRP formulation.  Recall from Section \ref{sssec:hyper} that (1) $\theta$ controls the strength of association between elements within clusters; (2) $\xi$ is the concentration parameter for the CRP which controls the prior probability of producing new clusters. In this experiment, we test sensitivity to $\theta$ and $\xi$, keeping all the remaining parameters fixed ($d_{EB} = 2$, $r=3$, $s = 4$). Fig. \ref{fig:euclidpara} shows the clustering results on three different $2$D Euclidean datasets (each dataset contains $60$ data points) with different $\theta$'s and $\xi$'s. Observations in Dataset 1 are clearly separable into three classes; Dataset 2 contains observations that can either be clustered into two or three classes; and Dataset 3 contains observation that shows no clear congregation of observations.

The histograms in Fig. \ref{fig:euclidpara} show posterior distributions of cluster number $K$.  In the upper left panel, results are provided for $\xi = 0.2$ and $\theta \in \{ 0.1,0.2,0.3,0.4,0.5 \}$. In the upper right panel, we use  $\xi=10$ and the same $\theta \in \{ 0.1,0.2,0.3,0.4,0.5 \}$. Clearly, when $\xi = 10$, we have more clusters with small sizes, although the dominant clusters appear to be similar to the case of $\xi = 0.2$.  In the lower left  panel, we set  $\xi= 0.2$ and  $\theta \in \{ 1000,2000,3000,4000,5000 \}$. Comparing this with the result in the upper left panel, we can see that 
 larger $\theta$ values lead to more but tighter clusters. This conforms to our intuition about the role of  $\theta$.  
    In the lower right panel, we used large $\xi=10$ and large $\theta \in \{ 1000,2000,3000,4000,5000\}$. Comparing this with the one in upper right panel, we find that when $\theta$ is large, the estimate of the number of clusters is less sensitive to $\xi$ than with a smaller value of $\theta$.

\begin{figure}
\begin{center}
\begin{tabular}{c}
\includegraphics[height=2.6in]{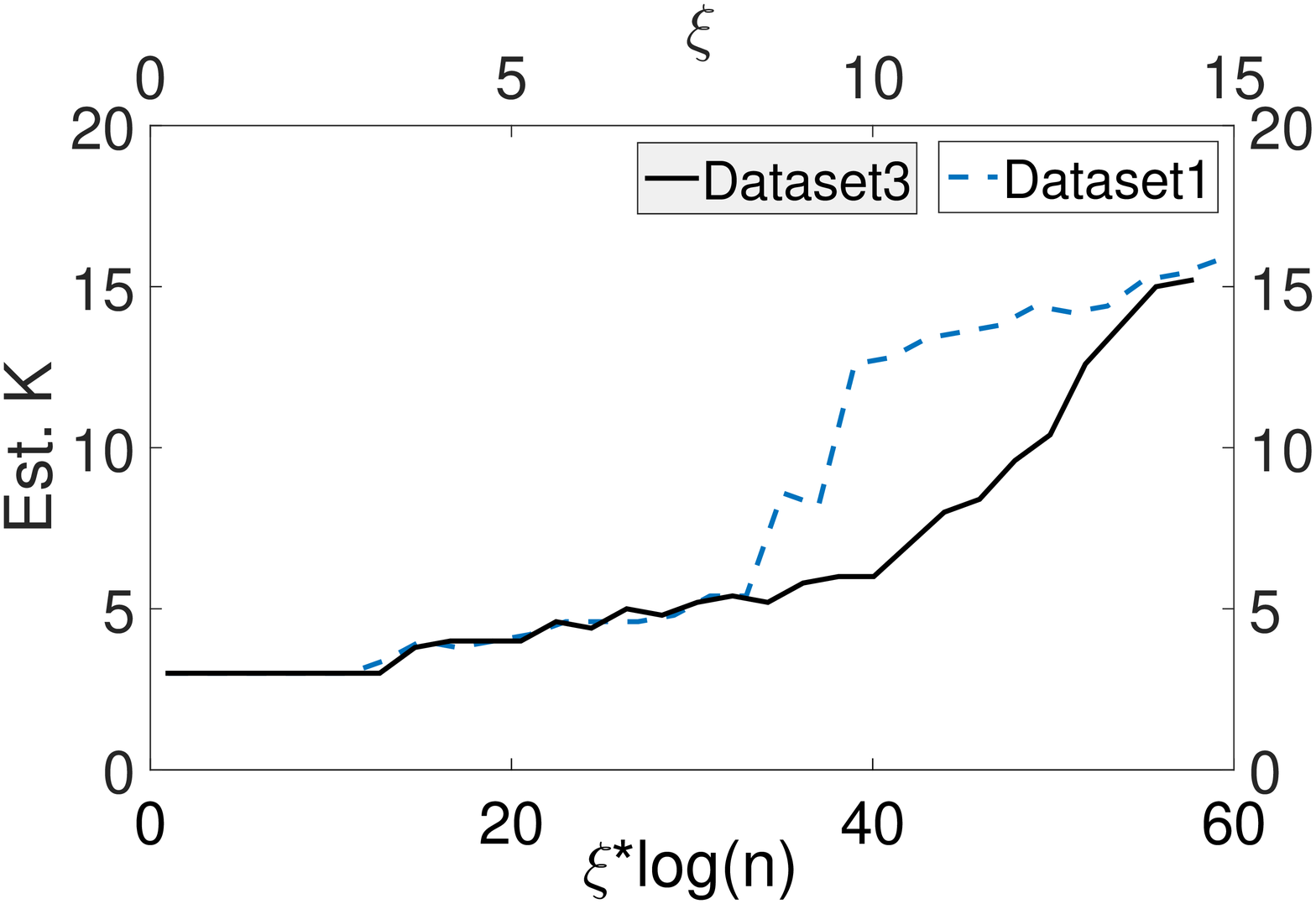}
\end{tabular}
\caption{Estimated number of clusters for different $\xi$ when we use a fix prior on $\theta$ ($ = \{ 100,200,300,400,500\}$).}
\label{fig:xiK}
\end{center}
\end{figure}

\begin{figure}
\begin{center}
\begin{tabular}{|cc|cc|}
\hline
\multicolumn{2}{|c|}{Dataset 1} & \multicolumn{2}{c|}{Dataset 3}\\
\includegraphics[height=0.6in]{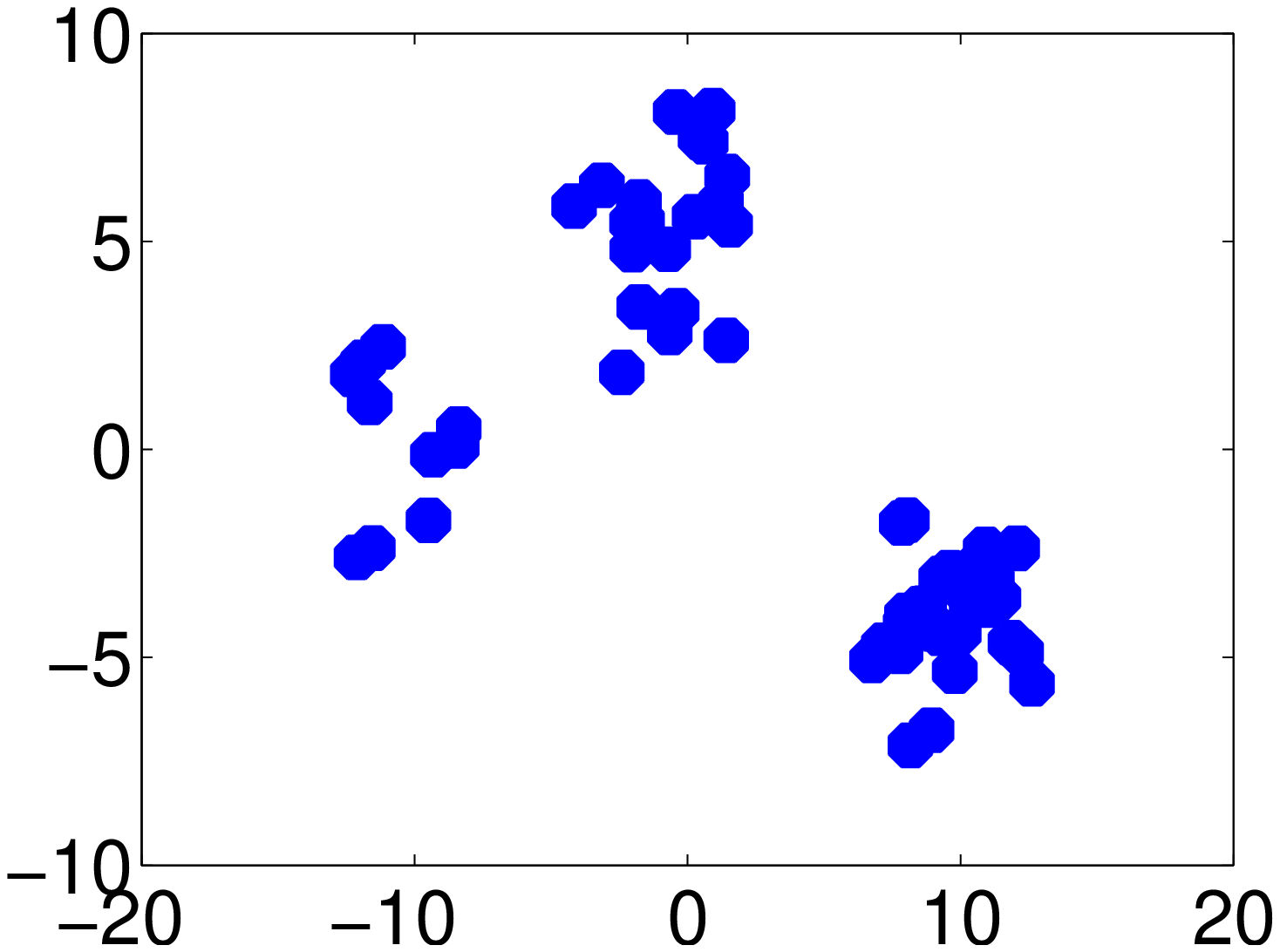}&
\includegraphics[height=1.7in]{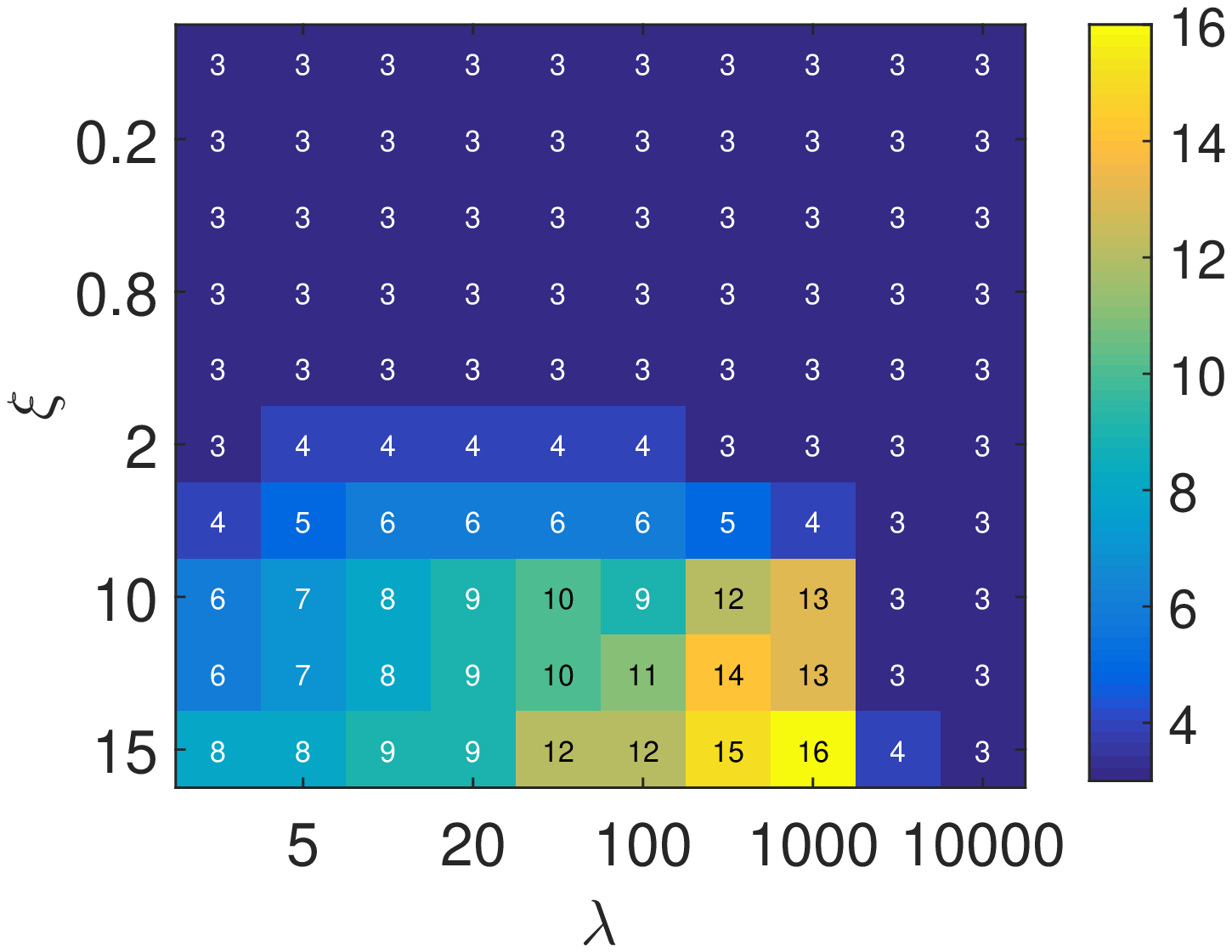}&
\includegraphics[height=0.6in]{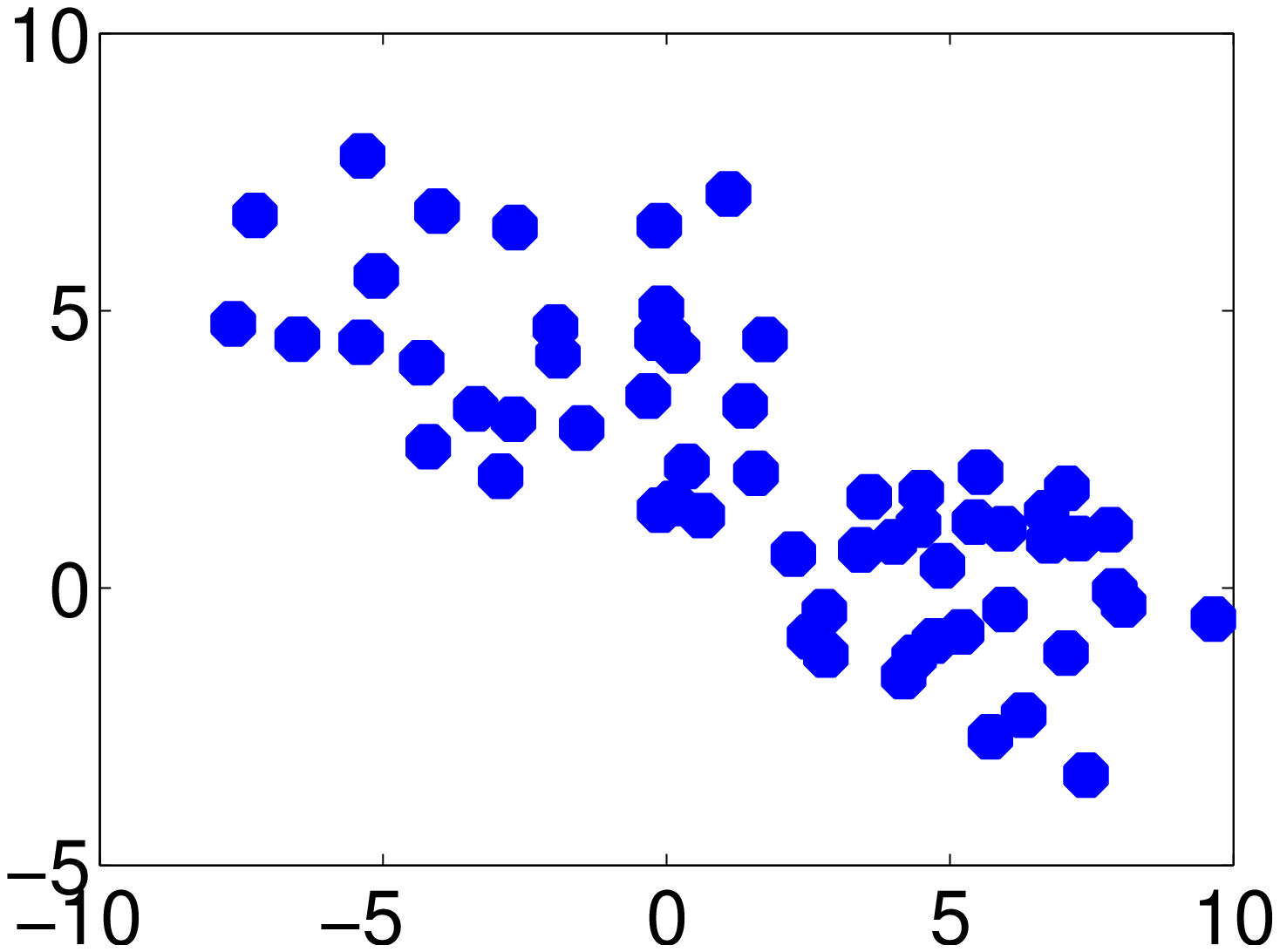}&
\includegraphics[height=1.7in]{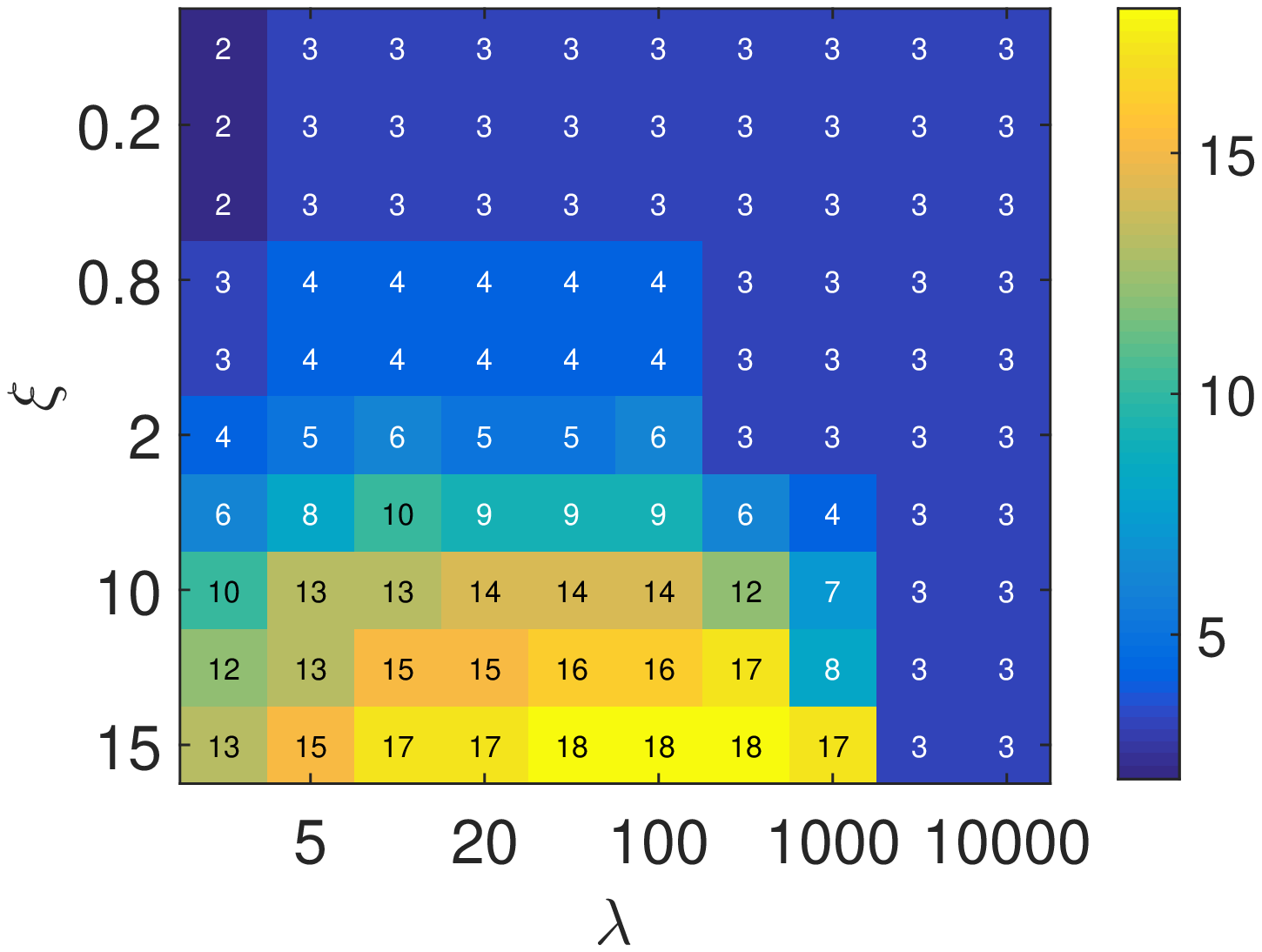}\\
\hline
\end{tabular}
\caption{Estimated number of inferred clusters for different $\lambda$ and $\xi$. The columns of the heat-map matrix are corresponding to $\lambda$ ($\lambda=\{1,5,10,20,50,100,200,500,1000,5000,10000\}$) and the rows are corresponding to $\xi $ ($\xi= \{0.1,0.2,0.4,0.8,1,2,5,10,12,15\}$). }
\label{fig:clusteringpara}
\end{center}
\end{figure}

The next experiment presents a detailed sensitivity analysis to the choices of $\xi$ and $\theta$. First, we fix the prior on $\theta$ ($ = \{ 100,200,300,400,500\}$) and investigate the relation between the inferred cluster number $K$ versus the choice of $\xi$.  Fig. \ref{fig:xiK} shows the result. The upper x-axis is $\xi$ and bottom x-axis corresponds to $\xi \times \log(n)$, where $n=60$ is the number of observations.
One can see that, the inferred cluster number is robust to the choice of $\xi$ in a large range. After $\xi$ exceeds certain value, the inferred $K$ increases. This is reasonable because a large $\xi$ (in CRP) produces a prior with large probability to induce new clusters, and it dominates the posterior inference.  Typically, one does not choose such a large $\xi$ in most practical situations, unless there is strong prior evidence.  
Next, we study the sensitivity of the relation between the inferred $K$ and $\xi$ to the choice of  $\theta$.  To address this, for $\xi \in \{0.1,0.2,0.4,0.8,1,2,5,10,12,15\} $, we specify priors on $\theta$ in the following way: $\theta = \lambda \times \{0.1,0.2,0.3,0.4,0.5\}$, where $\lambda = \{1,5,10,20,50,100,200,500,1000,5000,10000\}$. Fig. \ref{fig:clusteringpara} shows a heat-map of the matrix of the estimated $K$ for different values of $\xi$ (across the rows) and $\lambda$ (across the columns).  This figure shows  (1) The posterior of cluster number $K$ is robust to the choice of $\xi$ and $\theta$ in a large range. (2) For a fixed $\xi$, a small $\theta$ tends to produce more clusters and a large $\theta$ tends to produce less clusters. These results are coherent with the roles of $\xi$ and $\theta$:   (1) $\xi$ controls the probability of introducing new clusters. Hence, for a very large $\xi$, the posterior tends  to have more smaller extraneous clusters. We recommend choosing  $\xi = \tilde{K}/(\log(n))$, $\tilde{K}$ is a preliminary estimate of the number of clusters. Such a choice does not overestimate the value of $\xi$ and produces less extraneous clusters. (2) $\theta$ controls the strength of association between elements within clusters.  A large $\theta$ indicates a strong association, and tends to have more and tighter classes. However, $\theta$ tends to have a weak influence on the clustering result if $\xi$ is not very large.


\begin{figure}
\begin{center}
\begin{tabular}{cc||cc}
\hline
\hline
 \multicolumn{2}{c}{ Small $\theta$} &  \multicolumn{2}{c}{Big $\theta$}\\
\includegraphics[height=1.0in]{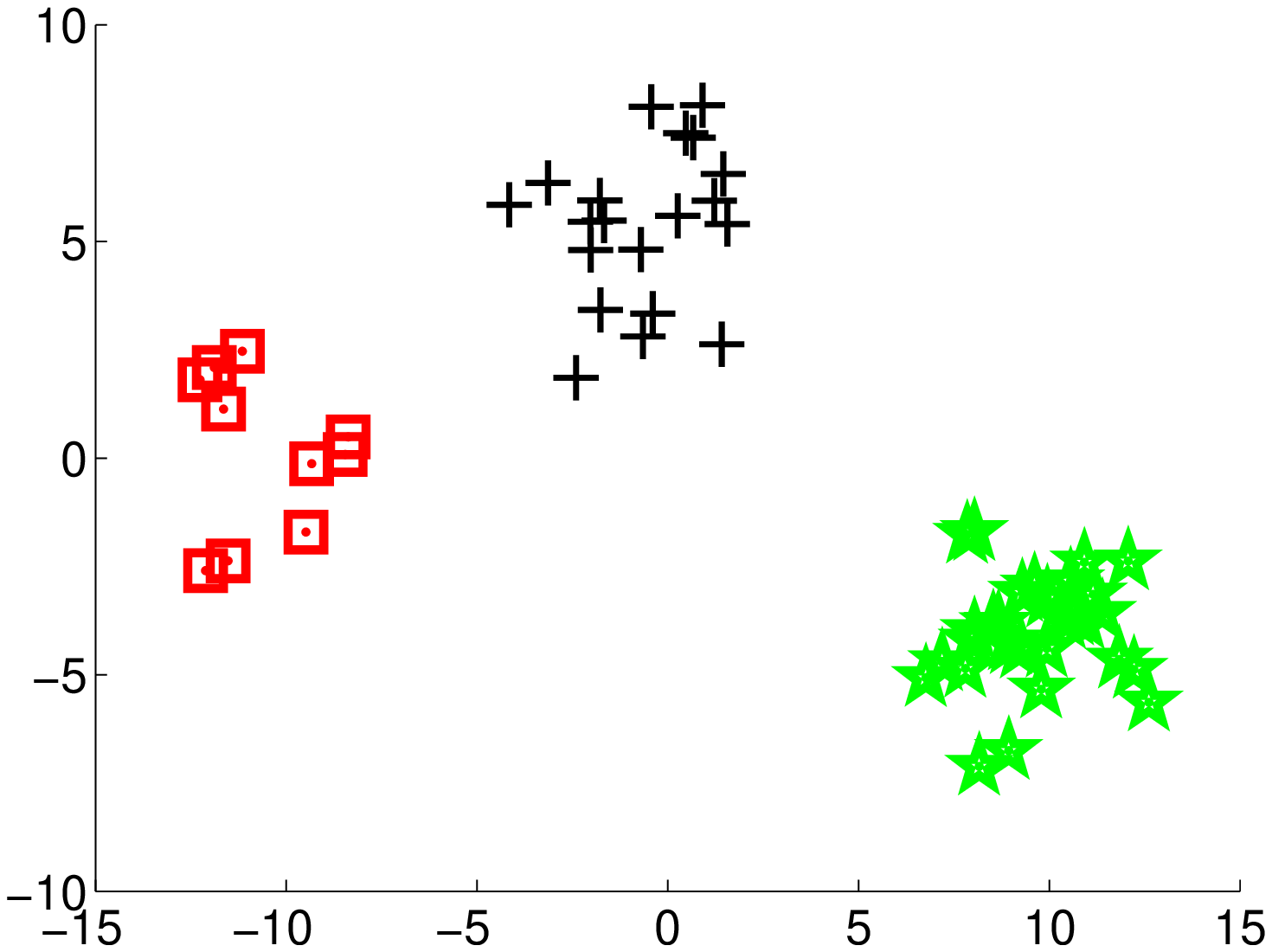}&
\includegraphics[height=1.0in]{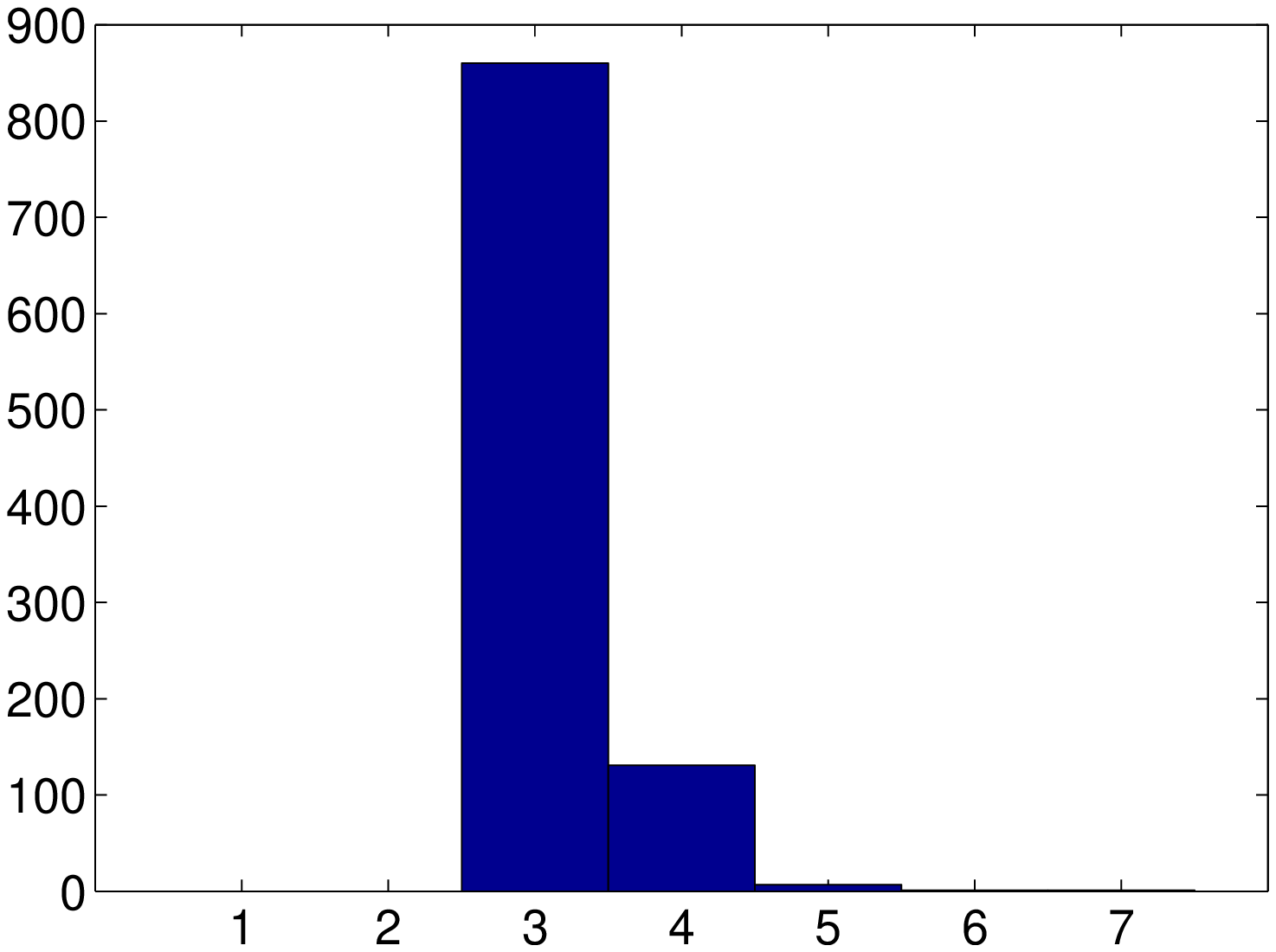}&
\includegraphics[height=1.0in]{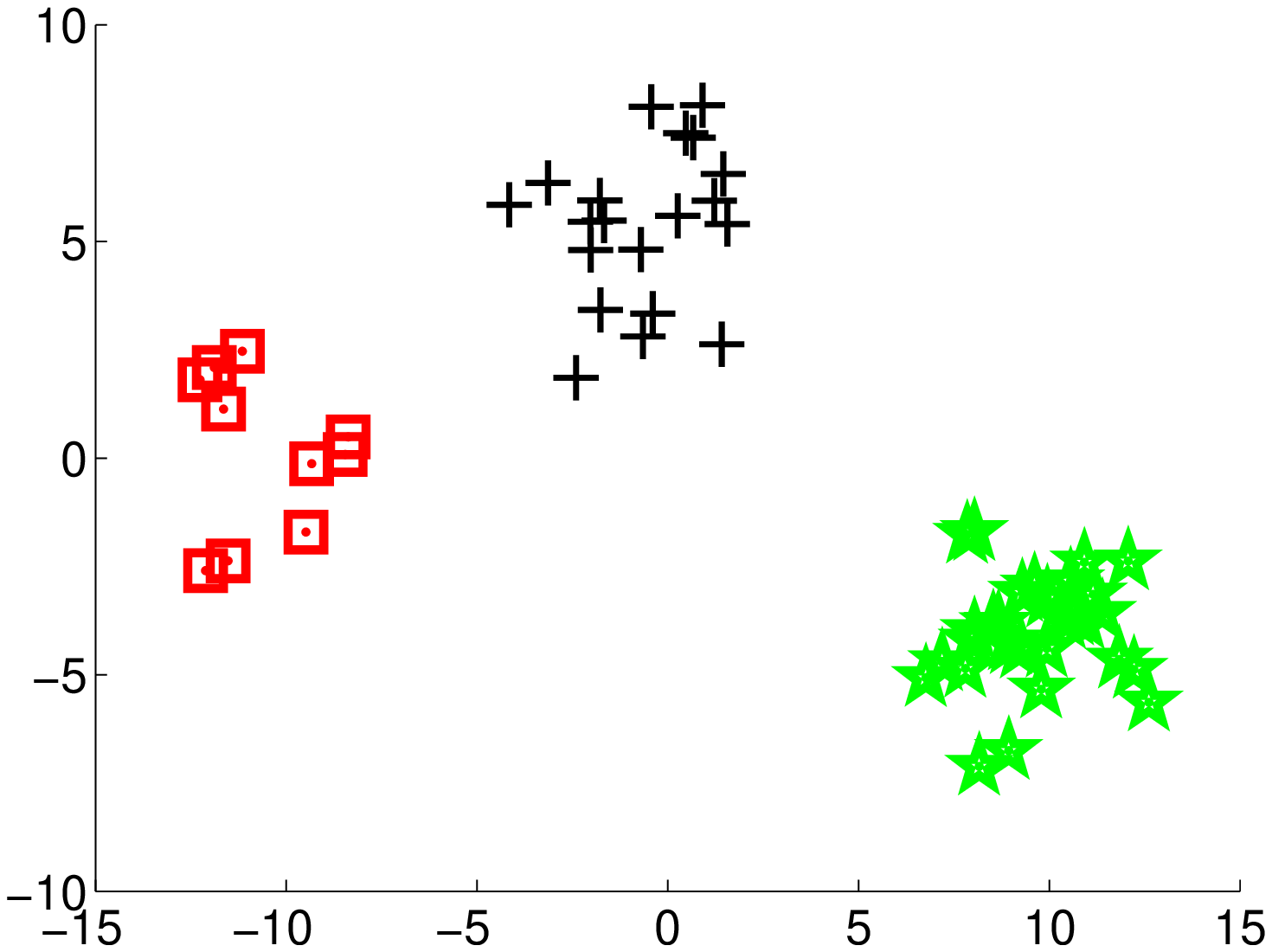}&
\includegraphics[height=1.0in]{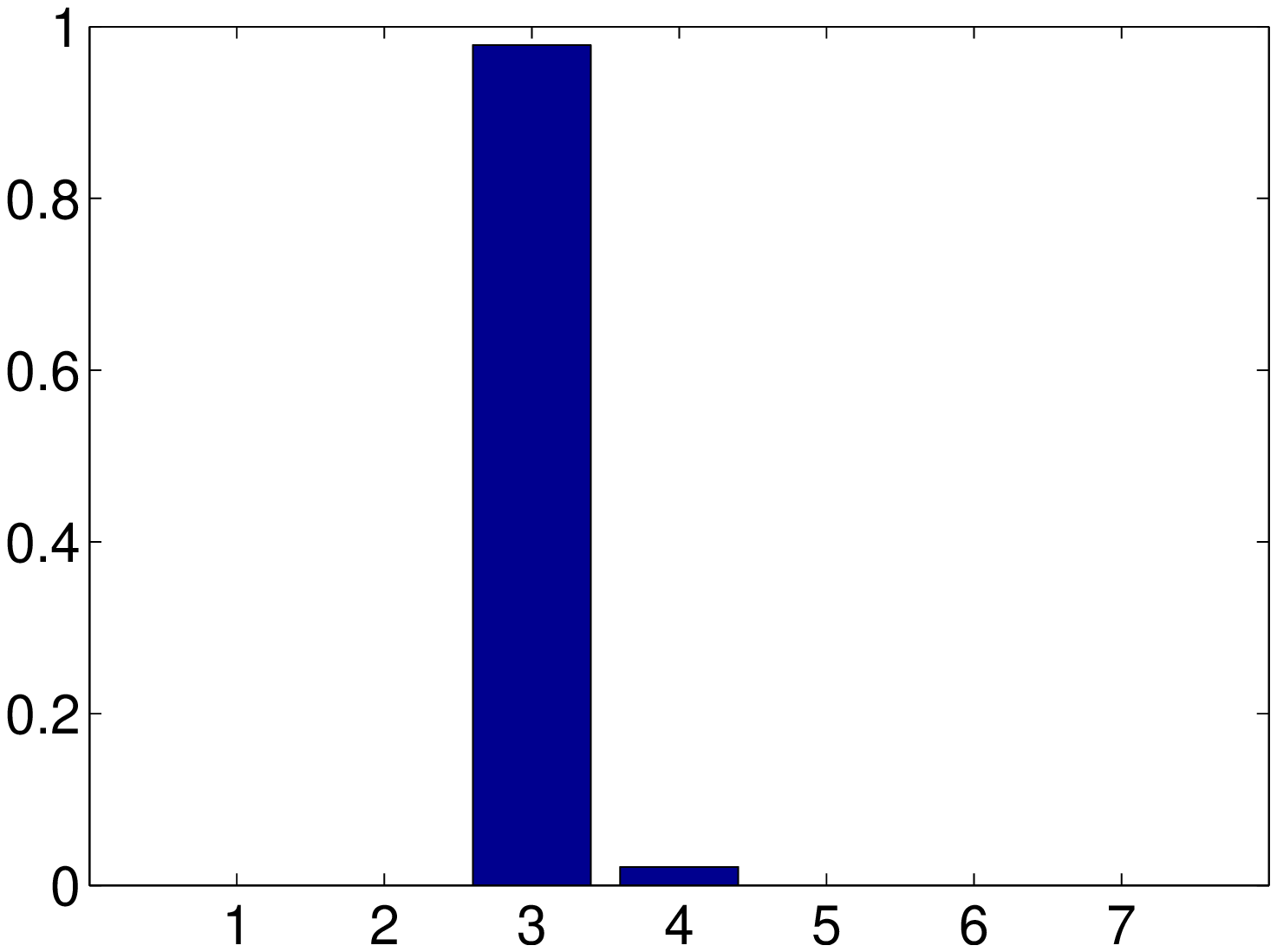}\\
\includegraphics[height=1.0in]{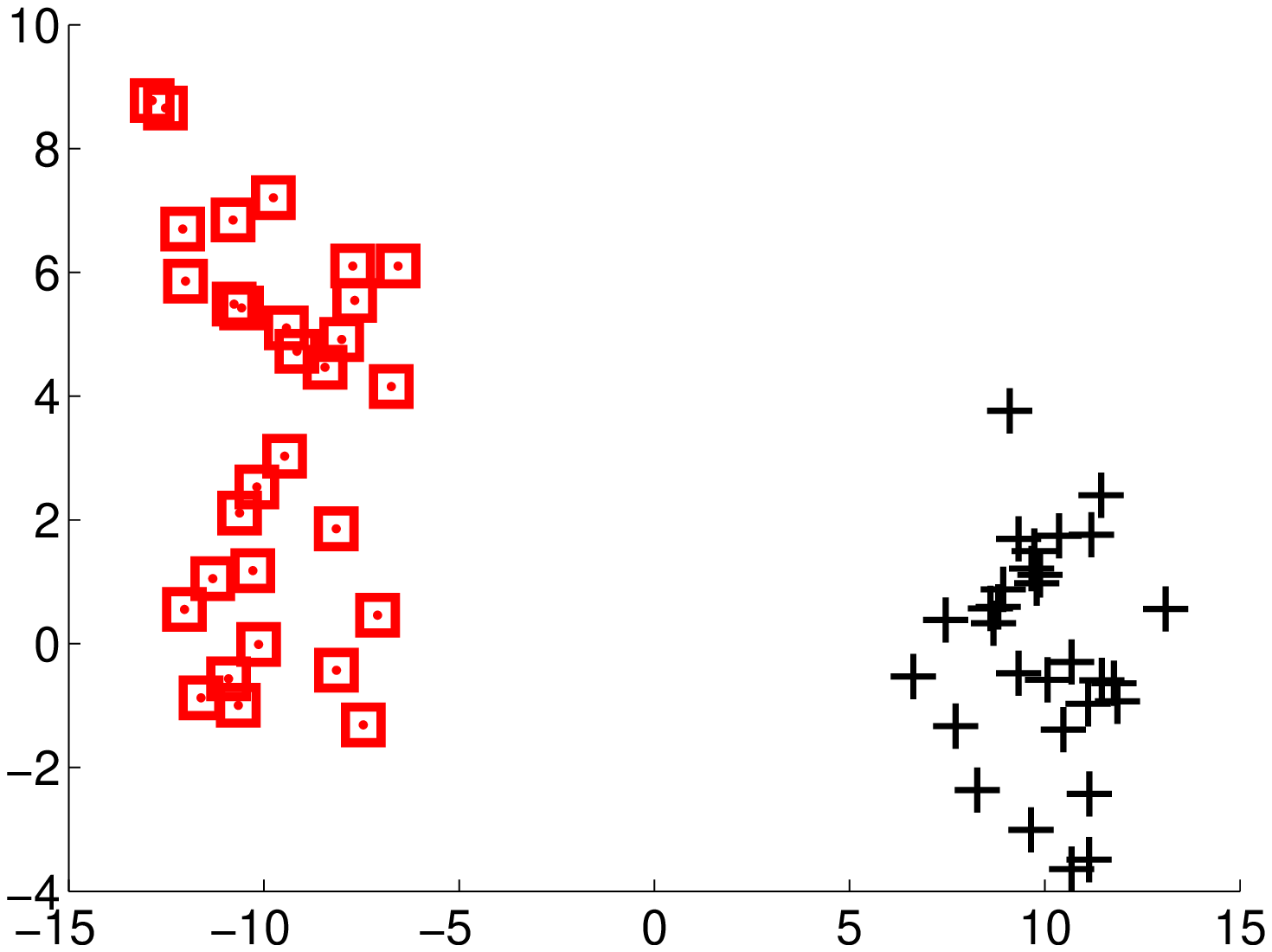}&
\includegraphics[height=1.0in]{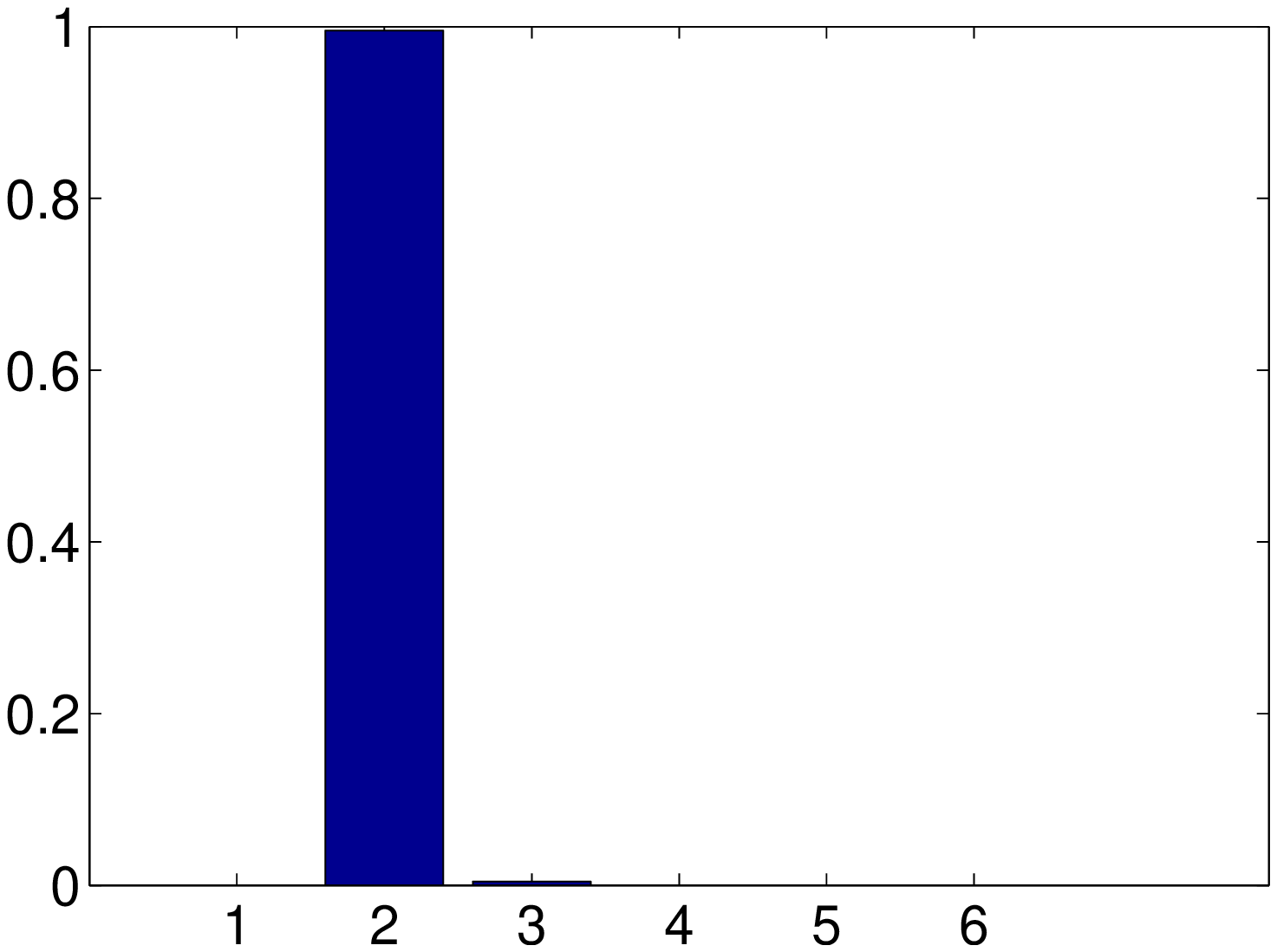}&
\includegraphics[height=1.0in]{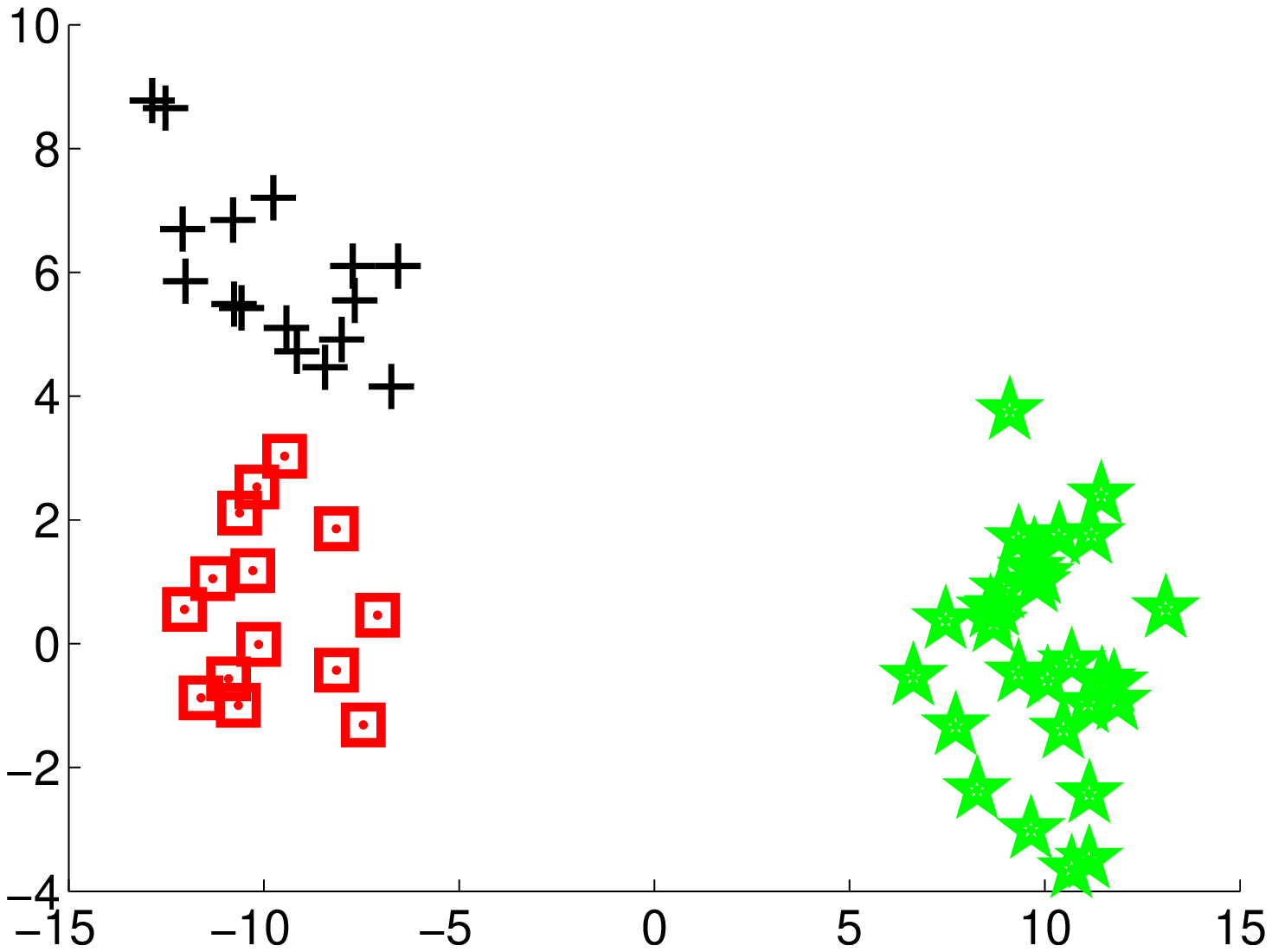}&
\includegraphics[height=1.0in]{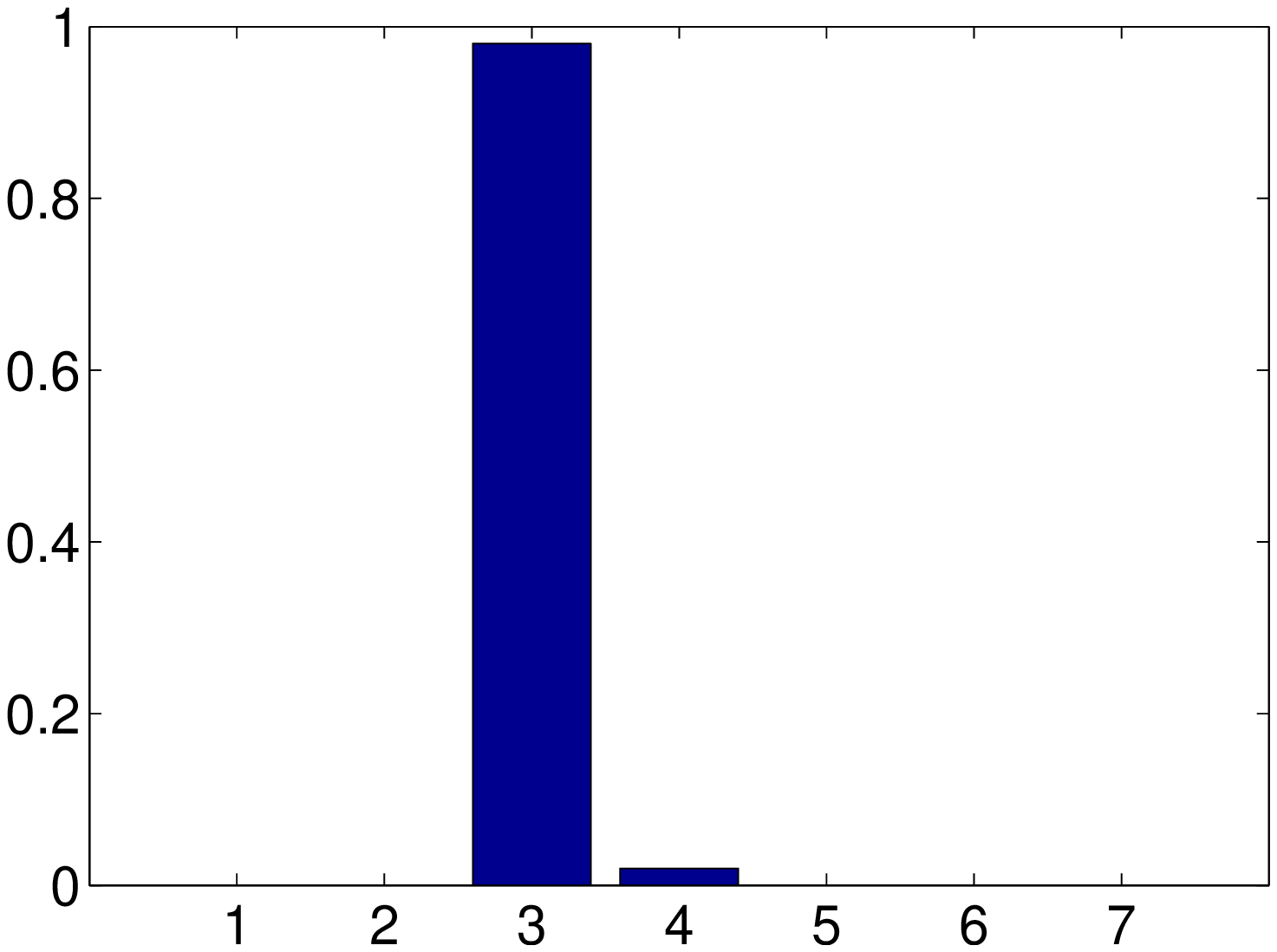}\\
\includegraphics[height=1.0in]{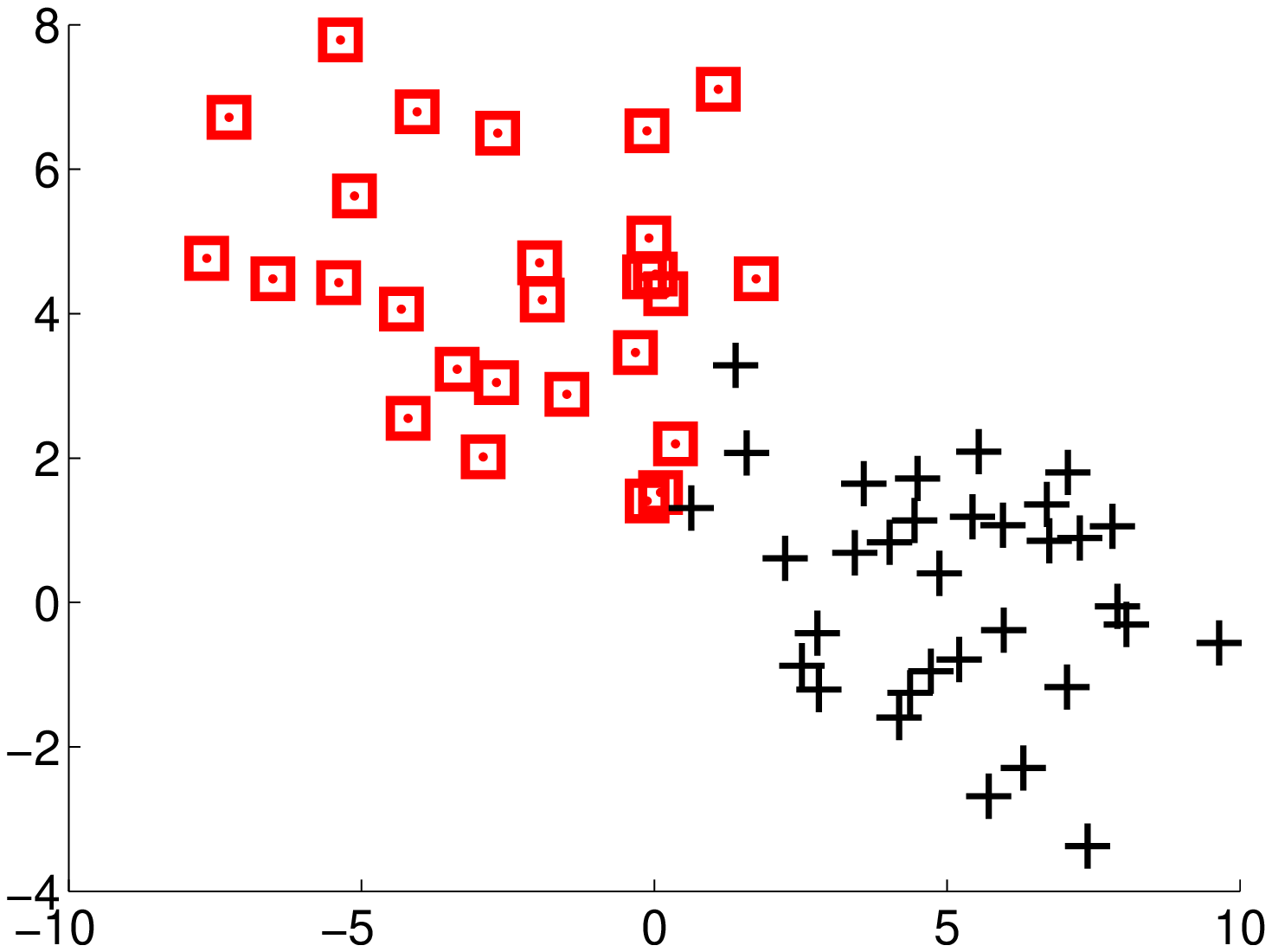}&
\includegraphics[height=1.0in]{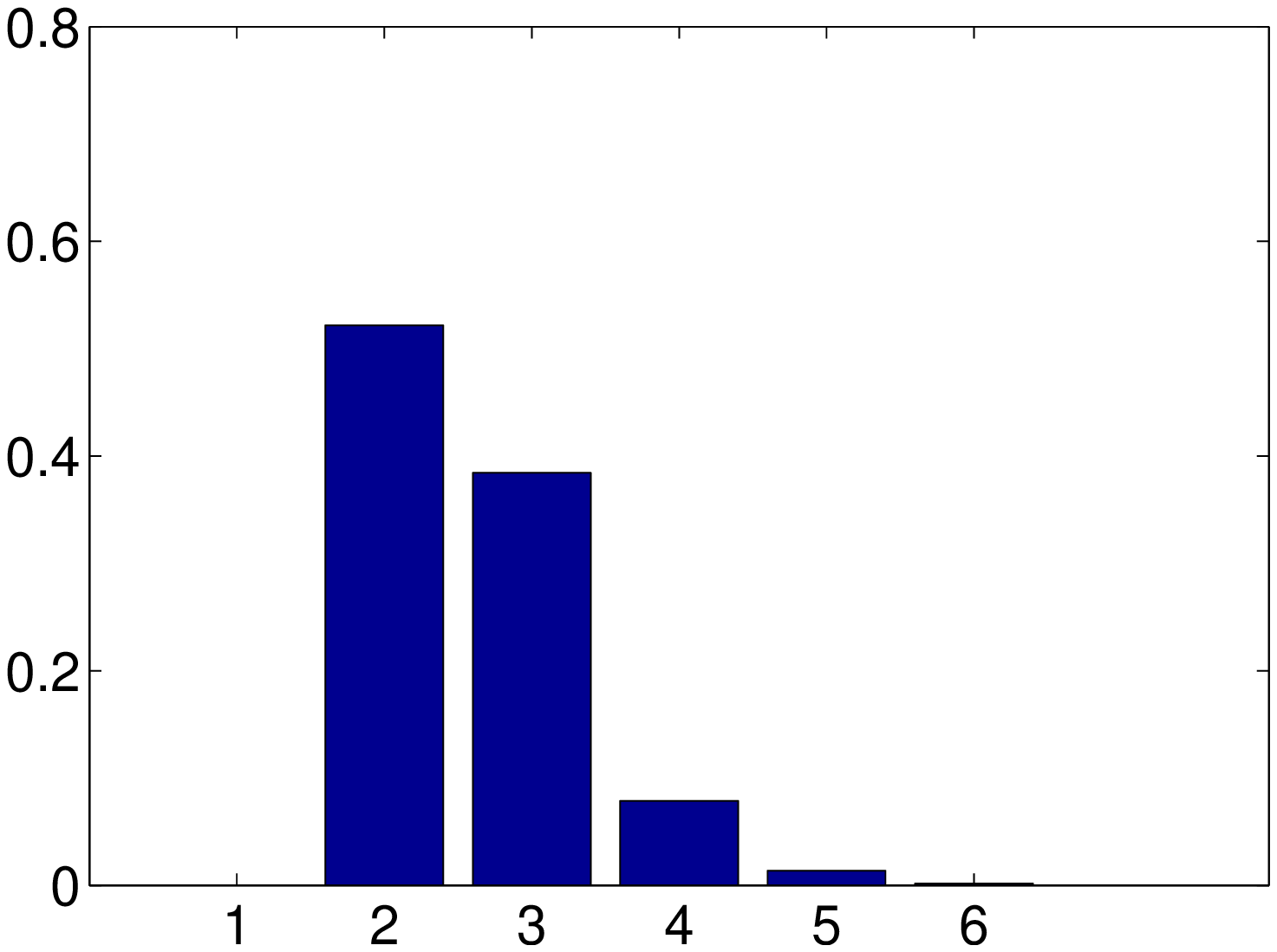}&
\includegraphics[height=1.0in]{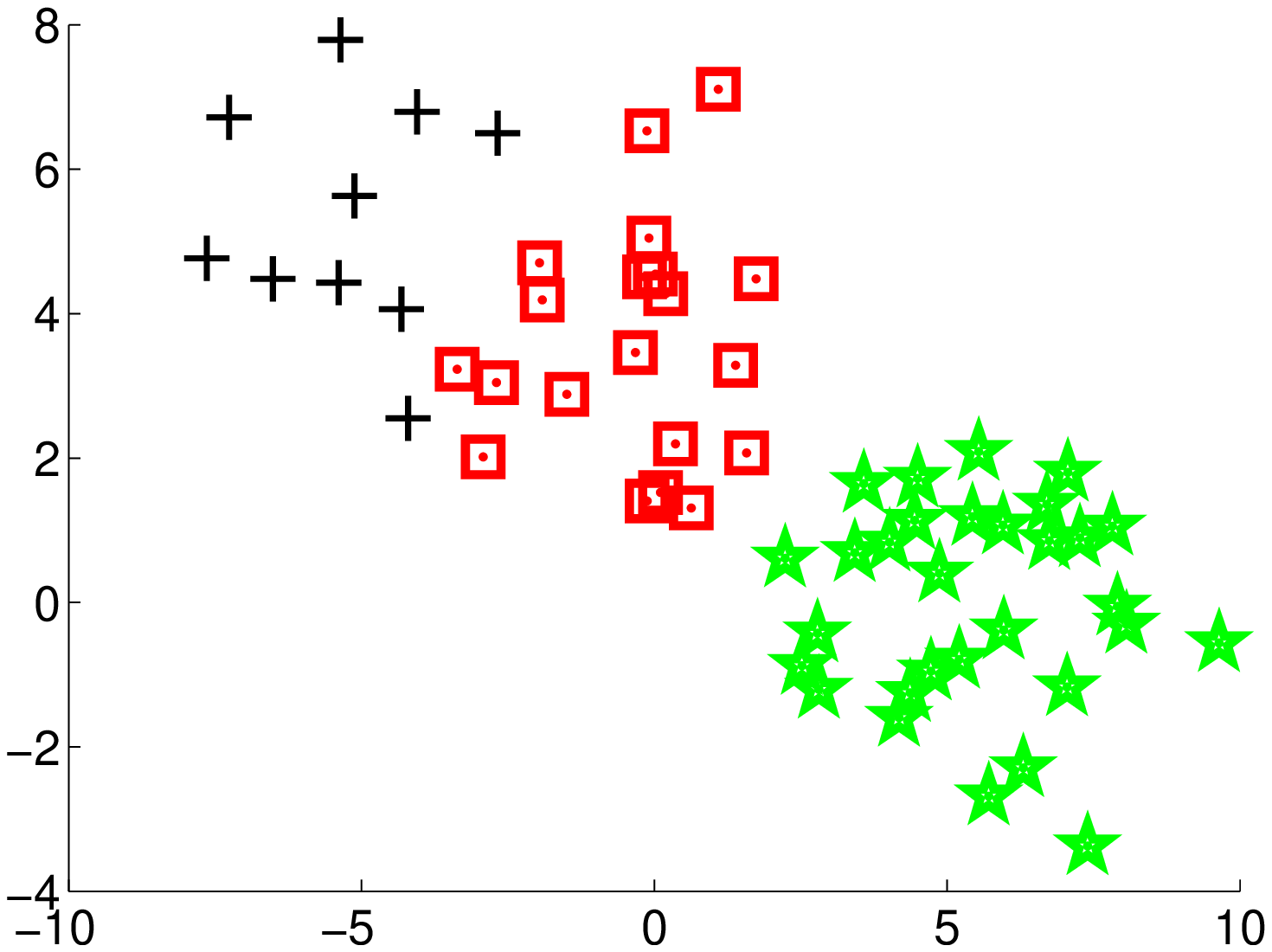}&
\includegraphics[height=1.0in]{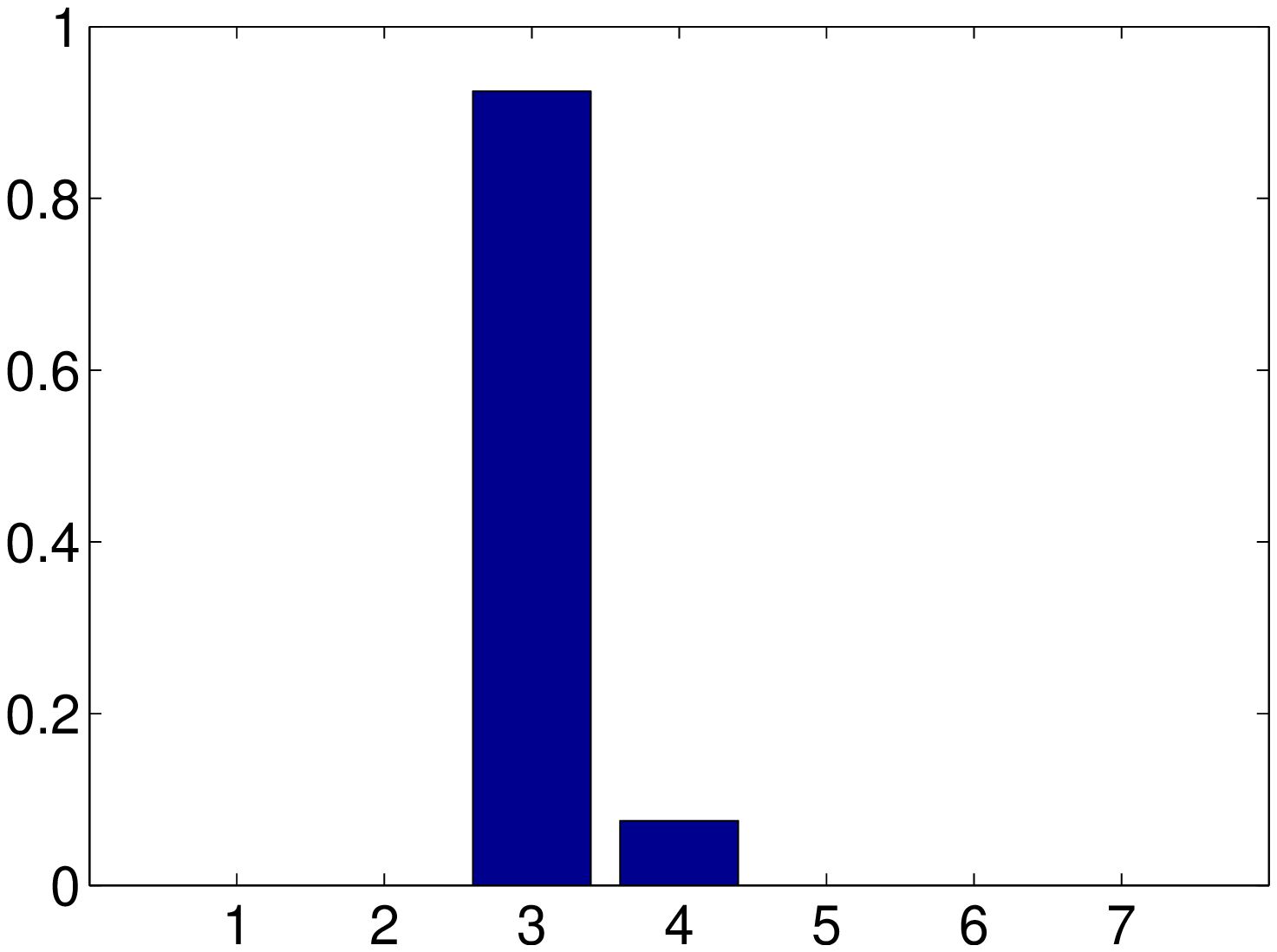}\\
\hline
\hline
\end{tabular}
\caption{Euclidean data clustering results using our model with MFM prior.  We use $\lambda=2$ and different $\theta$ in this experiment. The left part results are using $\theta \in \{ 0.1,0.2,0.3,0.4,0.5 \}$, and the right part results are using $\theta \in \{  1000,2000,3000,4000,5000 \}$.}
\label{fig:euclidmfm}
\end{center}
\end{figure}

 In a very recent technical report \citep{miller2013inconsistency},  it is shown that  Dirichlet process mixture (DPM) of Gaussians leads to an inconsistent estimation of the number of clusters.  Instead, mixture of finite mixtures model  (MFM) \citep{miller2013inconsistency} can guarantee the consistent estimation of the number of clusters. In next experiment,  we compare with mixture of finite mixtures (MFM) prior on the same data as in Fig. \ref{fig:euclidpara}. For a given finite mixture of Gaussians with number of components $K$, the MFM model assumes a Dirichlet distribution, $\mbox{Dir}(\alpha, \ldots, \alpha)$ conditionally on $K$ which  assigned a $\mbox{Poisson}(\lambda)$ distribution.  Fig. \ref{fig:euclidmfm} shows the clustering results with MFM prior, where we set  $\lambda = 2$. The left panel shows results using $\theta \in \{ 0.1,0.2,0.3,0.4,0.5 \}$ and the right panel shows results using $\theta \in \{  1000,2000,3000,4000,5000 \}$.  The clustering results of CRP and MFM are very similar.  

\begin{figure}
\begin{center}
\begin{tabular}{cc||cc}
\hline
\hline
 \multicolumn{2}{c}{ DPM of Gaussians} &  \multicolumn{2}{c}{MFM of Gaussians}\\
\includegraphics[height=1in]{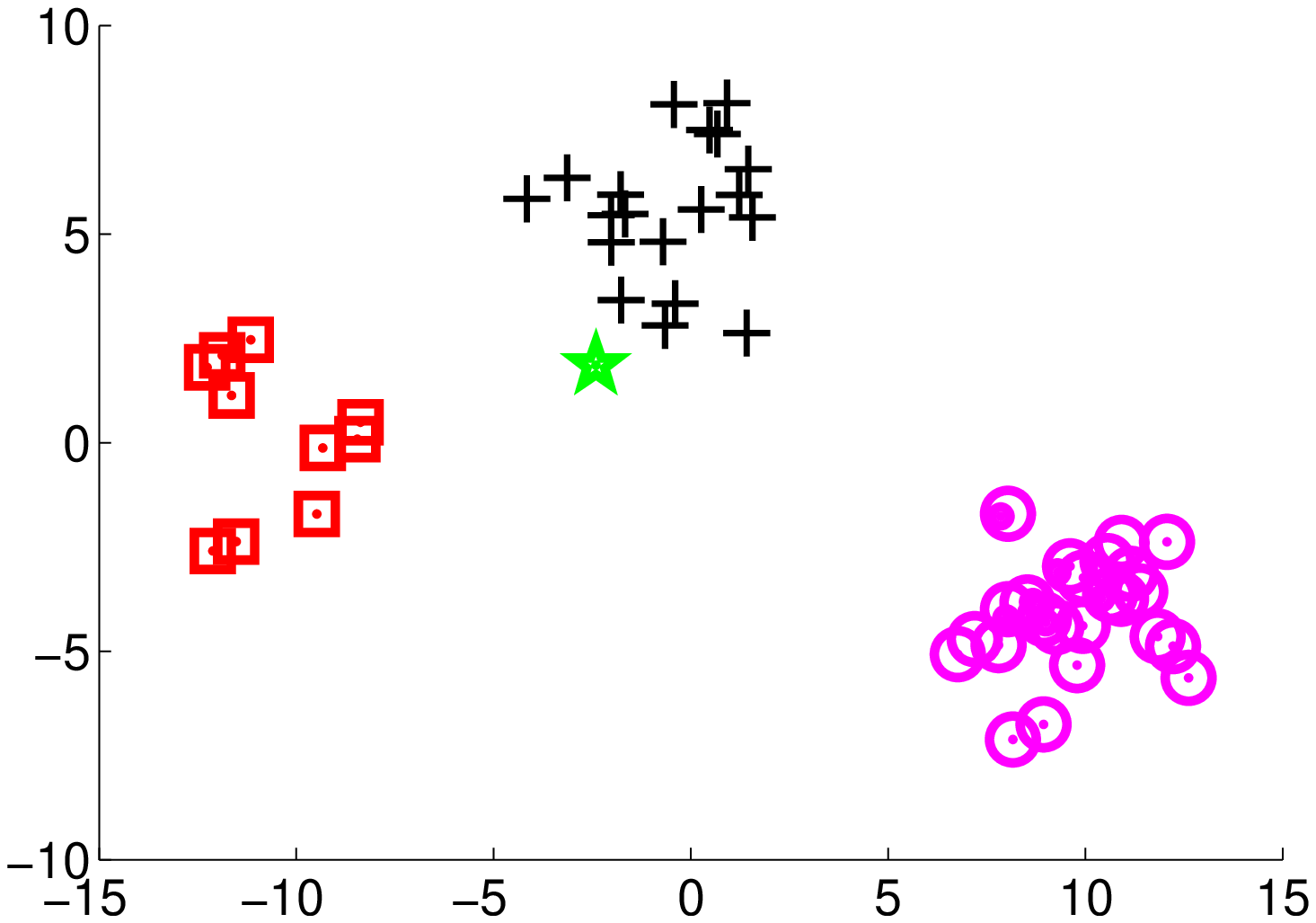}&
\includegraphics[height=1in]{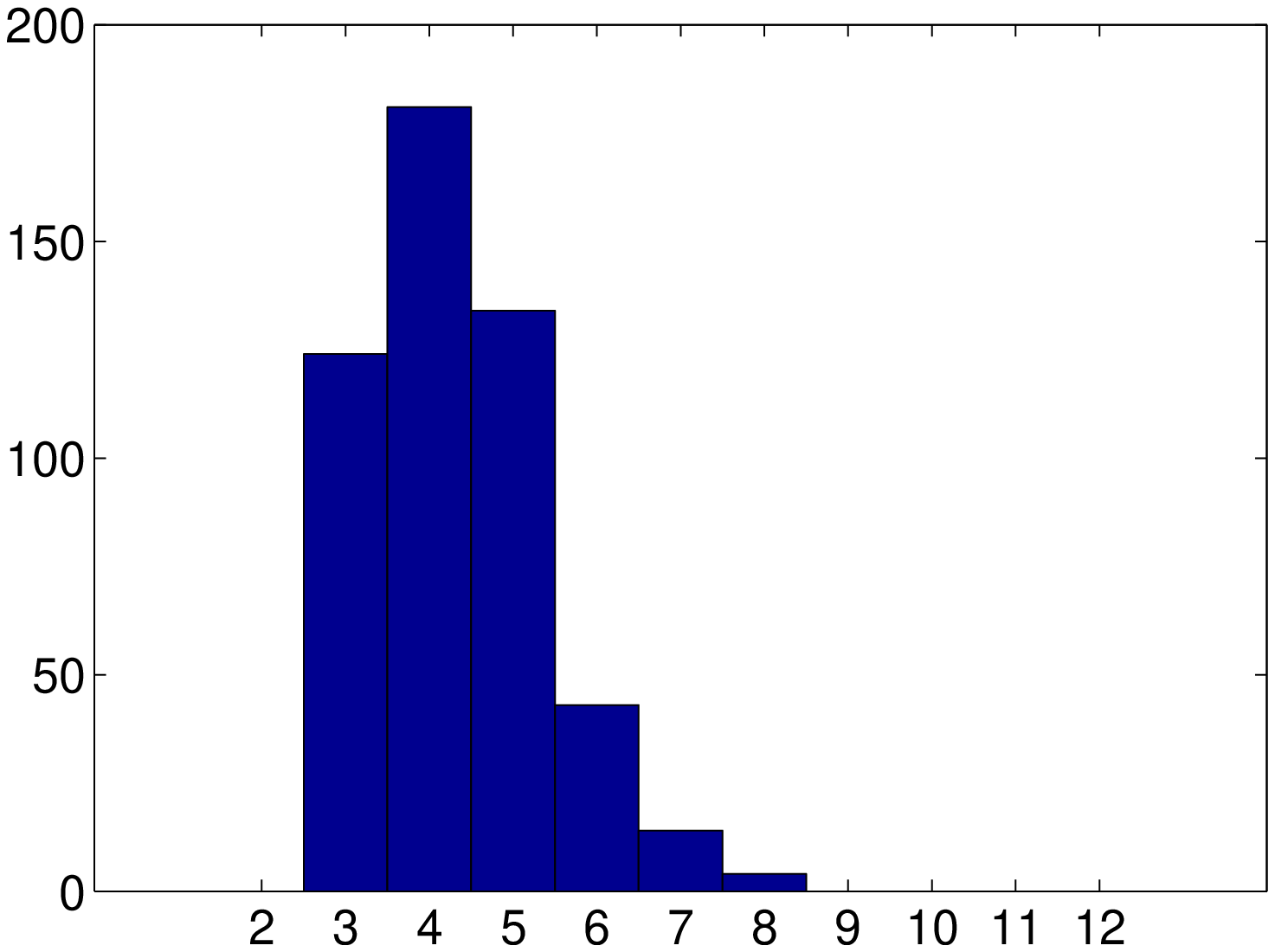}&
\includegraphics[height=1in]{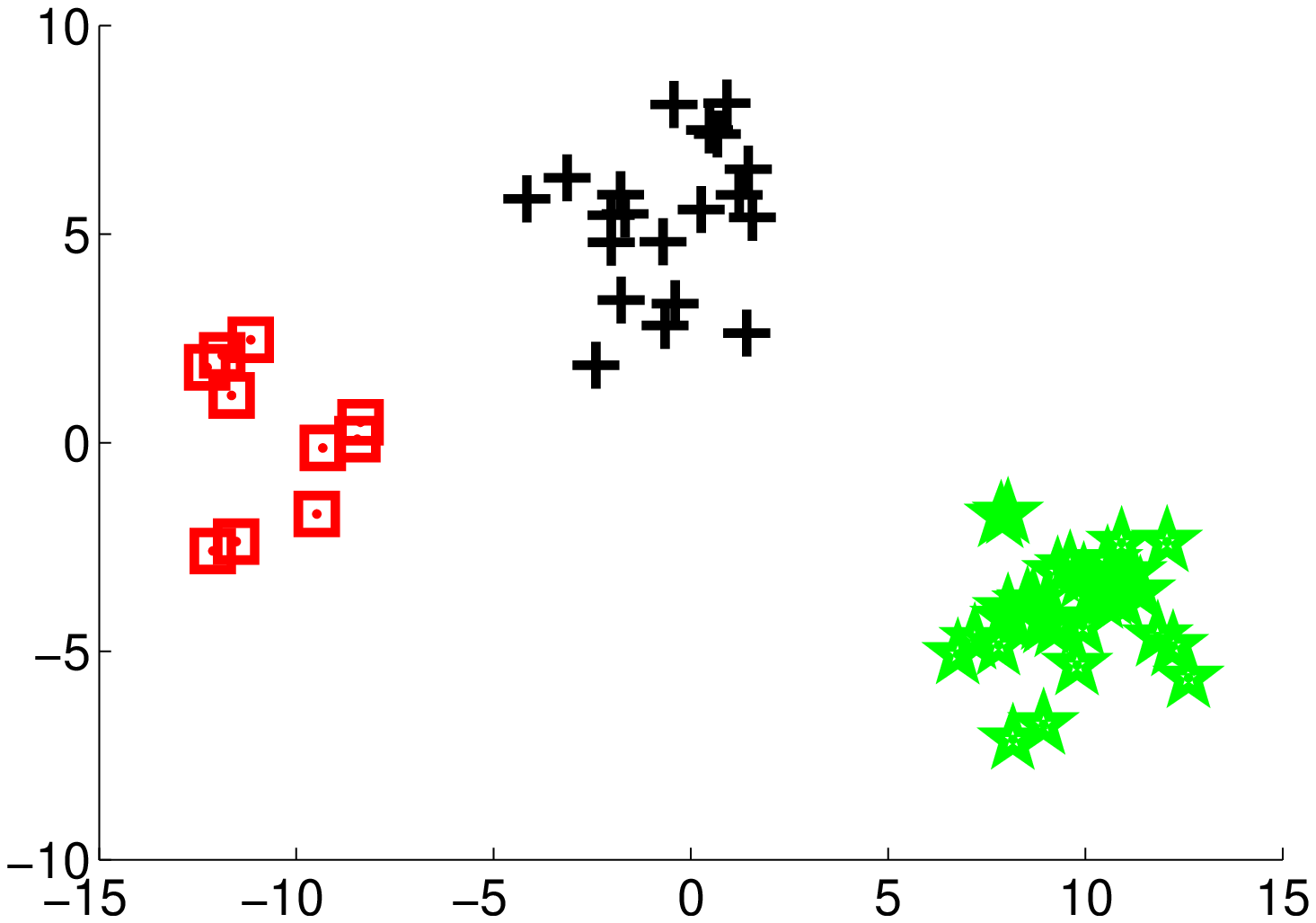}&
\includegraphics[height=1in]{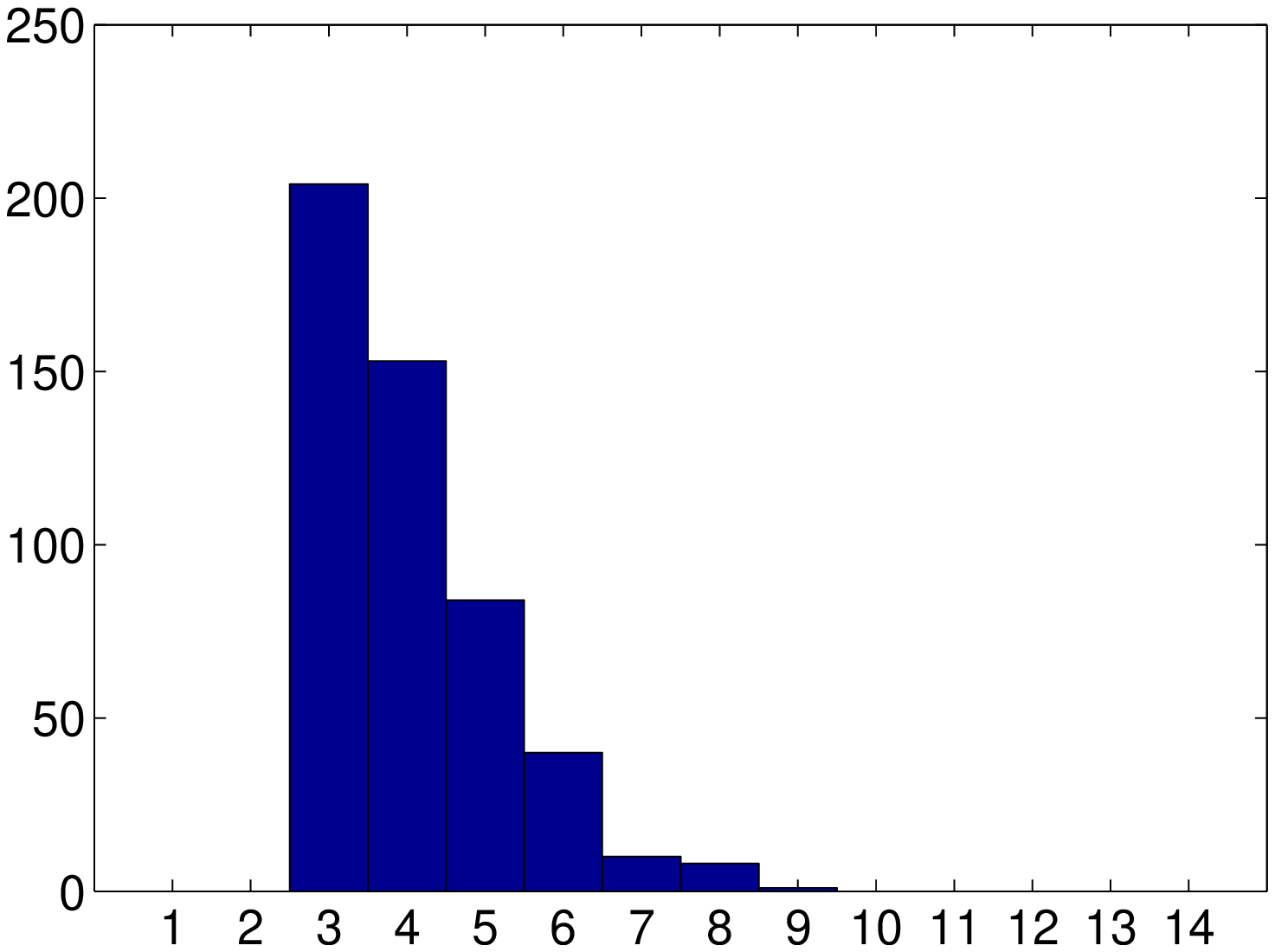}\\
\includegraphics[height=1in]{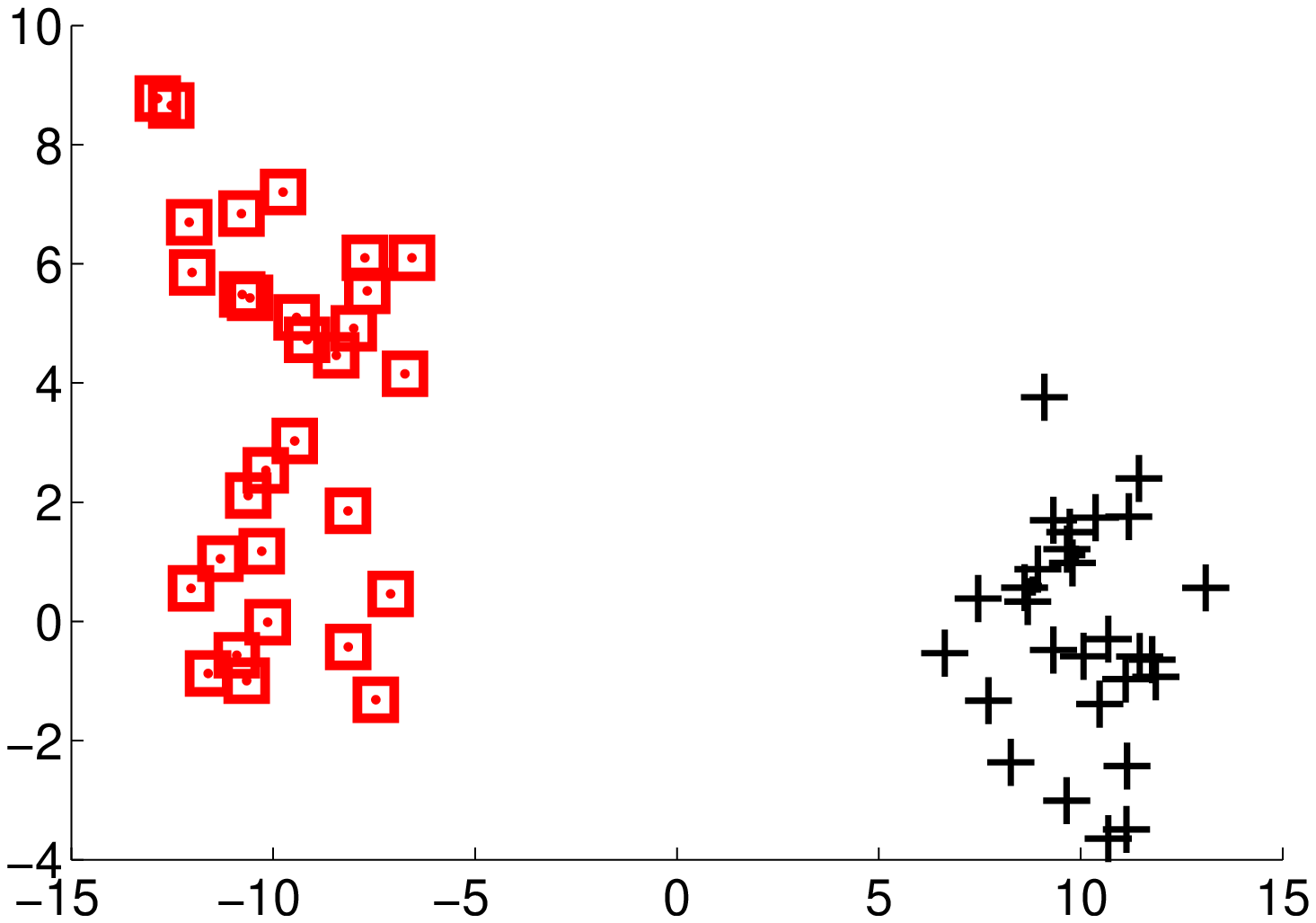}&
\includegraphics[height=1in]{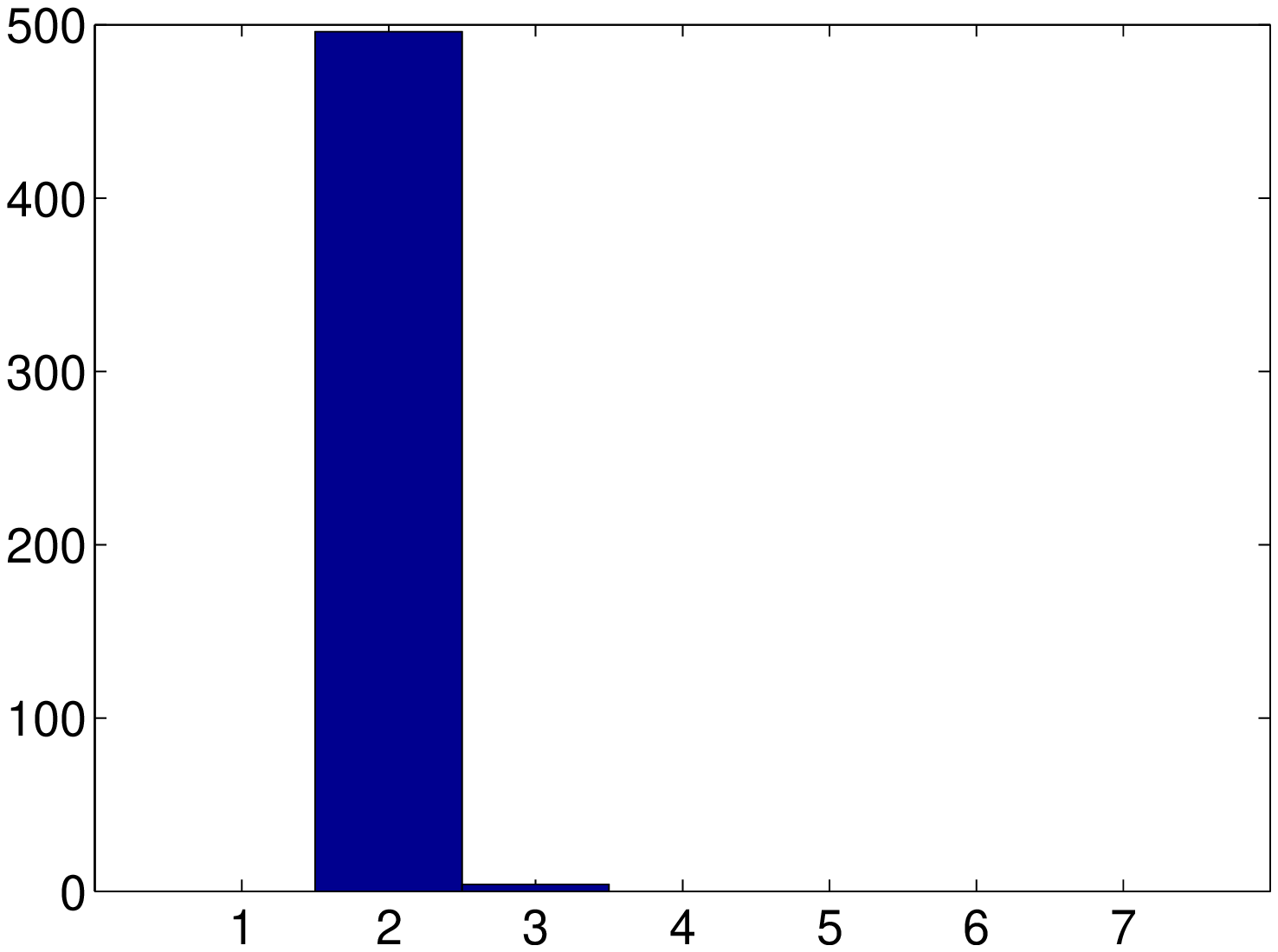}&
\includegraphics[height=1in]{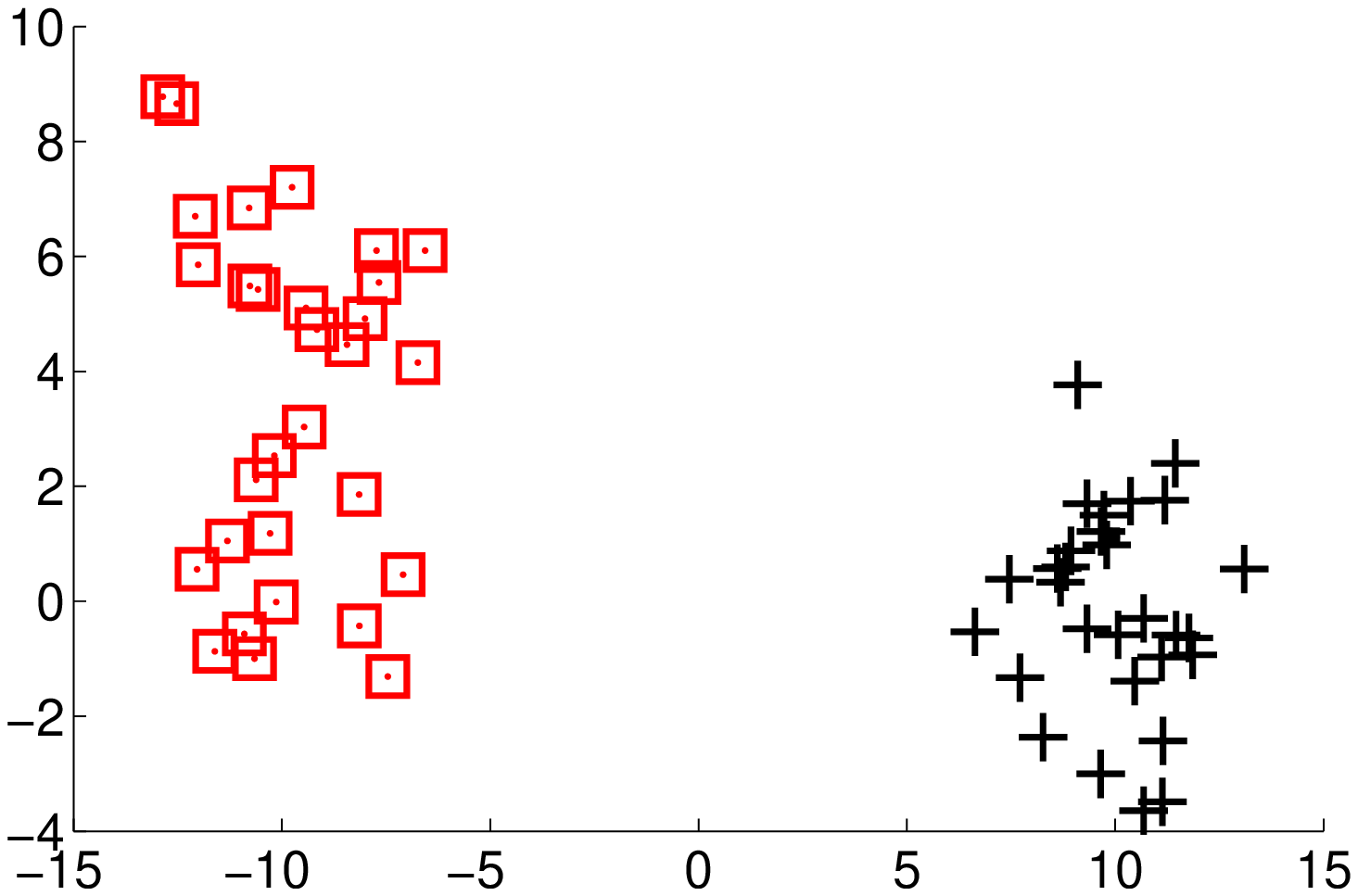}&
\includegraphics[height=1in]{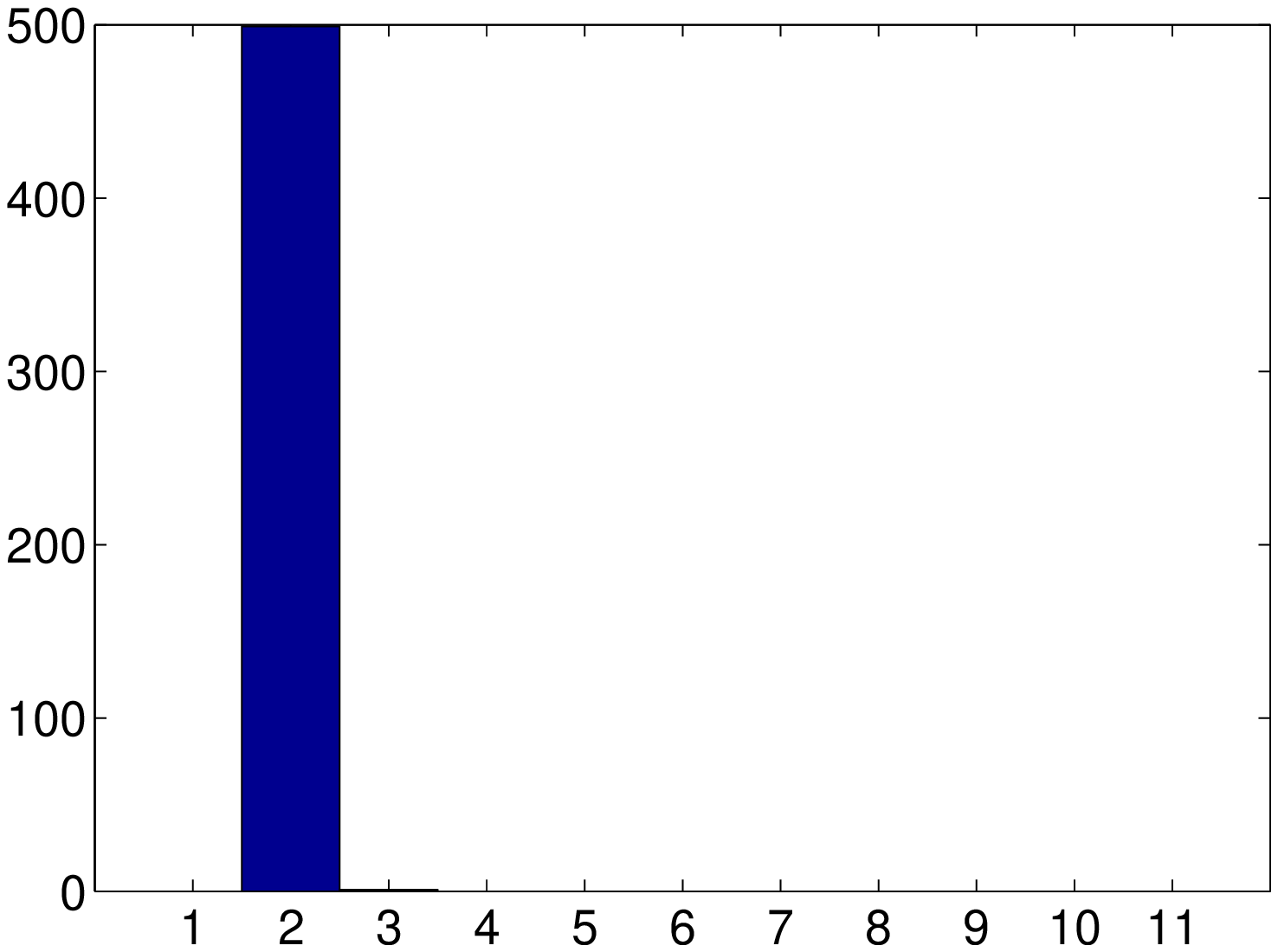}\\
\includegraphics[height=1in]{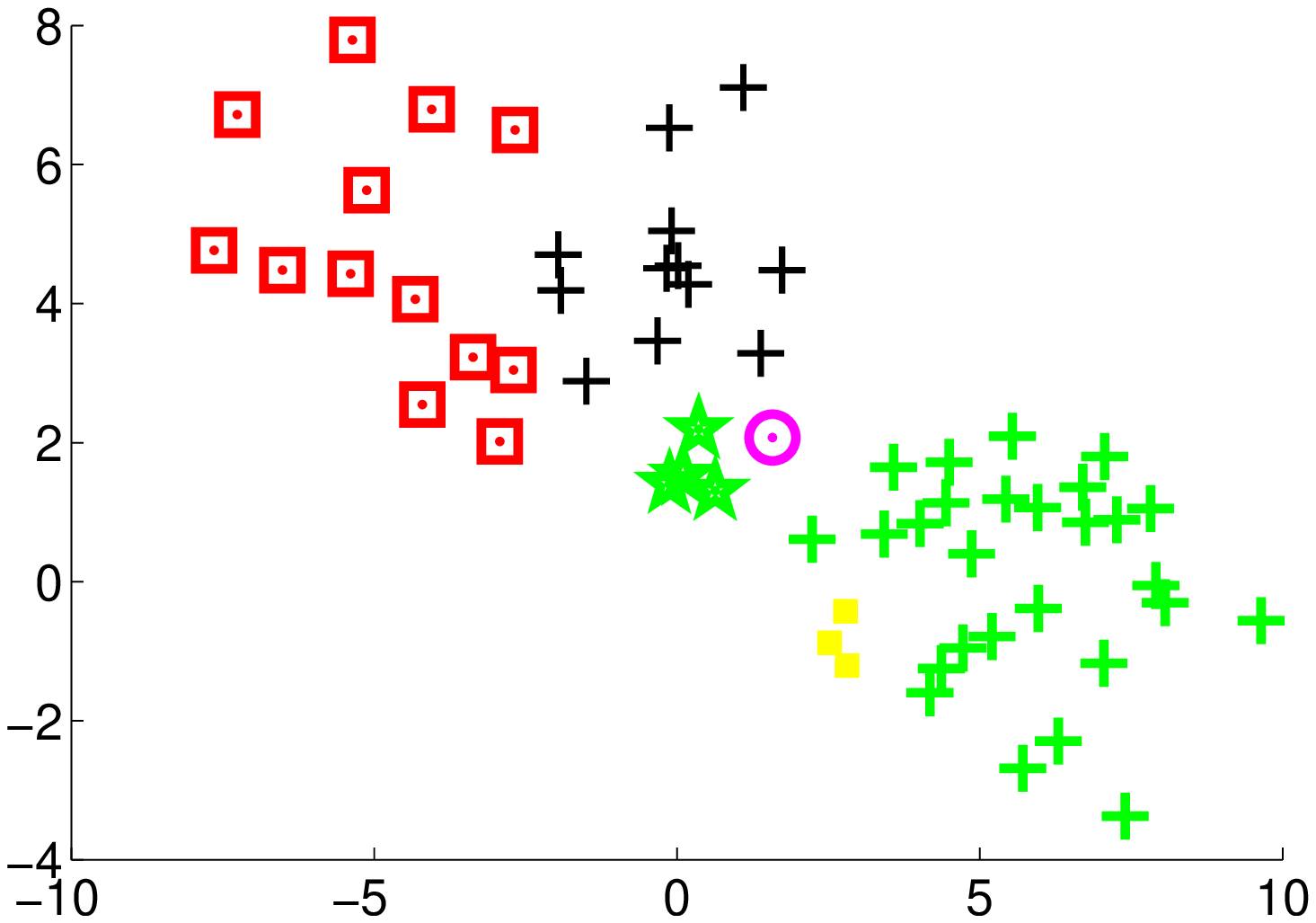}&
\includegraphics[height=1in]{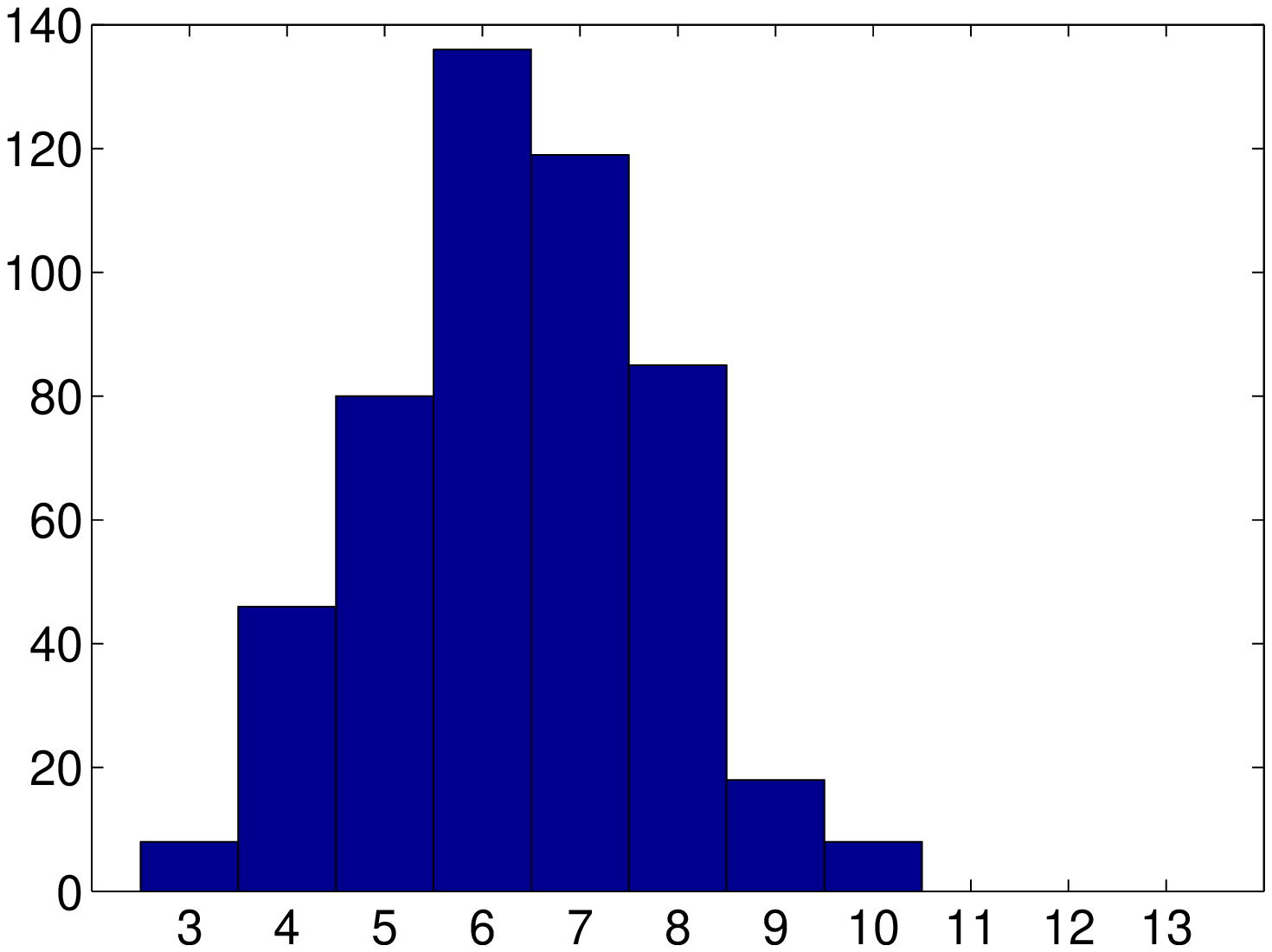}&
\includegraphics[height=1in]{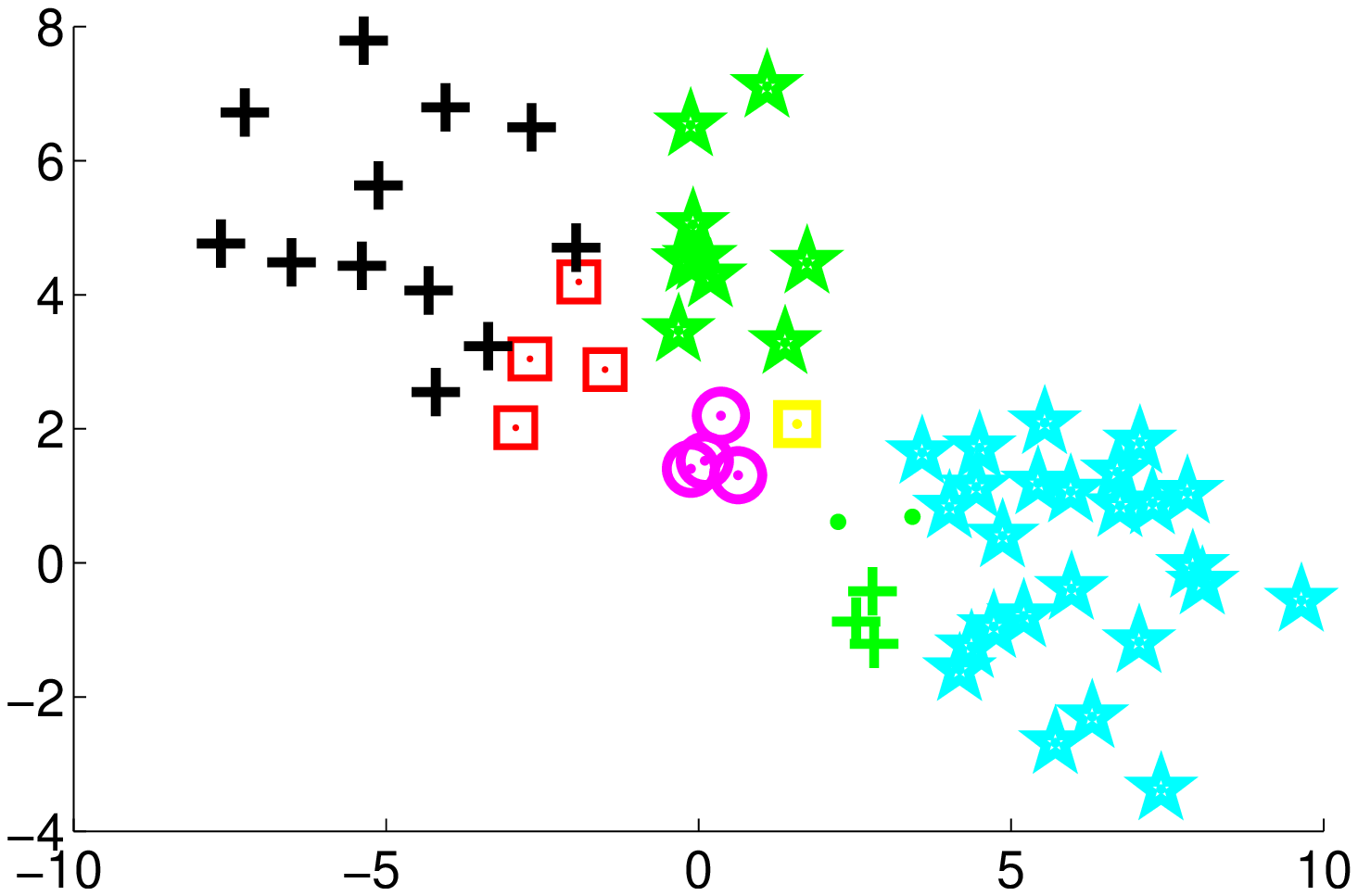}&
\includegraphics[height=1in]{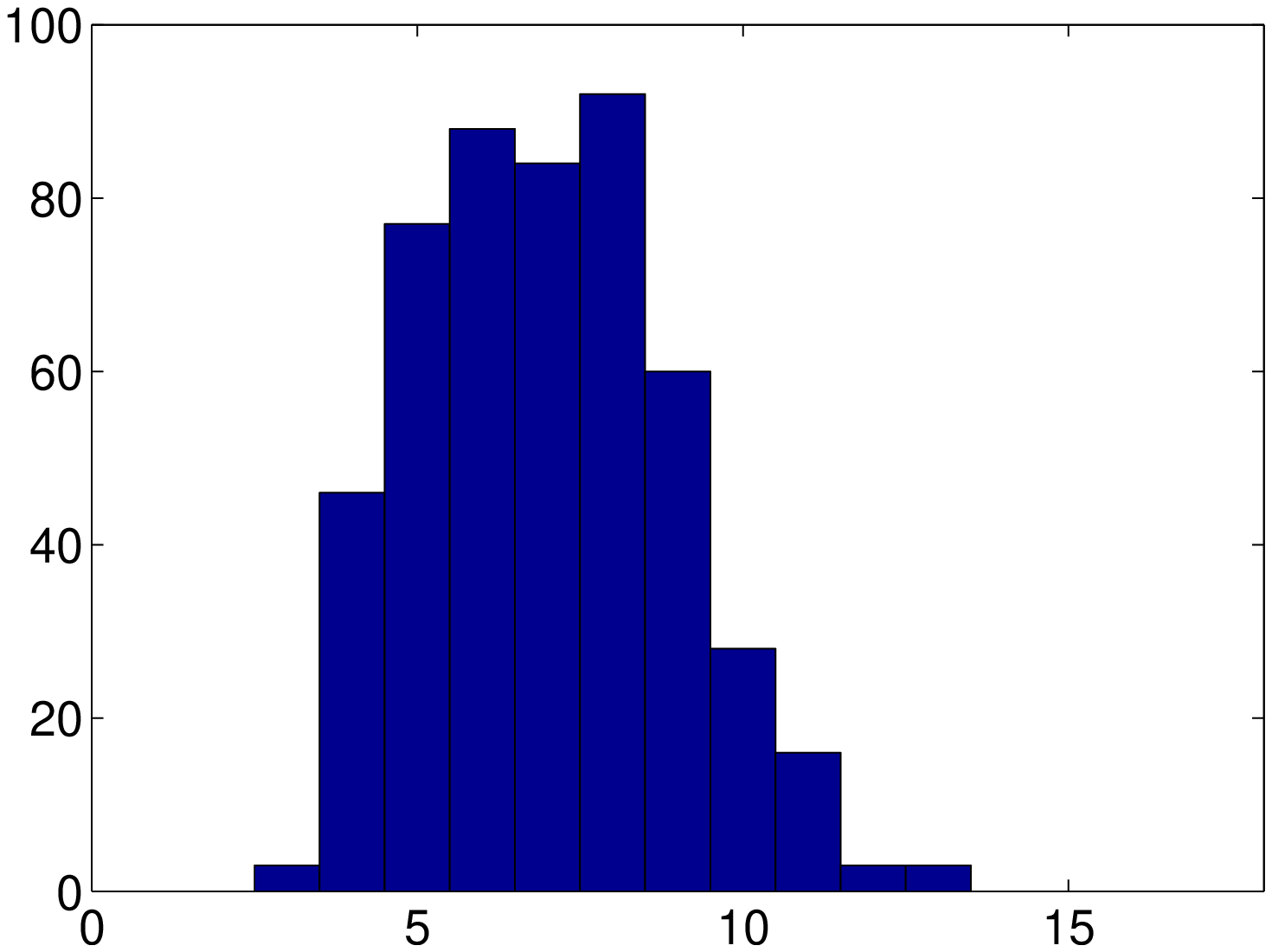}\\
\hline
\hline
\end{tabular}
\caption{Clustering with DPM of Gaussians and MFM of Gaussians. The left panel shows the results from DPM of Gaussians and the right panel shows the results form MFM of Gaussians. }
\label{fig:dmpmfm}
\end{center}
\end{figure}

For comparisons in the Euclidean case,  we  use the Dirichlet process mixture (DPM) of Gaussians \citep{Heller2005} and MFM of Gaussians directly on the observations $y_i, i=1, \ldots,n$.  Note that, unlike in the non-Euclidean case,  the mean and covariance matrix can be efficiently obtained in the Euclidean case. Fig. \ref{fig:dmpmfm} shows the results for the same dataset as in Fig. \ref{fig:euclidpara}. The left panel shows the results with DPM of Gaussians and the right panel shows the results with MFM of Gaussians. One can observe from Fig. \ref{fig:dmpmfm} that  
 both the methods tend to produce extraneous clusters compared to W-CRP.  This confirms the inconsistency of DPM as in \citep{miller2013inconsistency}  and also demonstrates that the convergence of MFM is slow, although theoretically it might produce consistent estimates of the number of clusters asymptotically. 

\begin{figure}
\begin{center}
\begin{tabular}{cc}
\includegraphics[height=1.8in]{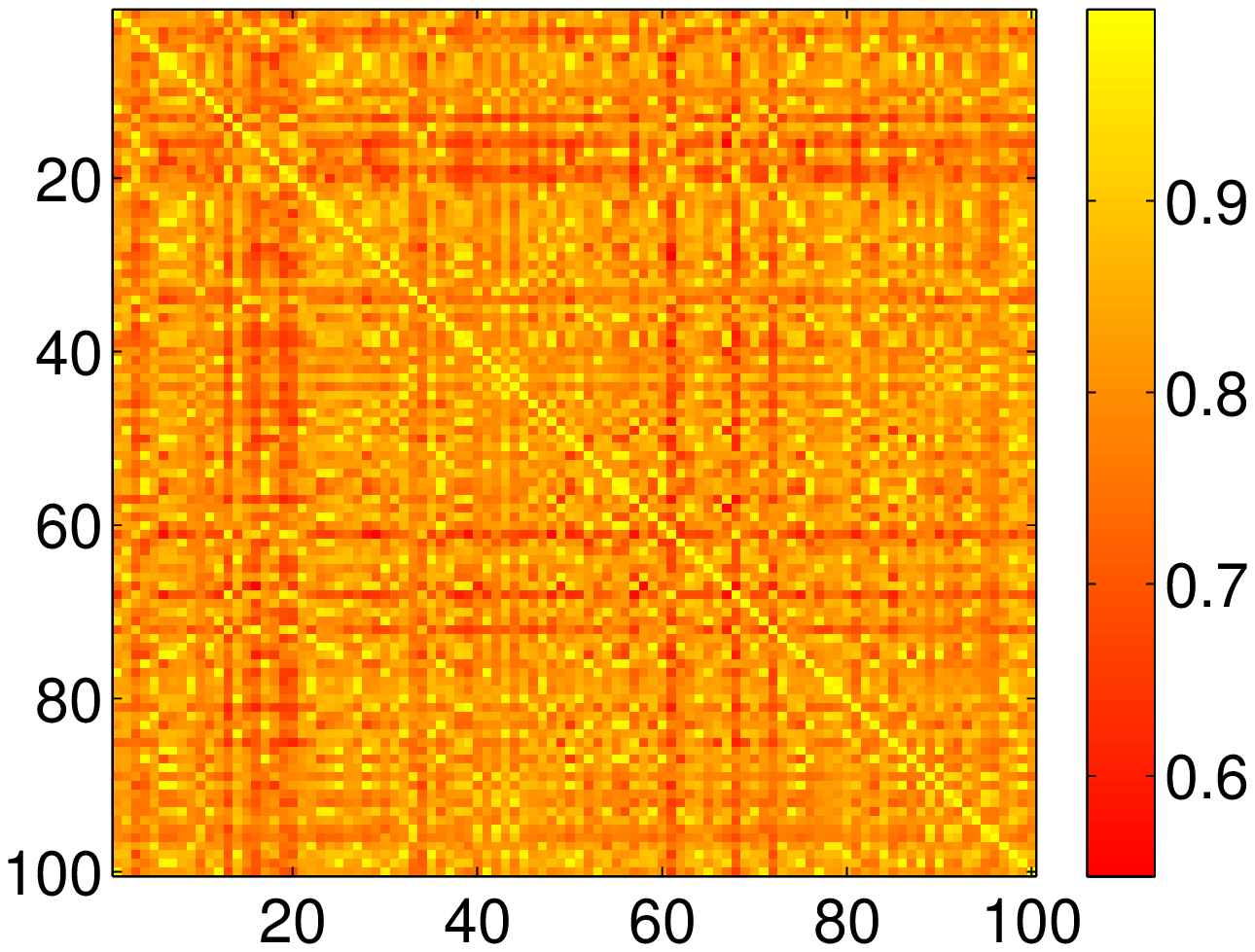}&
\includegraphics[height=1.8in]{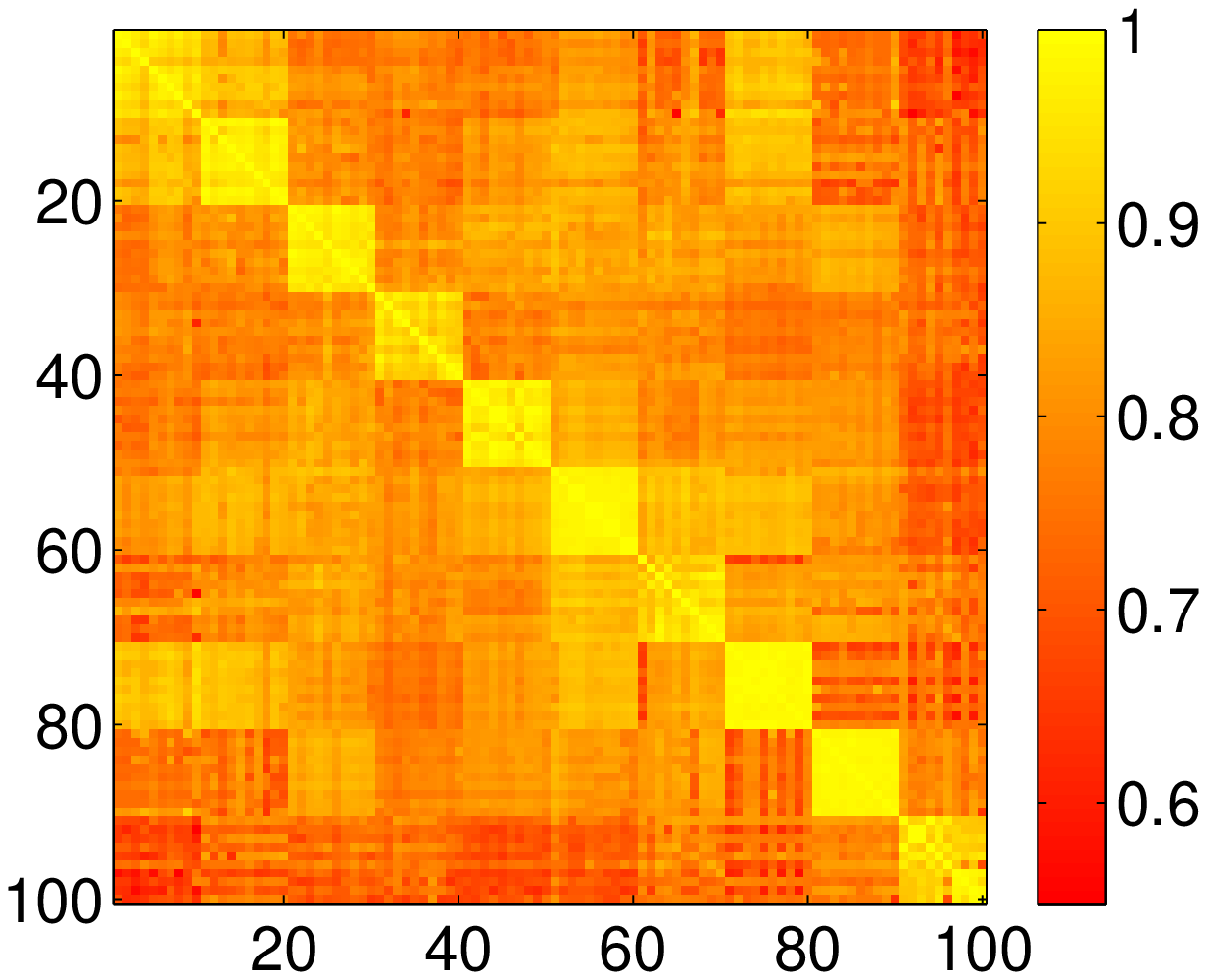}\\
(a) I-P Matrix Before & (b) I-P Matrix After \\
\includegraphics[height=2.4in]{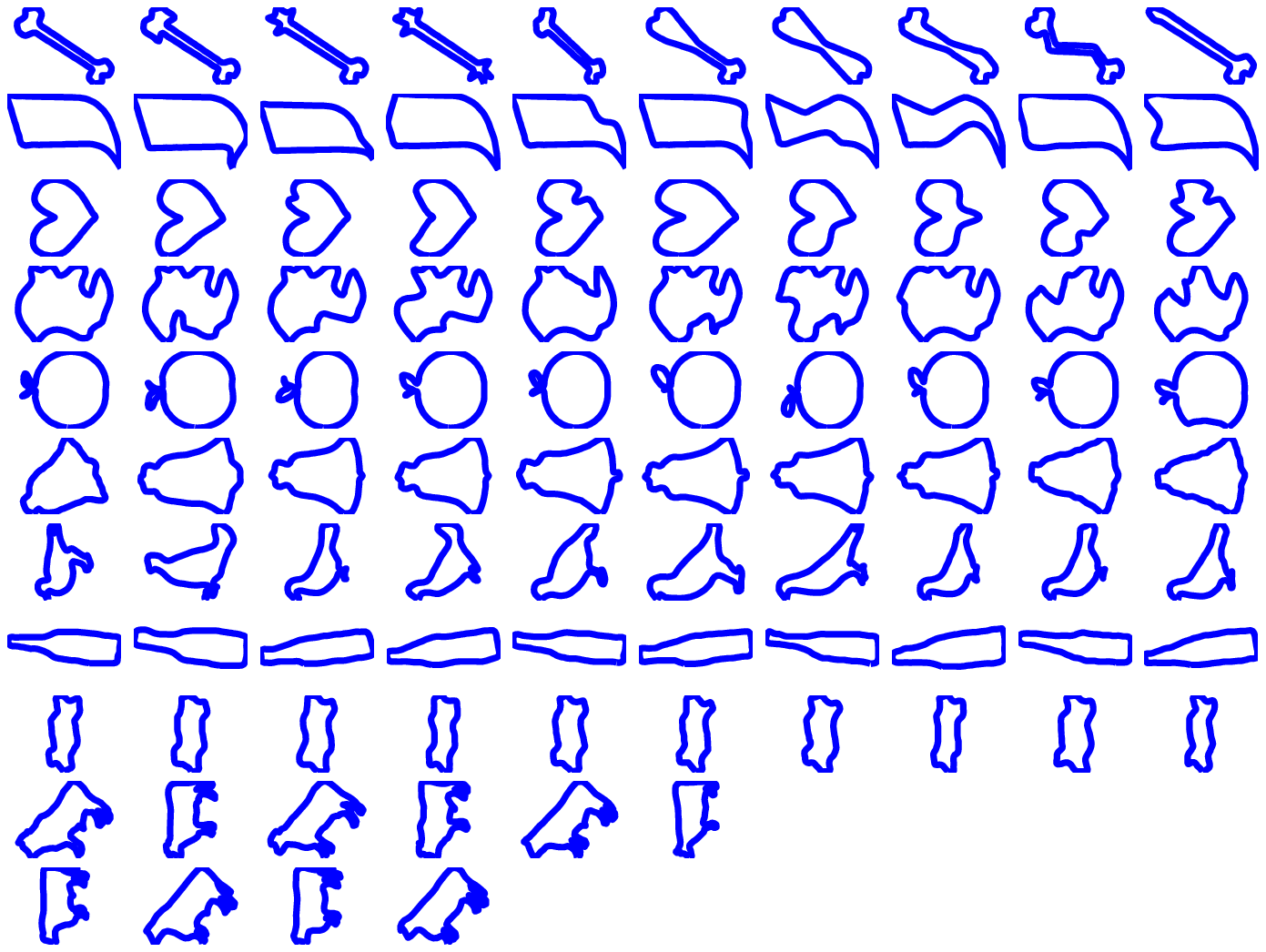}&
\includegraphics[height=1.6in]{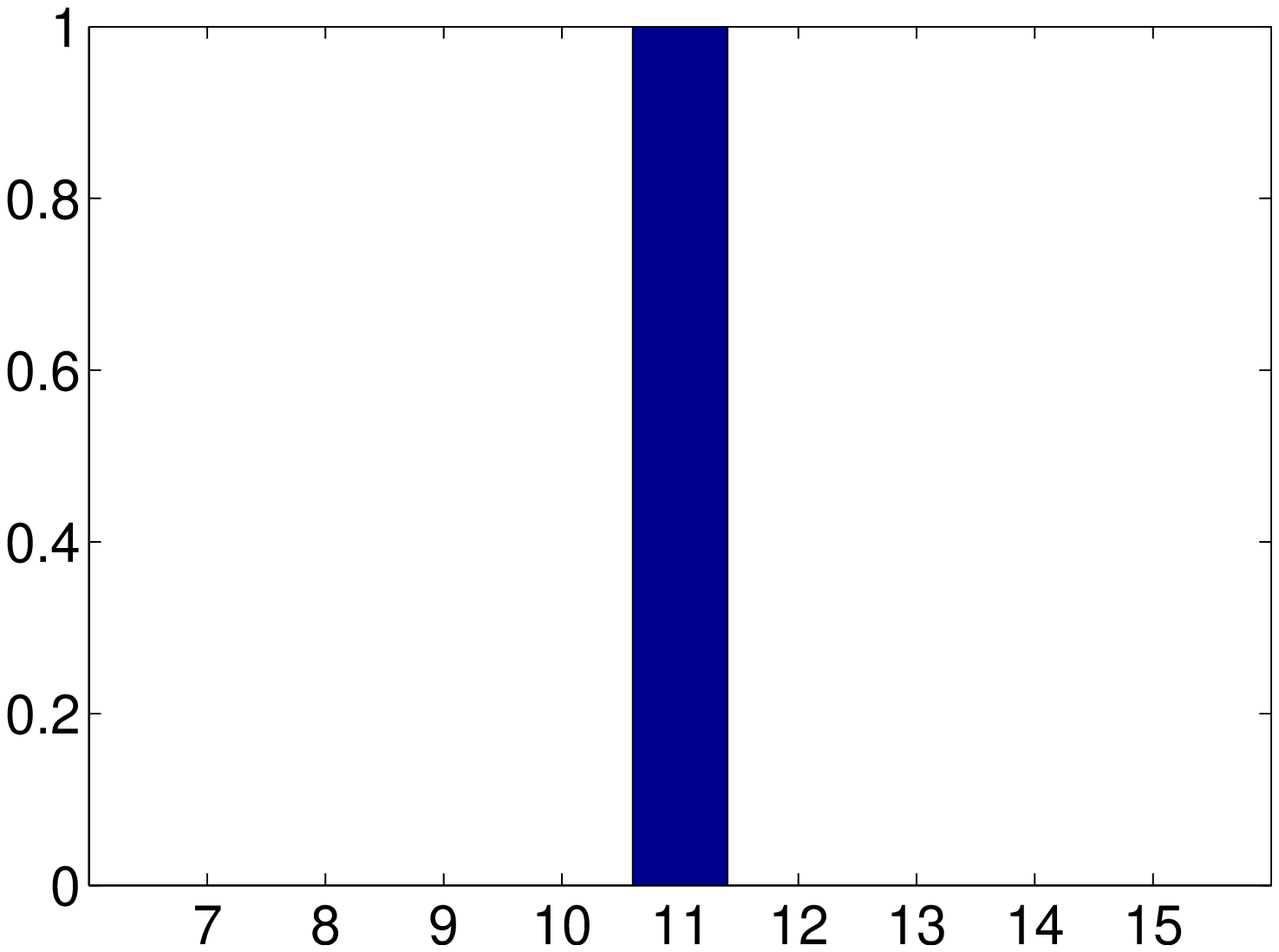}\\
(c) Clustering Result & (d) Histogram of k\\
\end{tabular}
\caption{Clustering process for $100$ shapes. The histogram shows the posterior distribution of the cluster number $k$ obtained from $4000$ MCMC samplings of $B$ without any burn-in. }
\label{fig:shape100}
\end{center}
\end{figure}

\noindent {\bf Non-Euclidean Shape Data}: \\ In this experiment, we study shapes taken from the MPEG-7 database \citep{mpeg7}. The full database has 1400 shape samples, 20 shapes for each class. We first choose 100 shapes to form a subset of 10 classes with 10 shapes from each class. The observations are randomly permuted and the inner product matrix $S$ is calculated using Definition \ref{def:innerpm}.  Then we perform our clustering method on $S$ (note that $S \in U_+(\real)$). 
We impose a prior on $\alpha$ with Inv-Gamma$(3,4)$, and $\xi$ is estimated by $\tilde{K}/\log(100)$, where $\tilde{K}$ is an estimate of number of clusters ($\tilde{K} = 15$ in this case).  The clustering result is shown in Fig. \ref{fig:shape100}, where (a) and (b) shows the inner product (I-P) matrix before and after clustering, (c) shows the final clustering result, and (d) shows the histogram of cluster number $K$ obtained from $4000$ MCMC samples of $B$. From the result, one can see that our algorithm clusters these $100$ shapes well other than splitting one class.

\begin{figure}
\begin{center}
\begin{tabular}{c}
\includegraphics[height=2.6in]{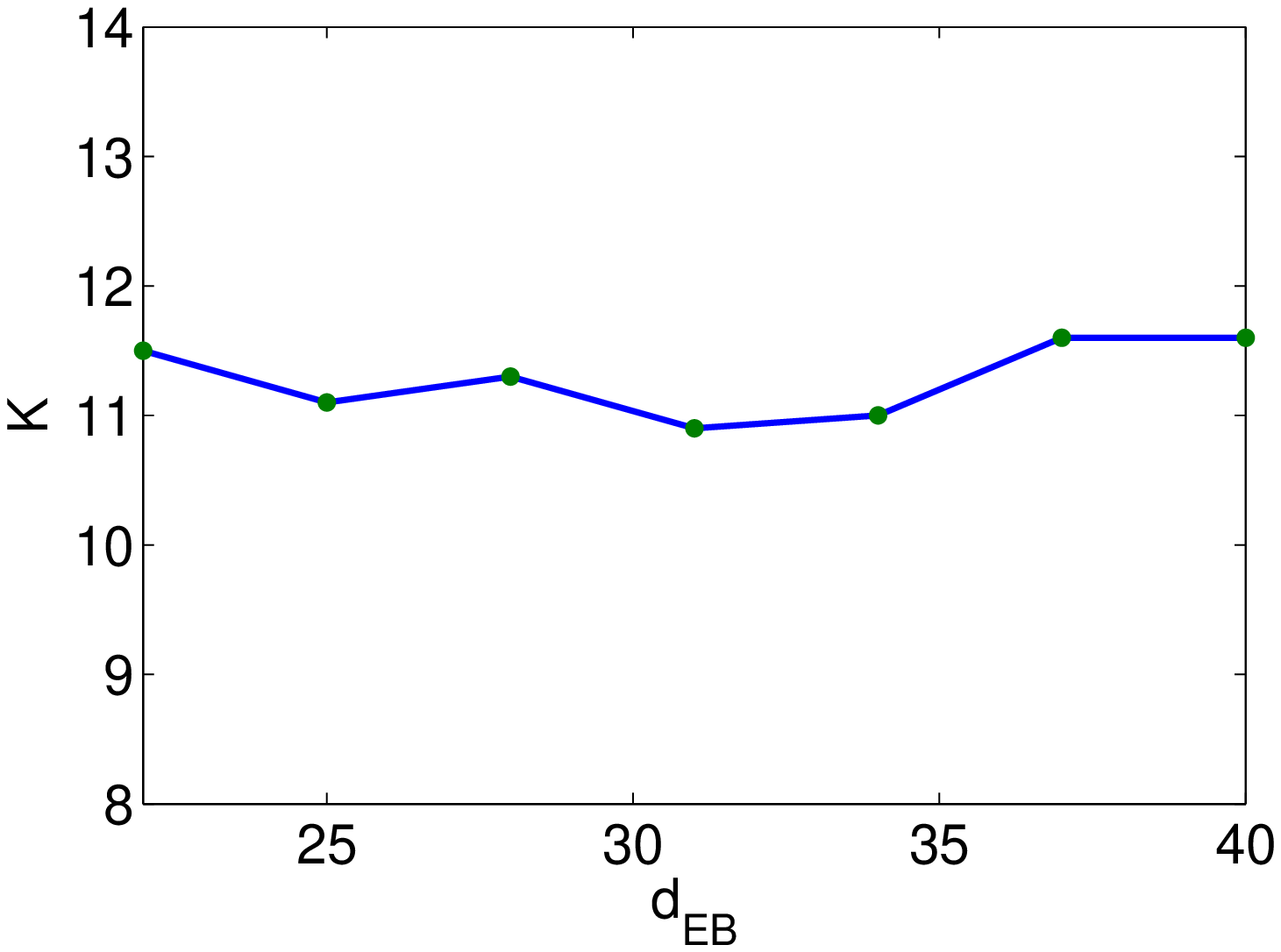}
\end{tabular}
\caption{Clustering sensitivity analysis of parameter $d$ in the dataset shown in Fig. \ref{fig:shape100}.}
\label{fig:analyzed}
\end{center}
\end{figure}

In next experiment, we analyze the sensitivity of the cluster number $K$ to the parameter $d$, degrees of freedom of the Wishart distribution. Note that in the Euclidean case, $d$ can be easily estimated since $d$ (in the case of $d<n$) is the dimension of the data. 
Fig. \ref{fig:analyzed} shows the estimated cluster number $K$ versus the value of parameter $d$ in the dataset shown in Fig. \ref{fig:shape100}. It is evident that the estimates of $K$ are robust to different choices of $d$.    

To compare with existing methods in the shape domain, we test our method on another subset of MPEG-7 dataset that was used in \citep{Bicego2004,Bicego2004pr,Thakoor2007}. The dataset contains $6$ classes of shapes with $20$ shapes per class. To quantify the clustering result, we use the ``classification rate'' defined in \citep{Jain1988}. For each cluster, we note the predominant shape class, and for those shapes assigned to the cluster which do not belong to the dominant class are recognized to be misclassified. The classification rate is the total number of dominant shapes for all classes divided by the total number of shapes. However, this measure is known to be sensitive towards larger clusters. 
  The Rand index  \citep{Torsello2007} is an alternative measure of the quality of classification which  measures the similarity between the clustering result and the ground truth, defined as $RI = a/\binom{n}{2}$. Here $a$ is the number of the ``agreements'' between the clustering and the ground truth, which is defined as the sum of two quantities: (1) the number of pairs of elements belonging to the same class that are assigned to the same cluster; (2) the number of pairs of elements belonging to different sets that are assigned to different classes. 
  If the clustering result is the same as the ground truth, $RI = 1$, otherwise $RI<1$. The Rand index penalizes the over-segmentation while the classification rate does not.   Table \ref{tabac} compares the overall classification rate and Rand index of our method with other methods, such as Fourier descriptor combined with support vector machine based classification (FD + SVM), hidden Markov model (HMM + Wtl) with weighted likelihood classification \citep{Thakoor2007}, HMM with OPC approach (HMM + OPC) \citep{Bicego2004pr}, elastic shape analysis (ESA) \citep{AnujShape} with k-medians (K-medians), ESA with pairwise clustering method (ESA + PW) \citep{Srivastava05shapte}. Our model, with Wishart-CRP applied on the elastic  inner product (EIP) matrix is denoted by EIP + W-CRP. The classification rate, Rand index and the computational time of K-medians, ESA + PW and our method are obtained based on the average of $5$ runs on a laptop with a i5-2450M CPU and 8GB memory. The computational time of our approach (EIP + DW) includes the cost of calculating the inner product matrix $S$ ($642.6$ s) and generating the $4000$ MCMC samples ($131.6$ s). An faster approach for calculating the elastic inner product matrix defined in our paper is available in \citep{wen2014}.  
    For ESA + PW and K-medians method, we set $K=6$ since we know the true $K$ in this case.  The classification rates for FD+ SVM, HMM + Wtl, and HMM + OPC are reported from \citep{Thakoor2007}, and these rates are based on the $1$-nearest neighbor classification. As evident from the results, our model can automatically find the cluster number $K=6$, and the classification rate is better than the competitors.

\begin{table*}
\caption{Comparison of the classification rate on MPEG-7 dataset}
\label{tabac}
\begin{center}
\begin{tabular}{ l|l|l|l|l|l|l }
\hline
Classifier & \begin{tabular}{@{}c@{}}FD+ \\ SVM\end{tabular} & \begin{tabular}{@{}c@{}}HMM+ \\ WtL\end{tabular} & \begin{tabular}{@{}c@{}}HMM+ \\ OPC\end{tabular} &\begin{tabular}{@{}c@{}}K-  \\ medians\end{tabular} &\begin{tabular}{@{}c@{}}ESA+ \\ PW \end{tabular} &  \begin{tabular}{@{}c@{}}EIP+ \\ W-CRP \end{tabular}  \\
\hline
\hline
Classification rate (\%)& 94.29 & 96.43 &97.4 & 81.5 & 96.67&  100.00 \\
\hline
Rand index & - & - & - &0.91& 0.98&  1.00 \\
\hline
Time (seconds) & - & - & - &648.5 & 707.4 &  774.2 \\
\hline
\end{tabular}
\end{center}
\end{table*}

\subsection{Real Data Study}\label{ssec:real}
In this section, we cluster cell and protein shapes introduced in Section 2.  Automated clustering is a crucial goal in real data applications where it is hard to provide a rough estimate of the number of clusters visually.   In the examples where the ground truth labels are available, our methods provide higher classification rates.

%

\noindent 
{\bf Cell shapes clustering:} We first cluster the cell shape data introduced in Section 2.2. The cell shape dataset includes cell shapes from different cell types: DLEX-p46 cells, and NIH-3T3 cells. DLEX-p46 cells are round, whereas NIH-3T3 cells have an elongated, spindly appearance.

\begin{figure}
\begin{center}
\begin{tabular}{|cc|c|}
\hline
\multicolumn{2}{|c|}{Small $\theta \in \{0.2,0.4,0.6,0.8,1 \}$}&\\
\includegraphics[height=1.86in]{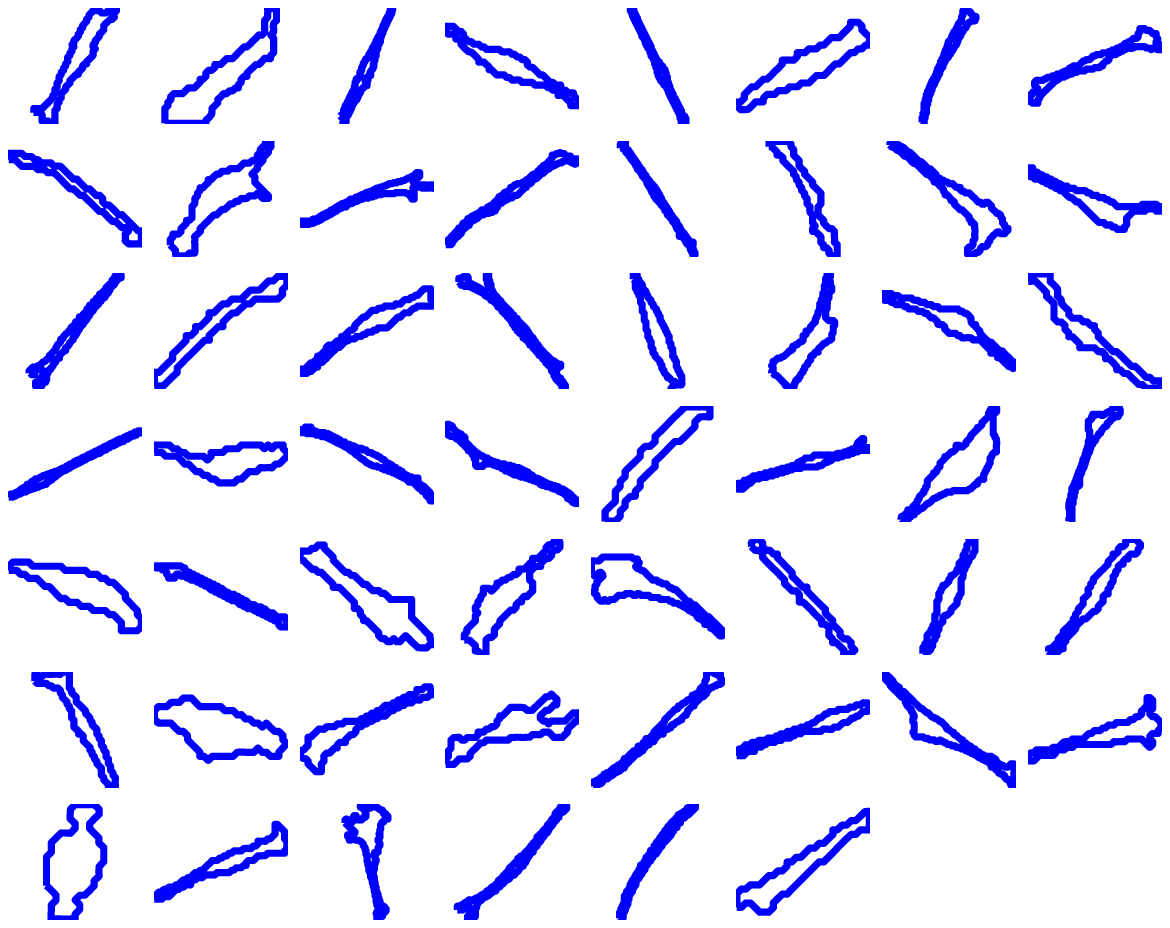}&
\includegraphics[height=1.86in]{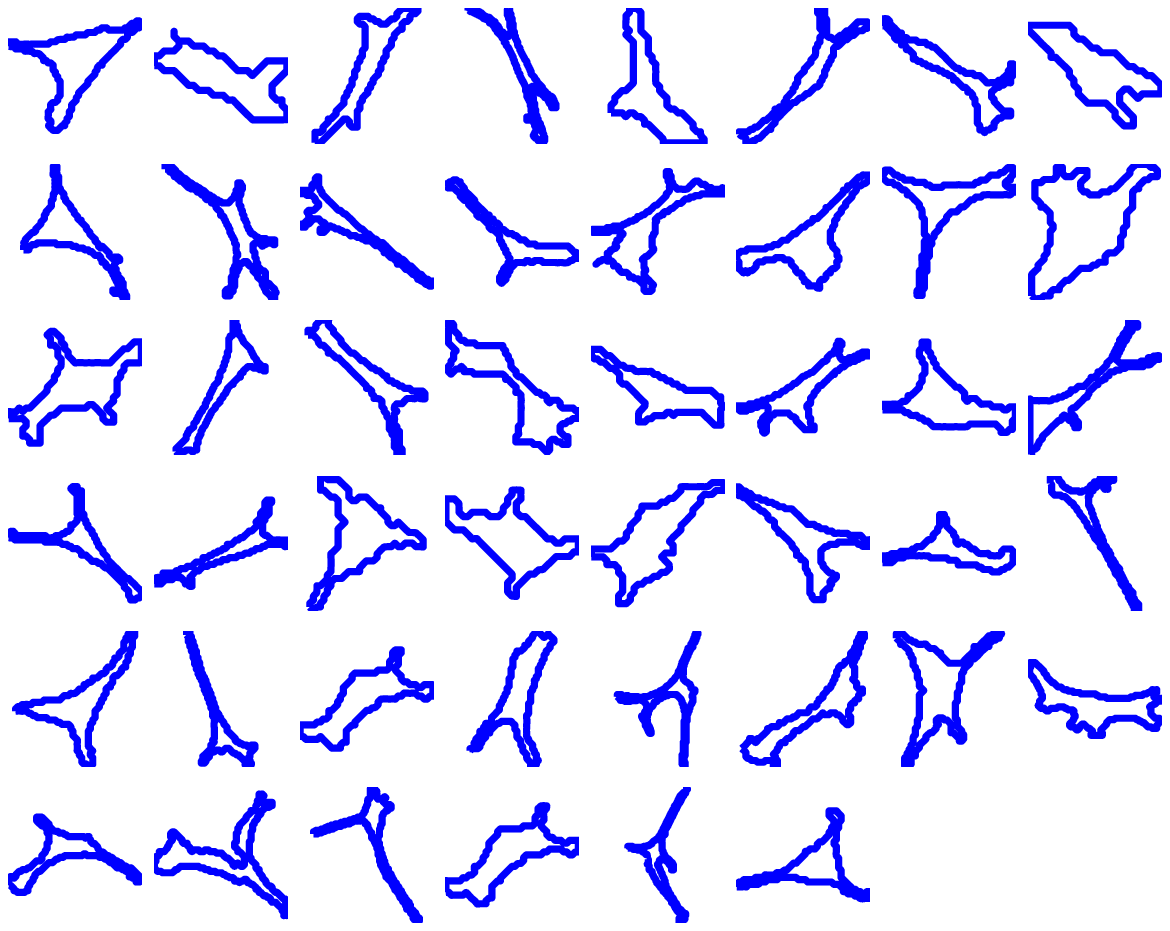}&
\includegraphics[height=1.2in]{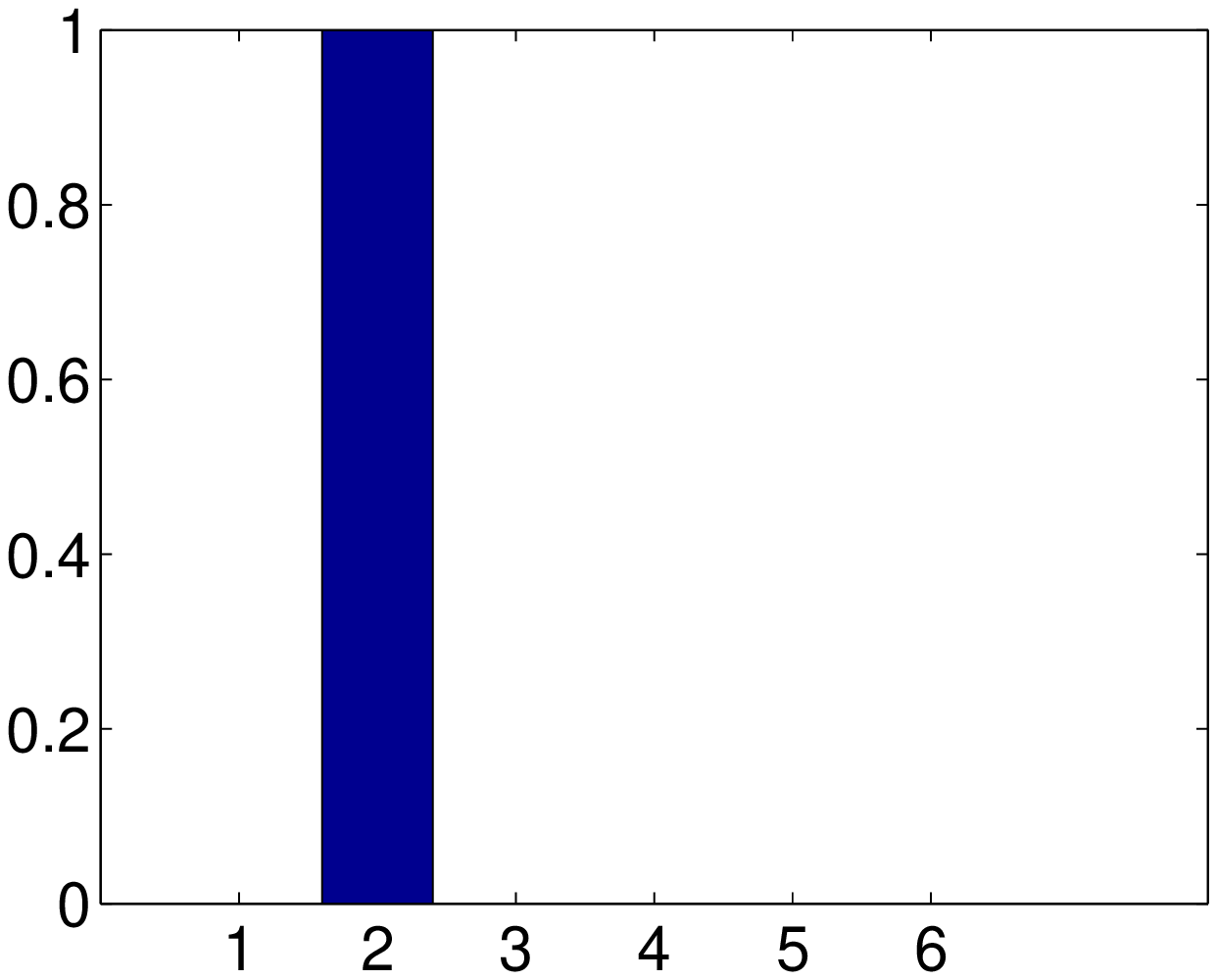}\\
Cluster 1 & Cluster 2 & Histogram of k \\
\hline
\end{tabular}
\begin{tabular}{|ccc|c|}
\multicolumn{3}{|c|}{Big $\theta \in \{200, 400,600, 800,1000\}$}&\\
\includegraphics[height= 1.2in]{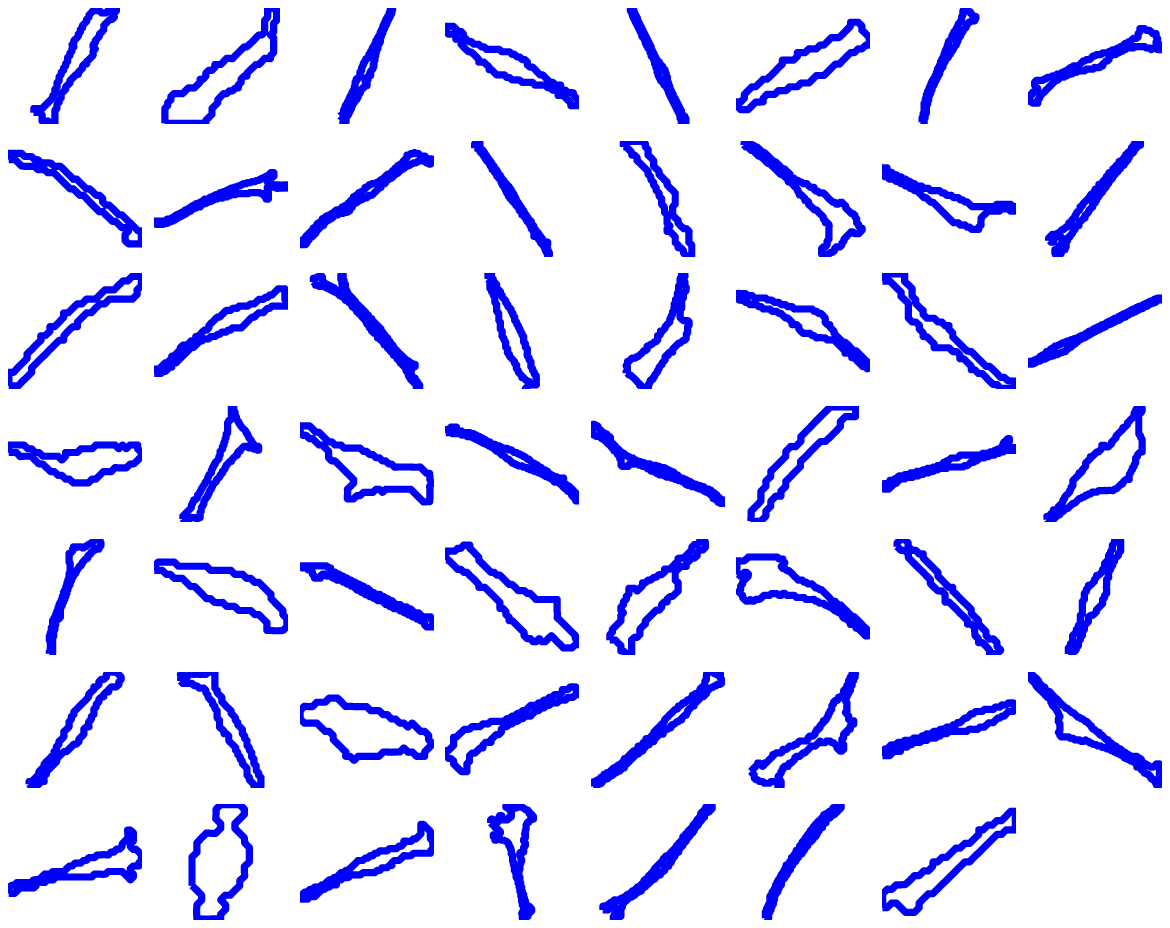}&
\includegraphics[height= 1.2in]{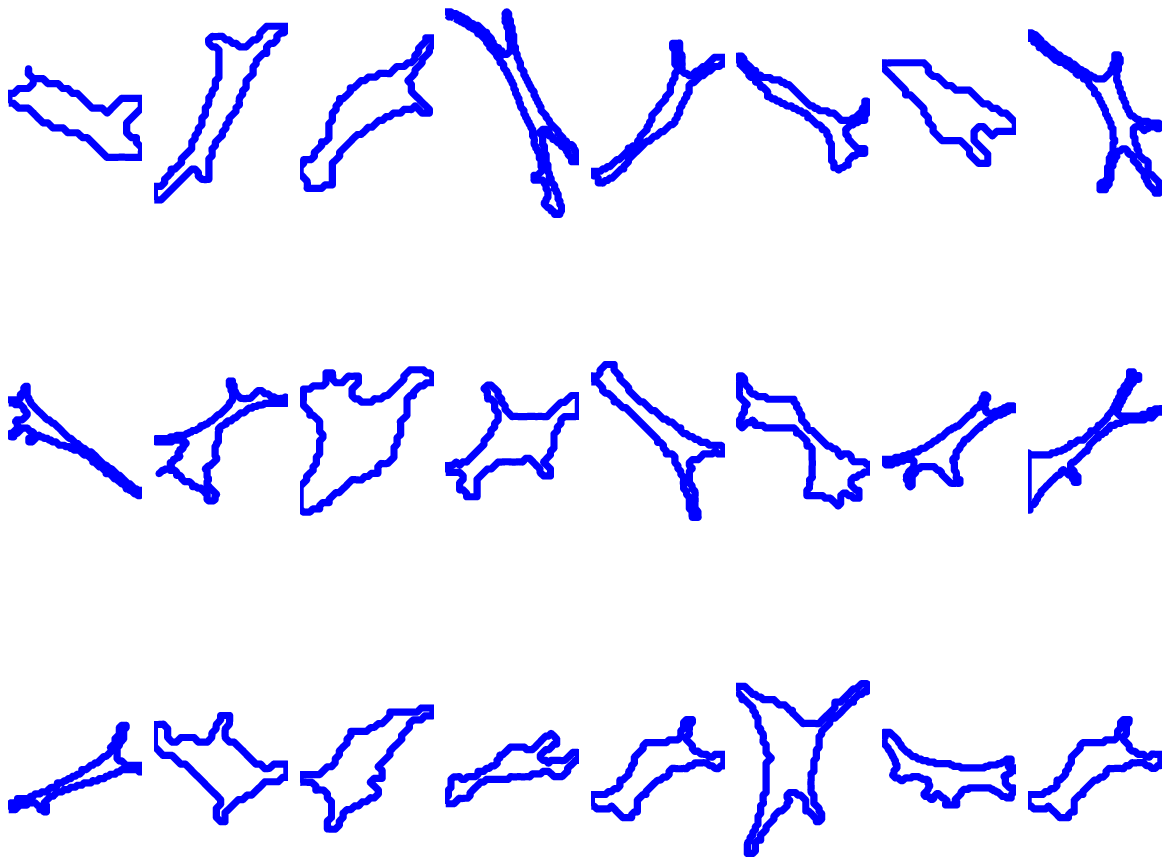}&
\includegraphics[height= 1.2 in]{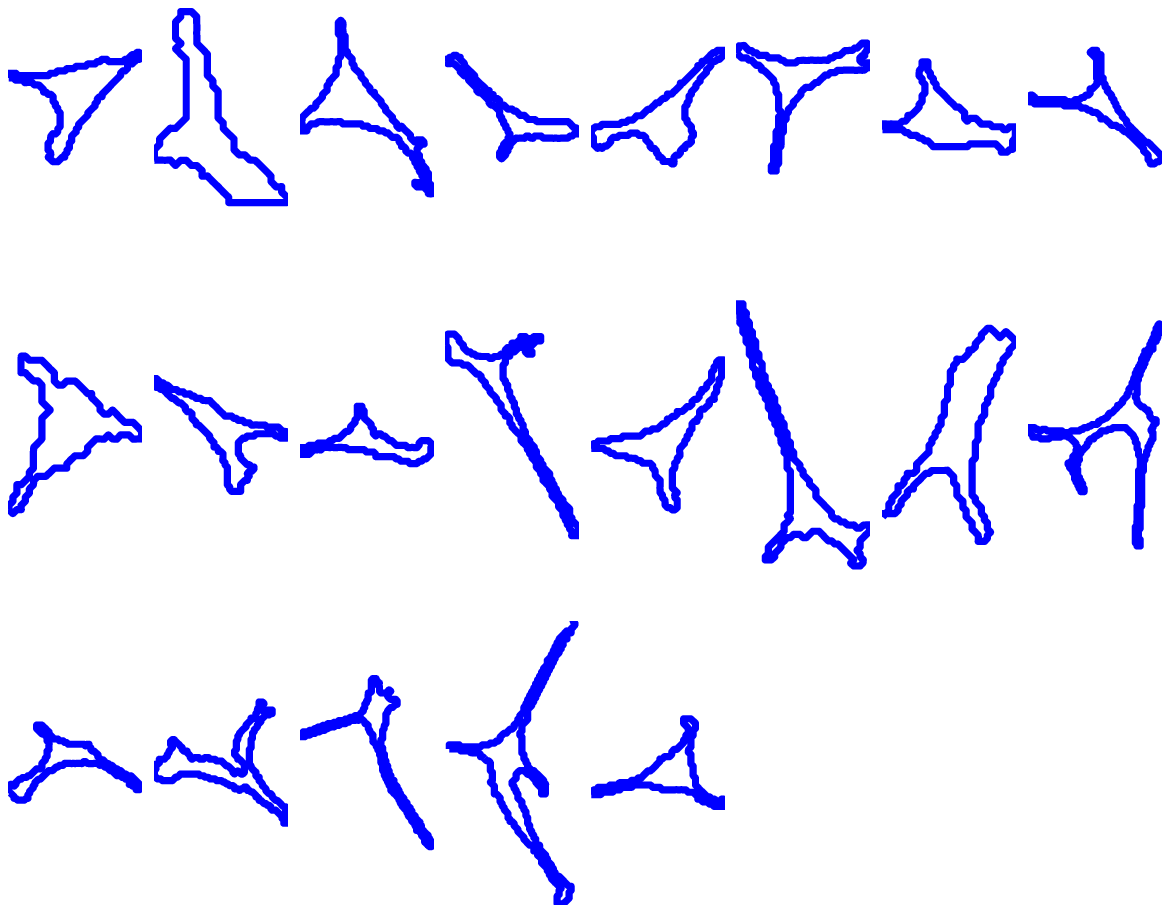}&
\includegraphics[height=1.2in]{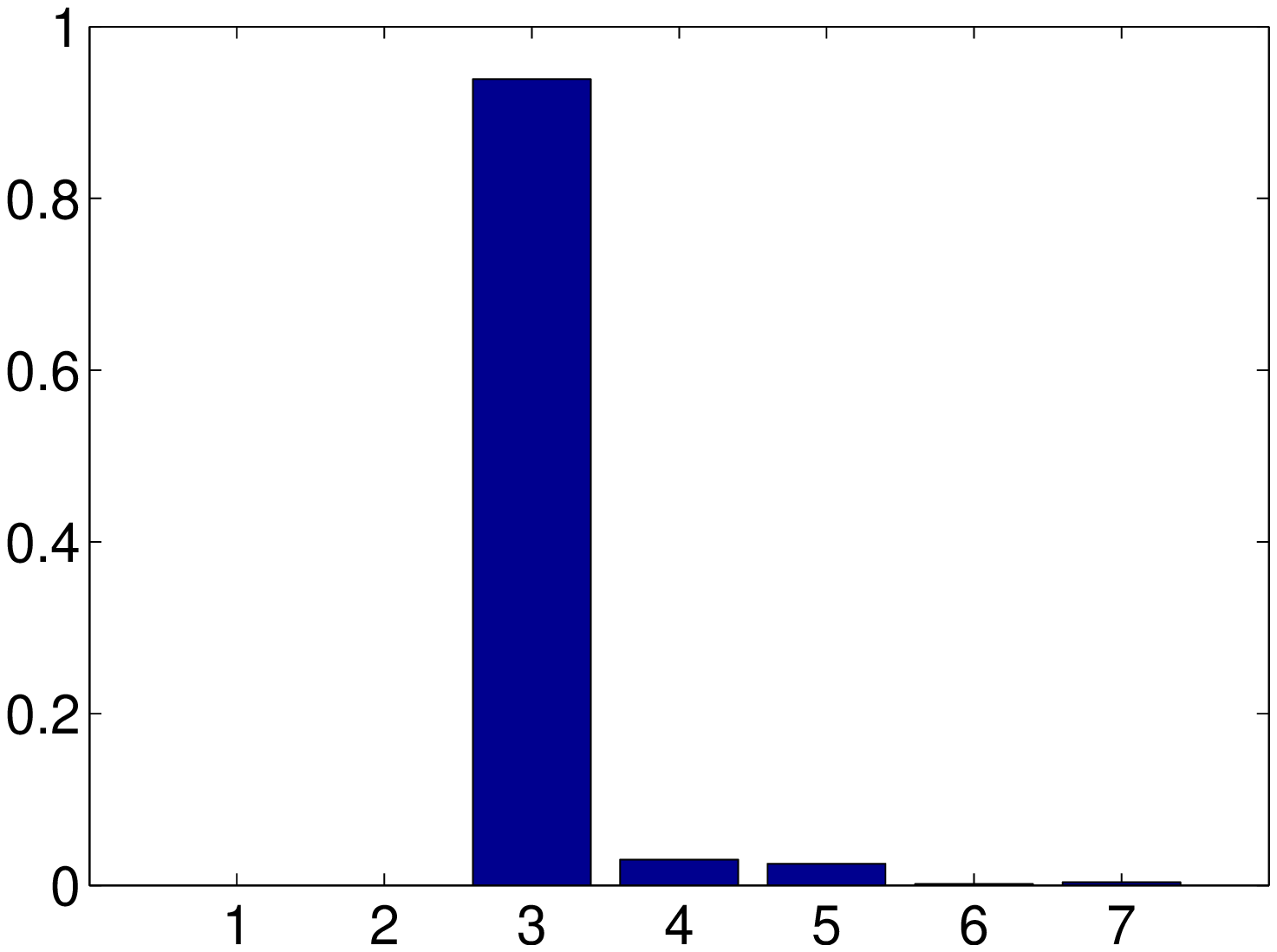}\\
Cluster 1 & Cluster 2 & Cluster 3 & Histogram of k\\
\hline
\end{tabular}
\caption{Clustering process for NIH-3T3 cell shapes with different $\theta$ values.}
\label{fig:shapeNIH}
\end{center}
\end{figure}

In the first experiment, we select $100$ shapes from NIH-3T3 cell shapes to form a subset, and cluster them with different priors on $\theta$: a set of small $\theta$ ($\theta \in \{0.2,0.4,0.6,0.8,1 \}$) and a set of large $\theta$ ($\theta \in \{200, 400,600, 800,1000\}$). Fig. \ref{fig:shapeNIH} shows the clustering result. The first row shows the clustering result with a small $\theta$ and the second row shows the result with a big $\theta$. 
One can see that with a small $\theta$, our method cluster the data into 2 classes: the first class of shapes only have two corners, and the second class has three or more corners. For a large $\theta$, we cluster the shapes into three classes: one with shapes have two corners, one with shapes have four or more corners and one with shapes have three corners.

\begin{figure}
\begin{center}
\begin{tabular}{|ccc|}
\hline
\includegraphics[height=1.5 in]{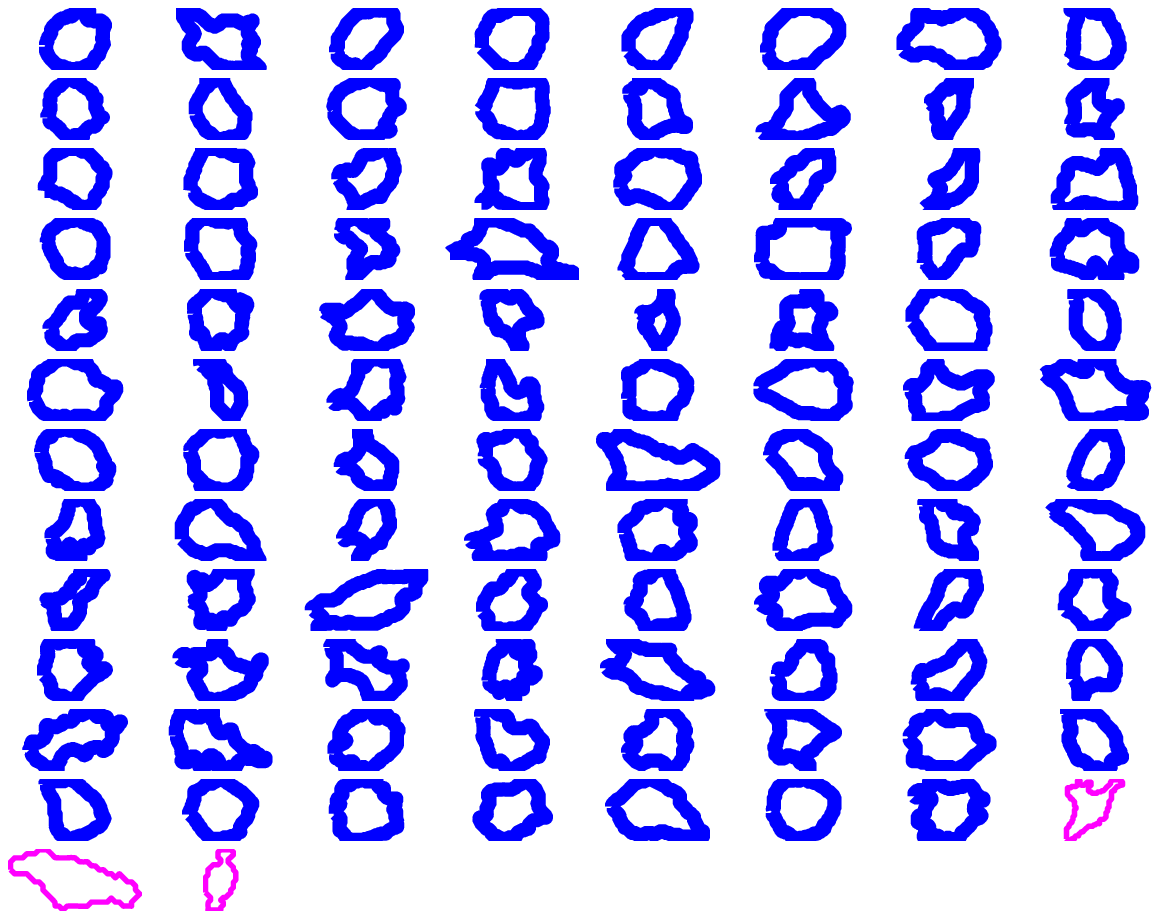}&
\includegraphics[height=1.5 in]{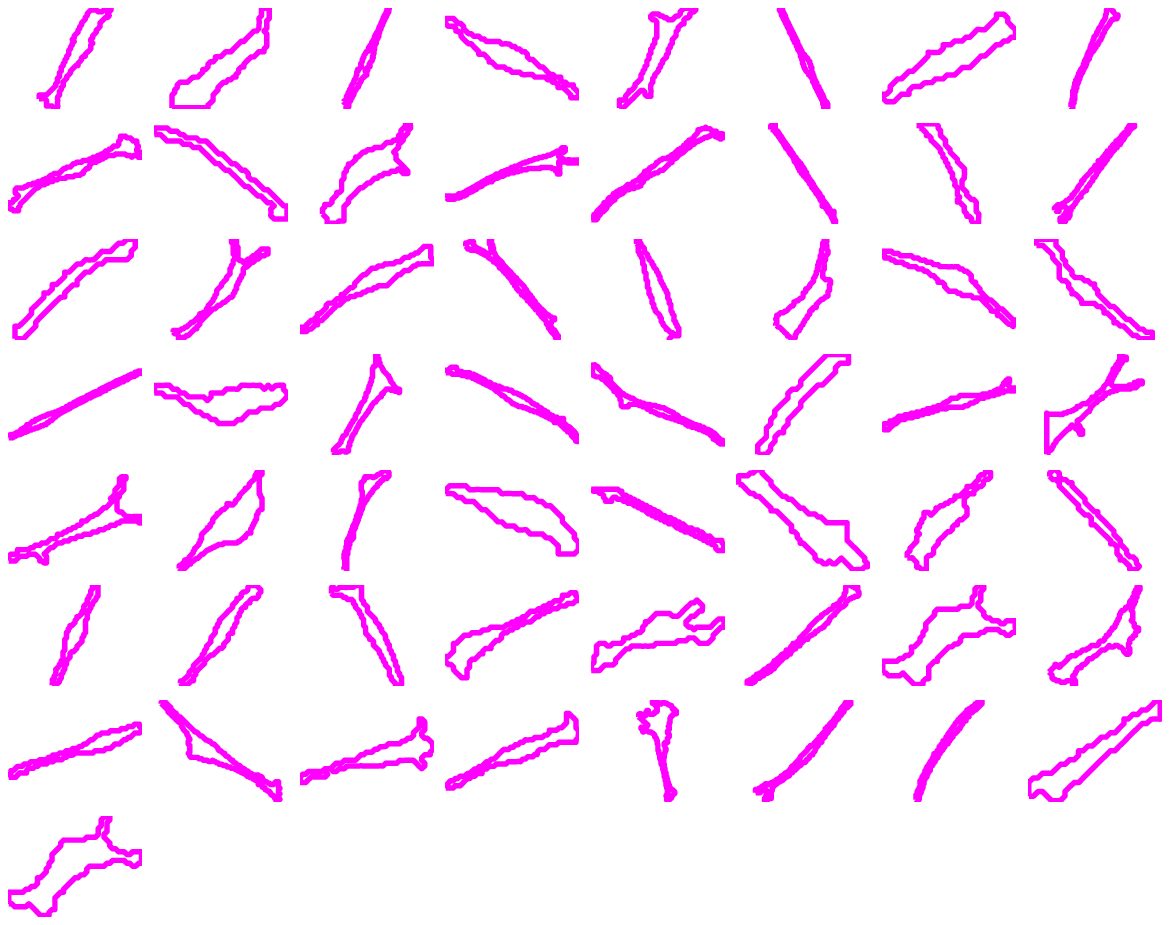}&
\includegraphics[height=1.5 in]{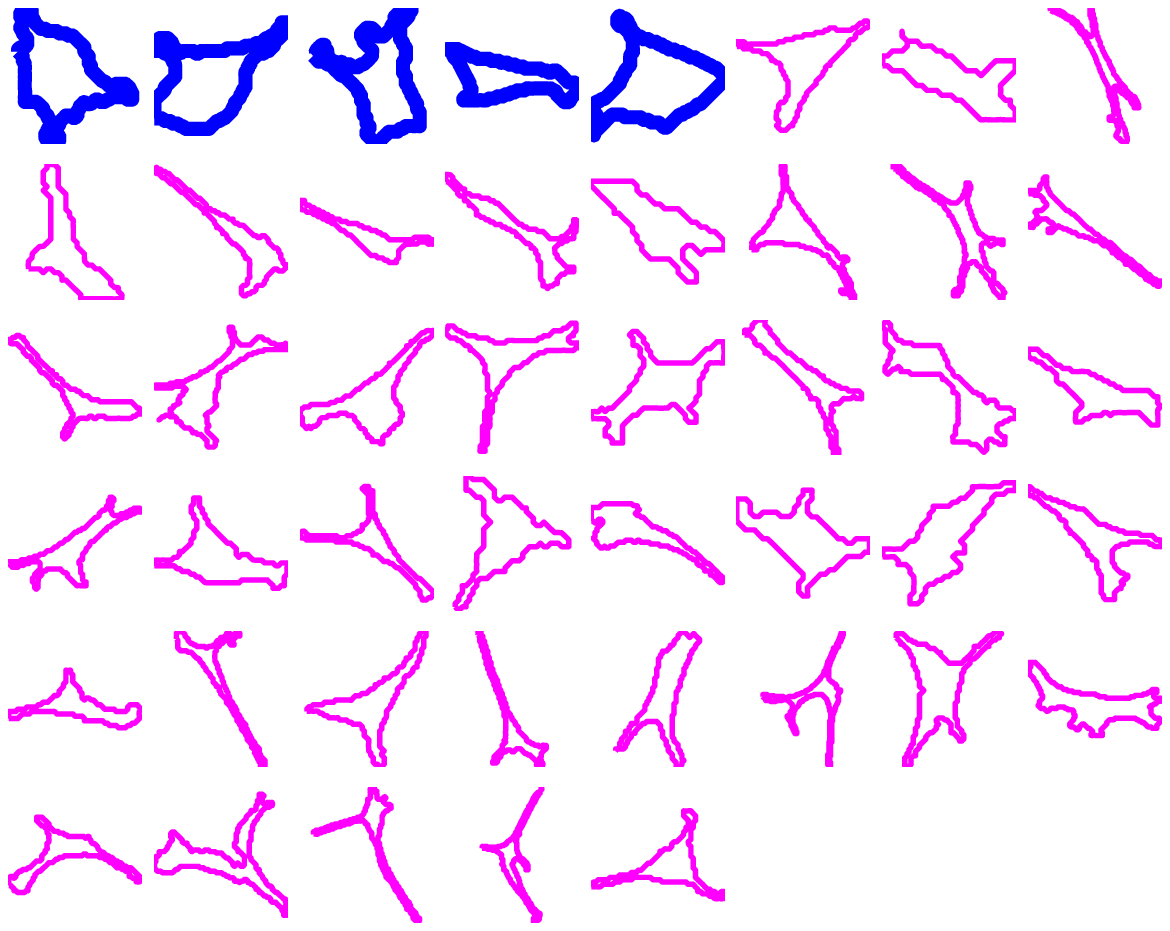}\\
\multicolumn{3}{|c|}{Our method (EIP+DW)} \\
\hline
\end{tabular}
\begin{tabular}{|cc|cc|}
\includegraphics[height=1.12 in]{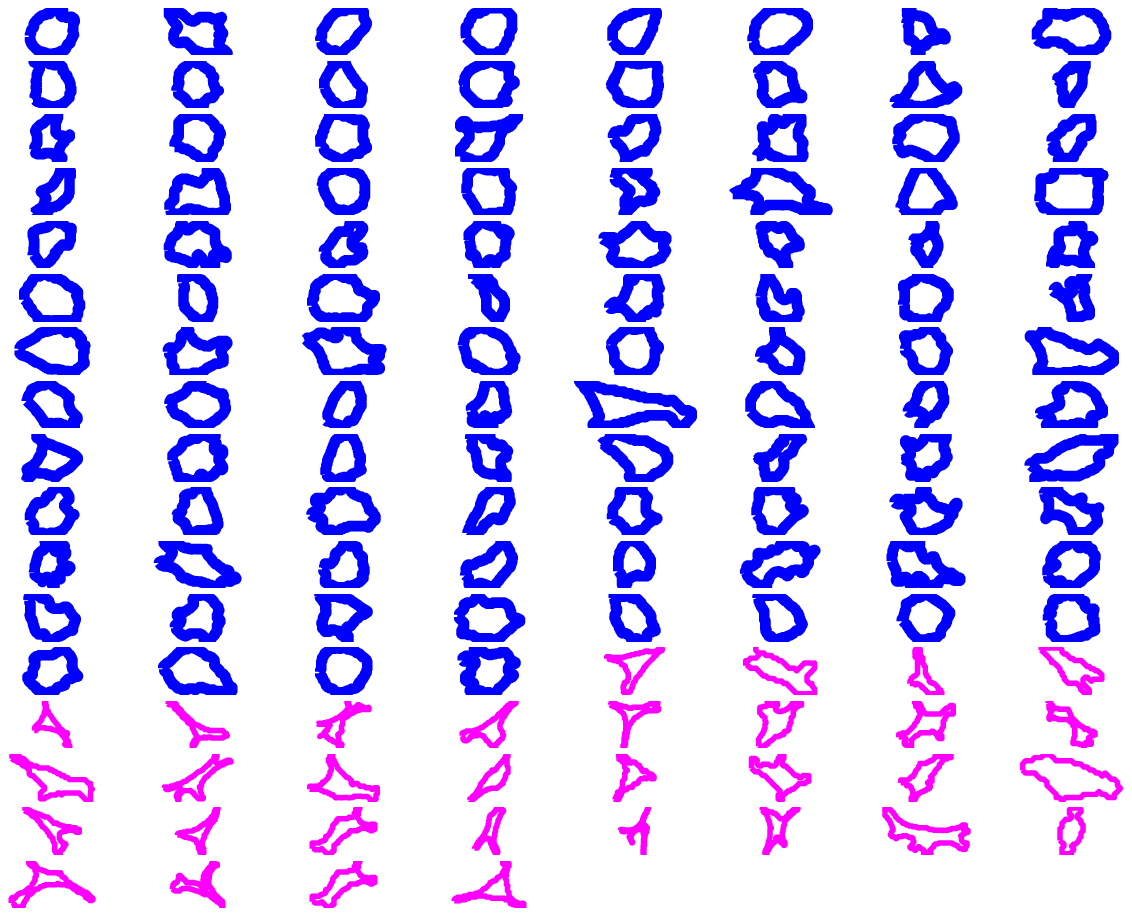}&
\includegraphics[height=1.12 in]{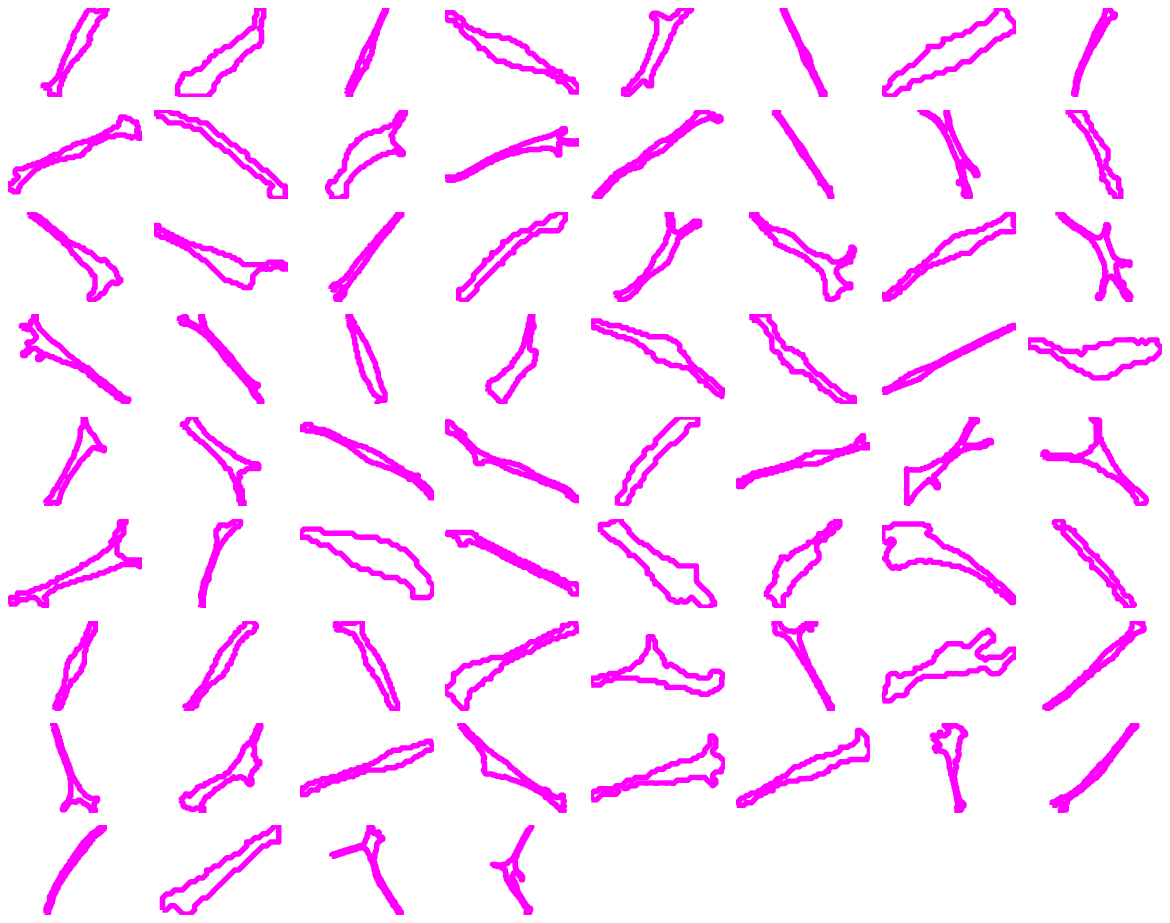}&
\includegraphics[height=1.10 in]{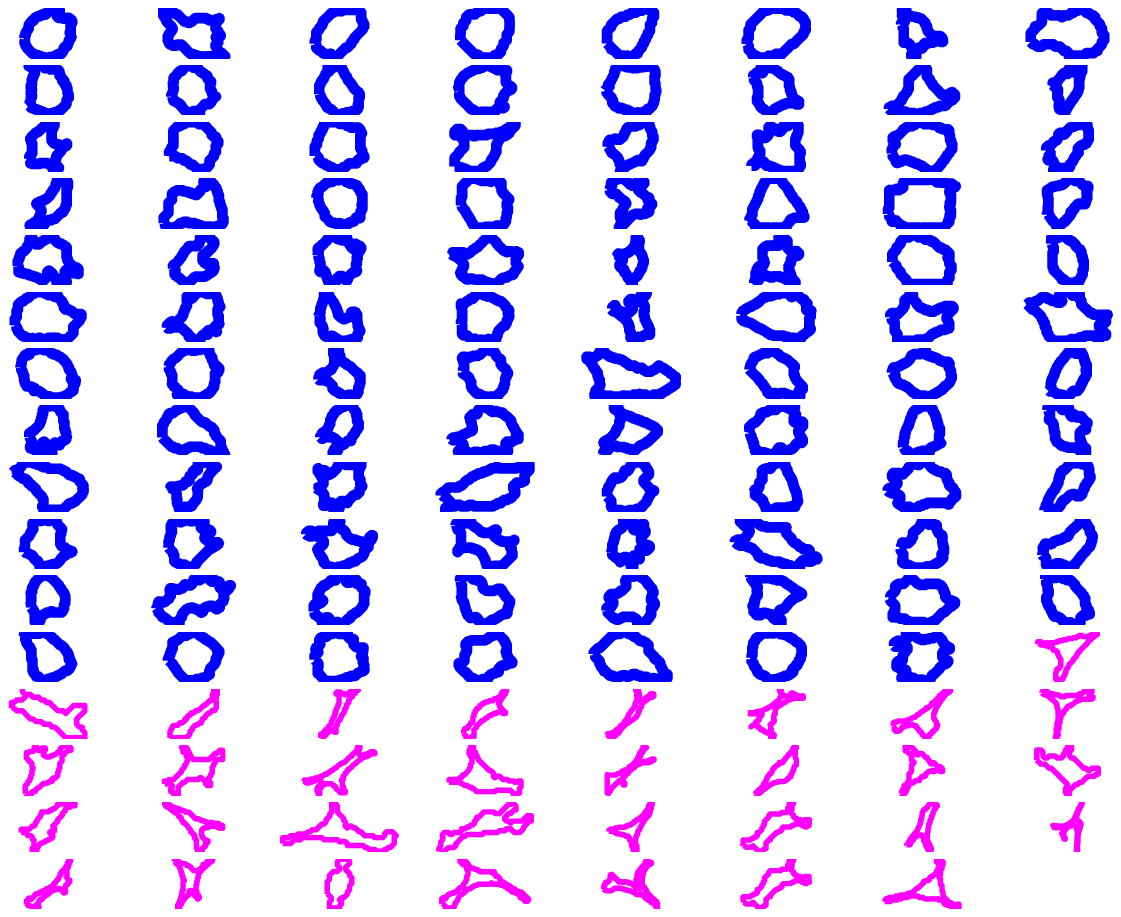}&
\includegraphics[height=1.10 in]{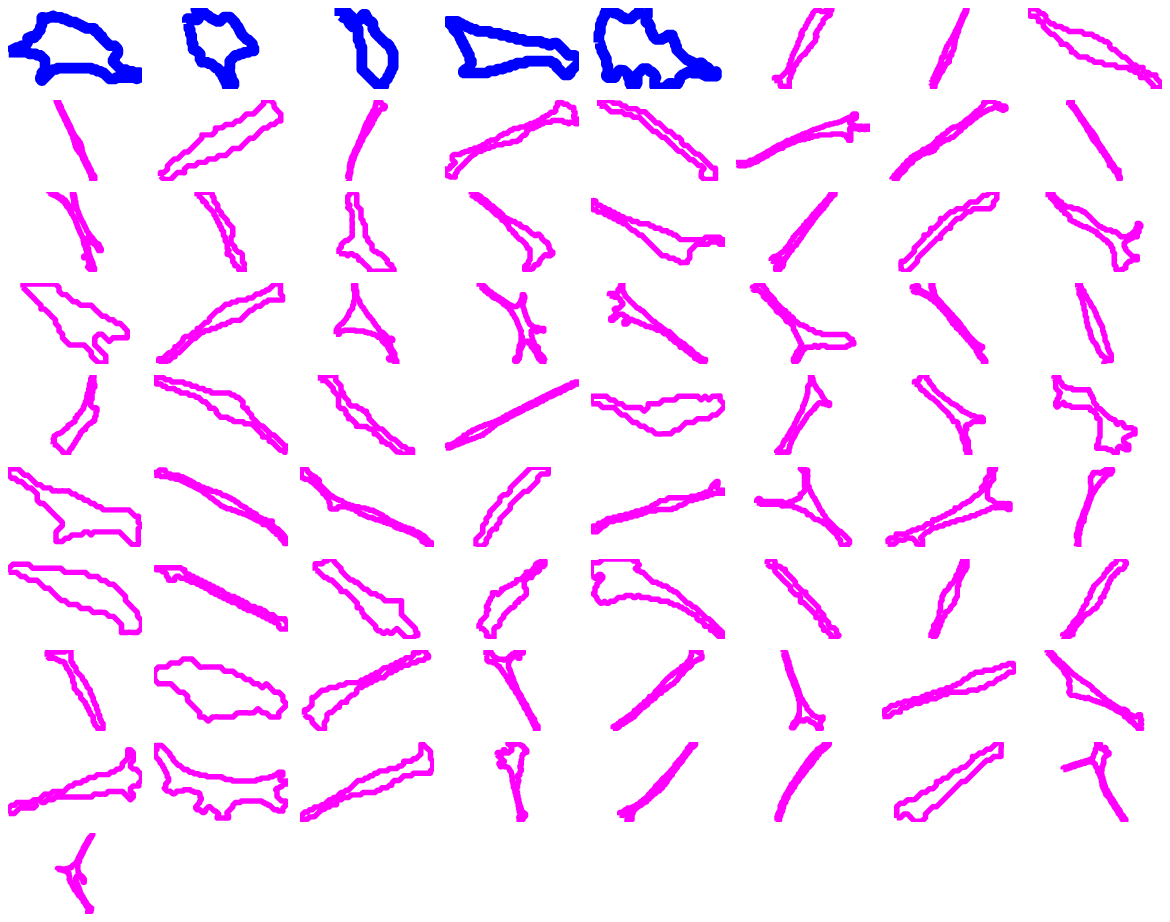}\\
 \multicolumn{2}{|c|}{ESA + PW} &  \multicolumn{2}{c|}{K-medians} \\
\hline
\end{tabular}
\caption{Clustering result for a subset of cell shapes containing both DLEX-P46 (thick blue) and NIH-3T3 (thin pink). The first row is the result of our method. The bottom left panel shows the result of ESA + PW and the bottom right shows the result of K-medians.}
\label{fig:realdlex}
\end{center}
\end{figure}

We pooled together $100$ shapes of DLEX-P46 cells and $100$ shapes of NIH-3T3 cells to form a set of $200$ cells.  Our goal is  to cluster this pooled dataset into two classes: one with the DLEX-p46 cell shapes and the other with NIH-3T3 cell shapes. 
Since we expect a smaller number of clusters here, we use a set of small $\theta$, $\theta \in \{0.1, 0.2, 0.3, 0.4\}$, $\xi = 1$ as our model parameters. The clustering result are shown in Fig. \ref{fig:realdlex} first row, where each plot is a cluster. The $200$ cell shapes are automatically clustered into $3$ classes. Cell shapes of DLEX-p46 are plotted with  thick blue lines,  and NIH-3T3 cells are plotted with thin pink lines. The NIH-3T3 cell shapes have a large variance, thus our method separates them into 2 classes, one with long strip shapes, and the other with star shapes. Setting $K=2$, we compared with ESA + PW method  in the left panel of the  bottom row and the K-medians method in the right. From the result, one can see that our method identifies $3$ meaningful clusters instead of $2$, and the clustering quality (both the classification rate and Rand index) also is higher than ESA + PW and K-median method.  The classification rates for our method, ESA+PW and K-medians are $96\%$, $83.5\%$, and $81.5\%$,  respectively, and the Rand indexes are 0.82, 0.73, and 0.69, respectively.

\noindent
{\bf Protein structure data clustering:} In the following experiments, we will use our model to cluster the protein structure data introduced in Section 2.1.

\begin{figure}
\begin{center}
\begin{tabular}{|cc|}
\hline
\includegraphics[height=1.94in]{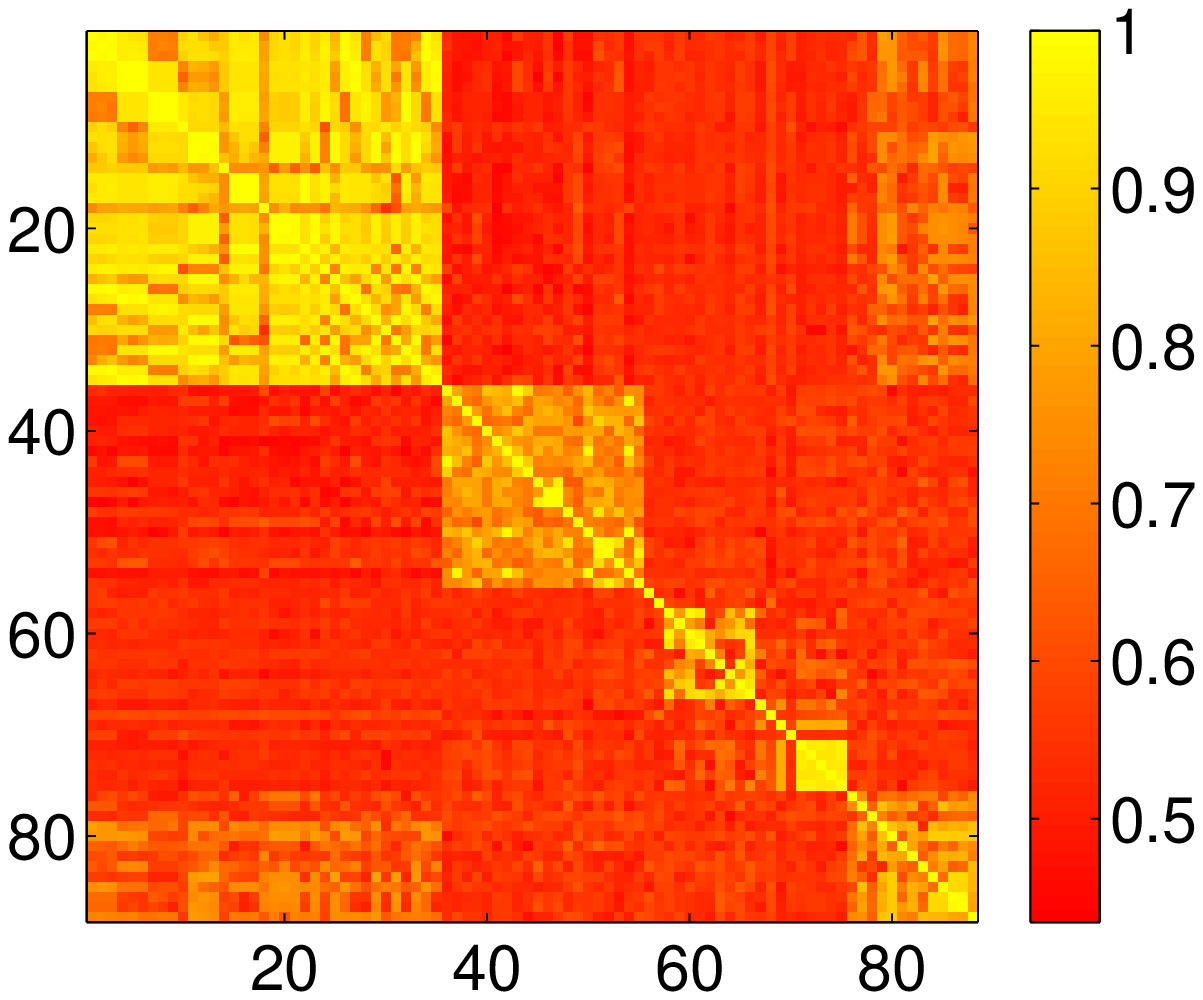}&
\includegraphics[height=1.84in]{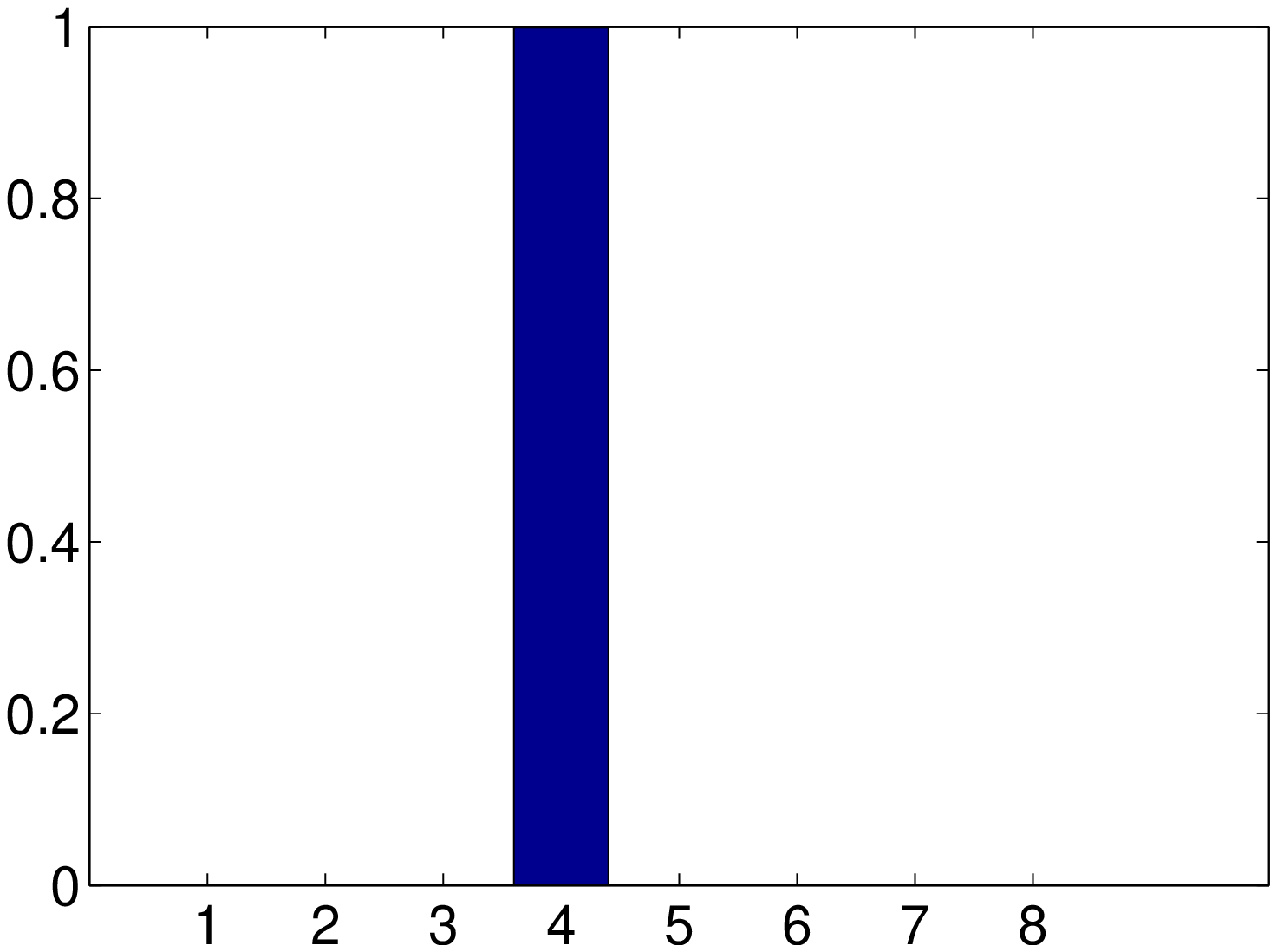}\\
I-P Matrix After & Histogram of $k$ \\
\end{tabular}
\begin{tabular}{|cccc|}
\hline
\includegraphics[height=0.9in]{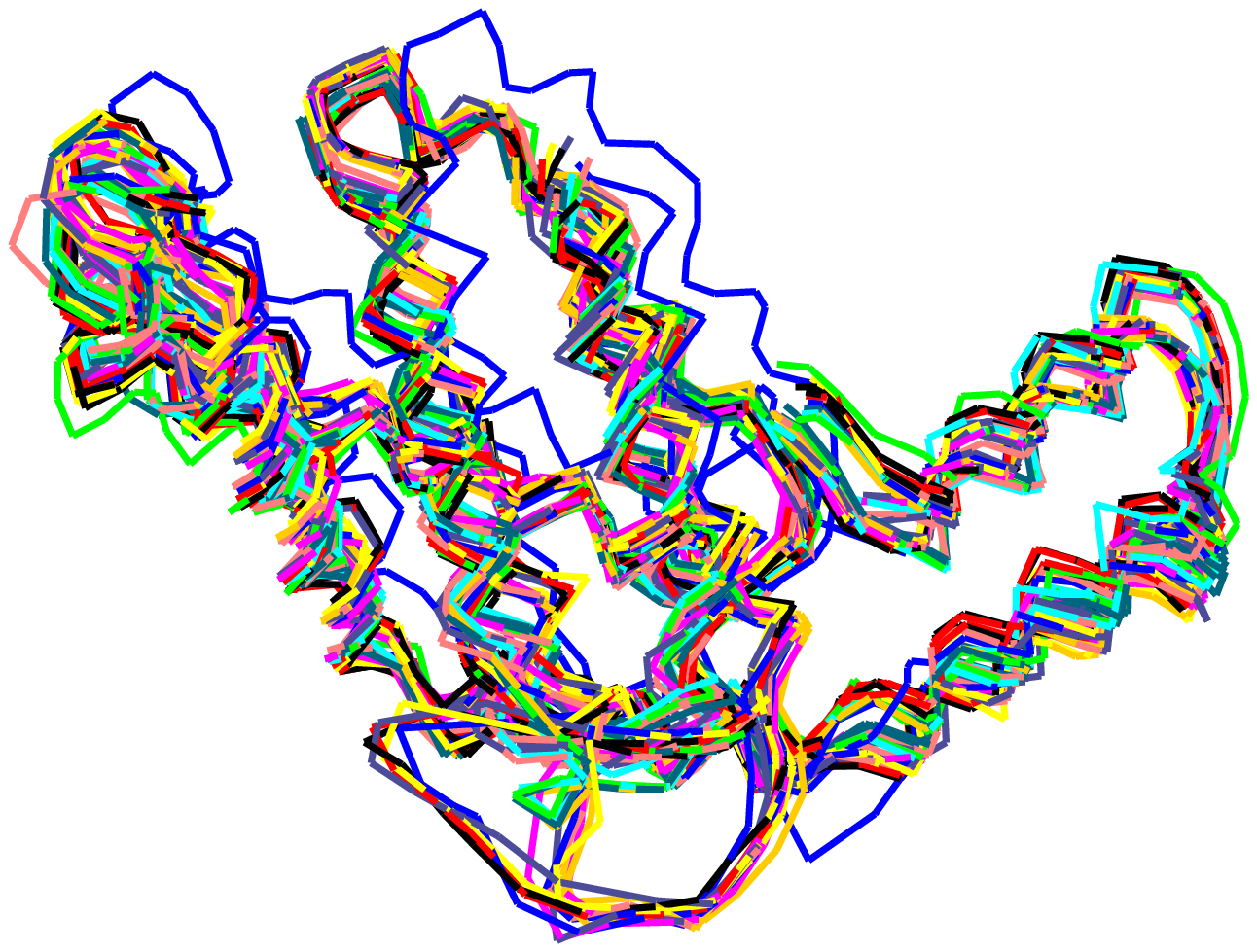}&
\includegraphics[height=0.9in]{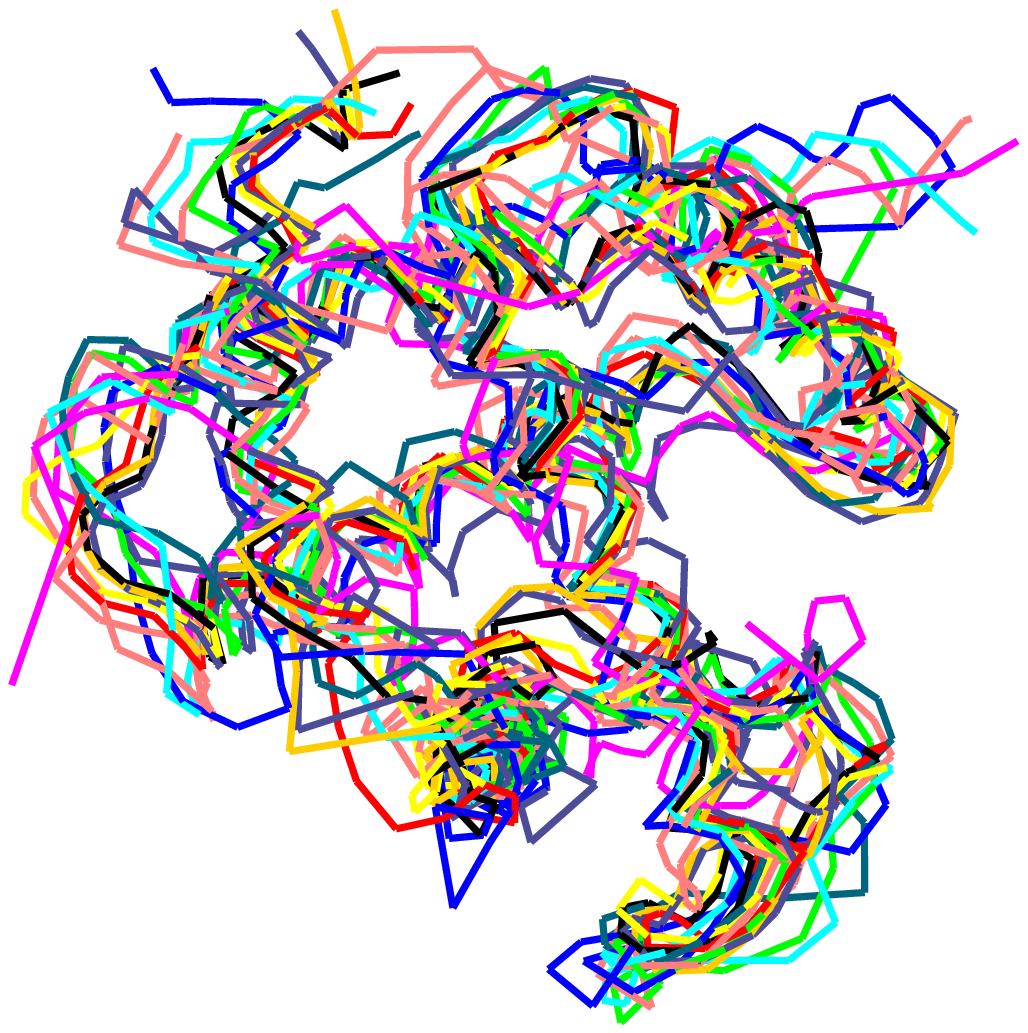}&
\includegraphics[height=0.9in]{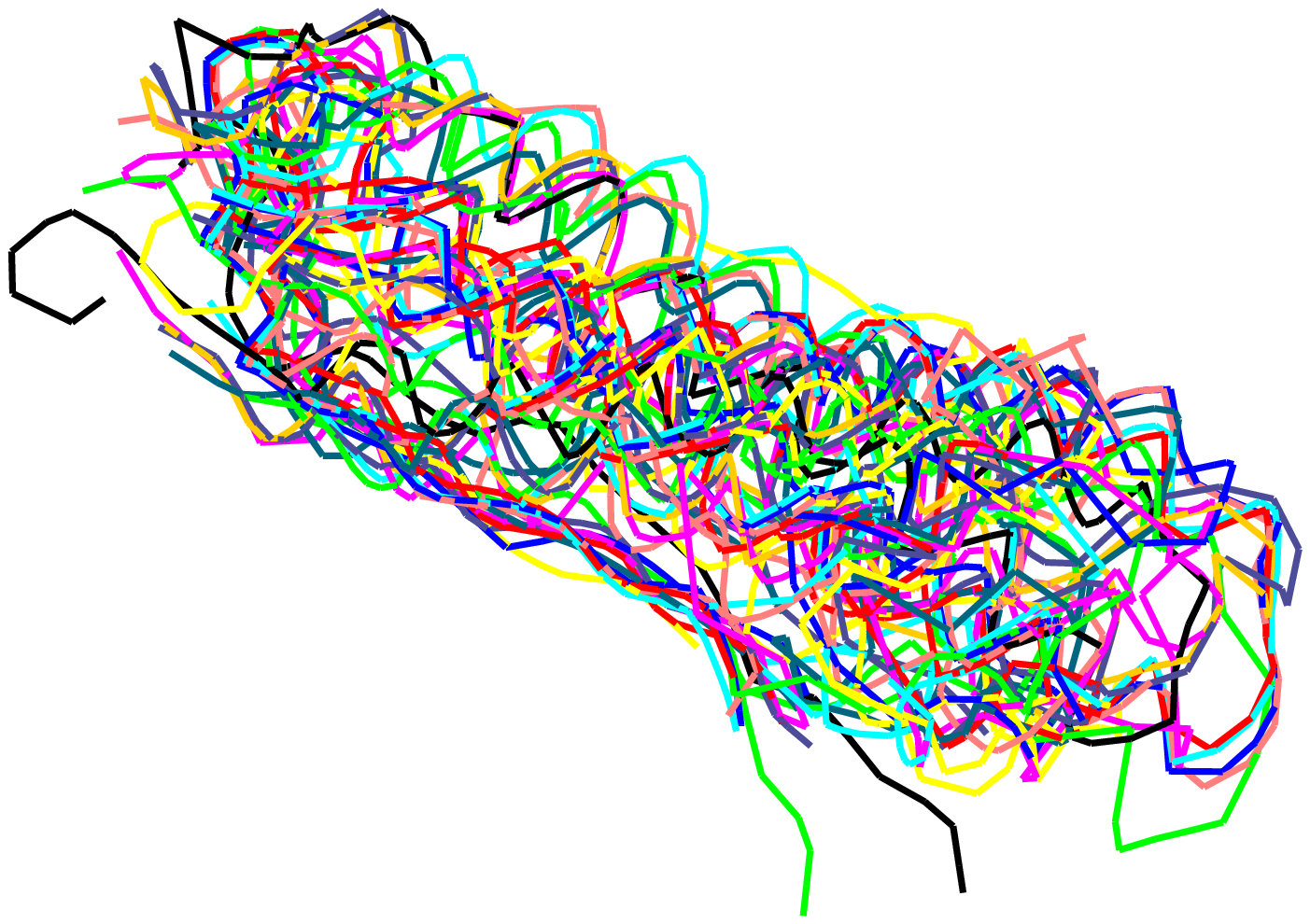}&
\includegraphics[height=0.9in]{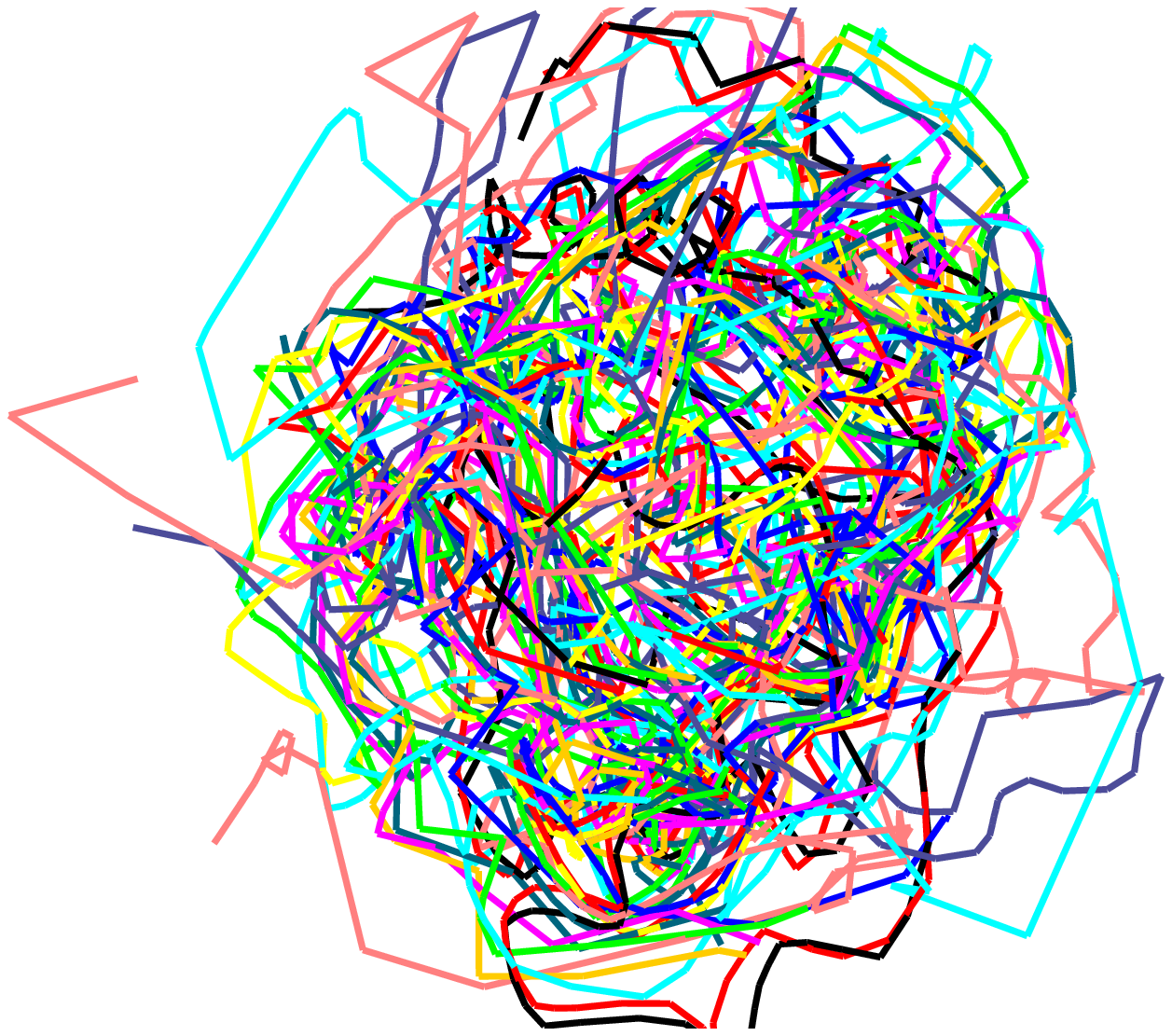}\\
Class 1 & Class 2 & Class 3 & Class 4\\
\hline
\end{tabular}
\caption{Protein structure classification on a SCOP subset containing $88$ proteins. The first row shows the inner product matrix between protein structures after clustering and the histogram of cluster number $k$. The second row shows the four clusters.}
\label{fig:proteinsmall}
\end{center}
\end{figure}

In the first experiment, we choose a small protein structure dataset obtained from SCOP with only 88 proteins. Based on SCOP, these proteins are from 4 classes (SCOP provides the ground truth). Those proteins are pre-processed similar to an earlier study \citep{Weiliu2011}.  To have a good estimate of the SRVFs from the raw data, we smooth the resampled protein structures with a Gaussian kernel. We also added one residue at both N and C terminal of each protein chain by extrapolating from the two terminal residues to allow some degrees of freedom on matching boundary residues. The added residues are removed after matching. Note that these smoothed SRVFs will only be used for searching optimal re-parameterizations $\gamma$ and rotations $SO(3)$ to get the inner product between protein structures. Then we apply our mixture of Wisharts model to the inner product matrix $S \in U_{+}(\real)$ and get the clustering result, where we use parameters $\theta \in \{0.1,0.2,0.3,0.4\}$ and $\xi = 1$. The final clustering results are shown in Fig. \ref{fig:proteinsmall}.  The clustering rate is $100\%$ compare with the ground truth provided by SCOP.  

\begin{figure}
\begin{center}
\begin{tabular}{|cc|}
\hline
\includegraphics[height=1.94in]{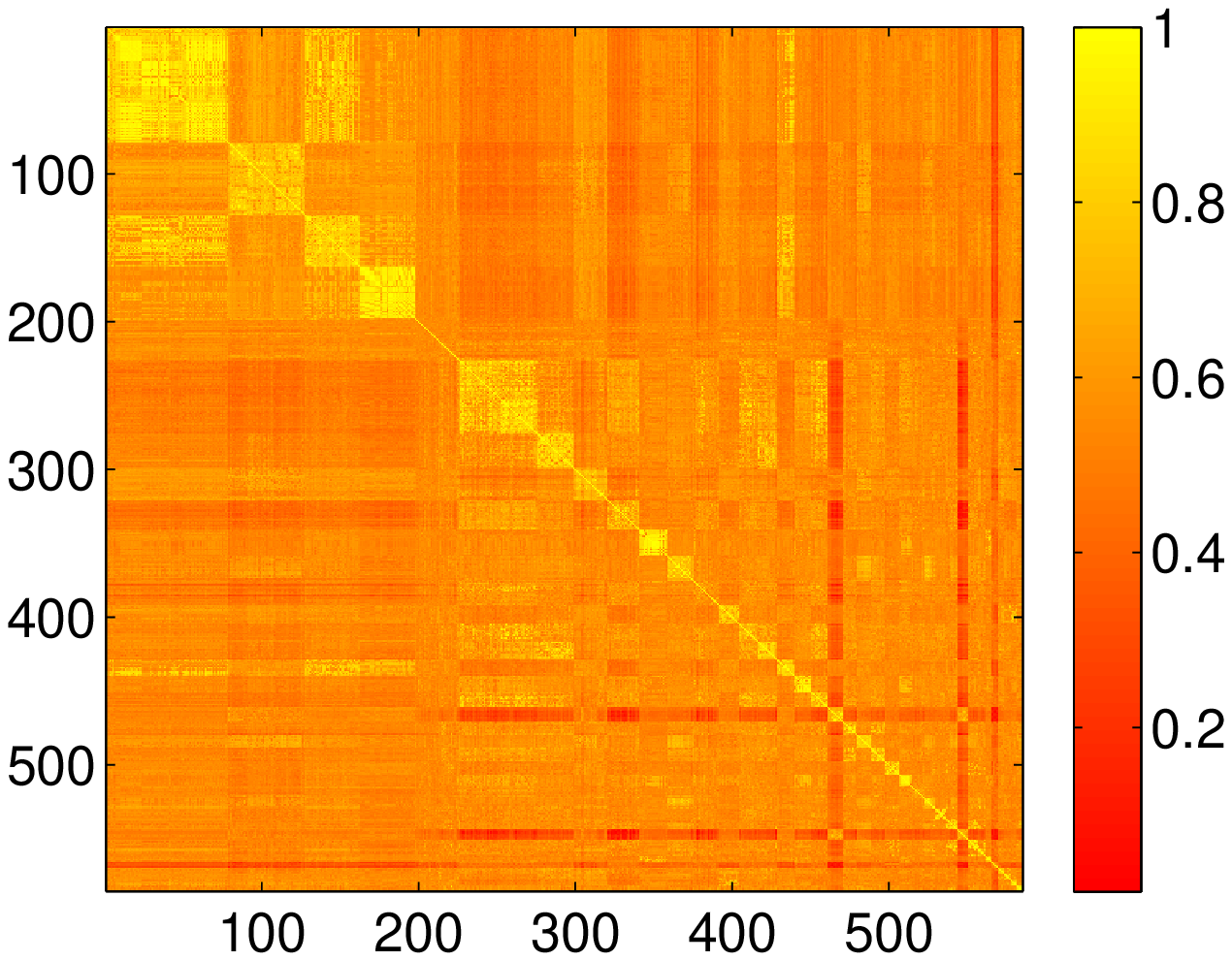}&
\includegraphics[height=1.84in]{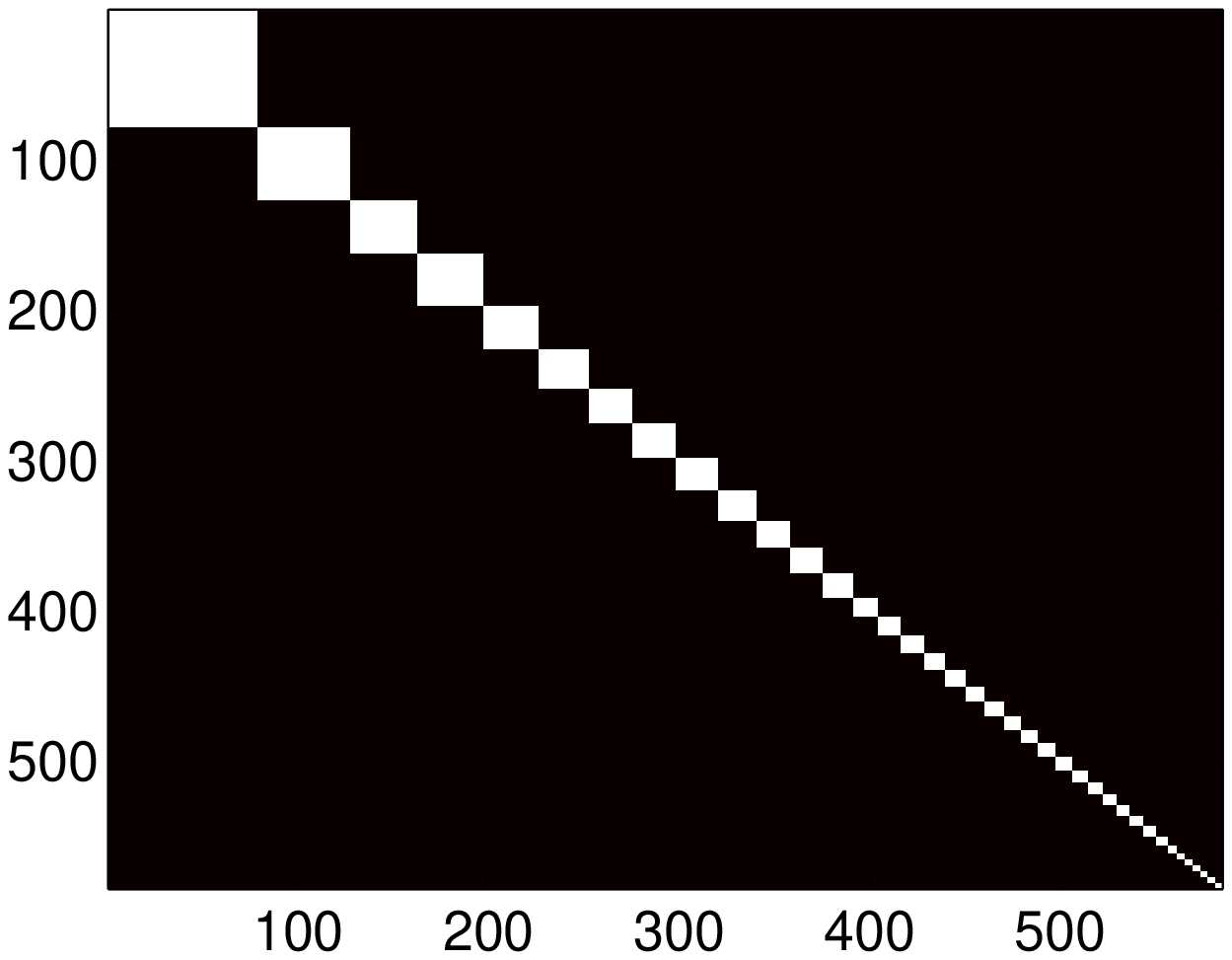}\\
I-P Matrix After & Membership Matrix $B$ \\
\end{tabular}
\begin{tabular}{|cccc|}
\hline
\includegraphics[height=0.9in]{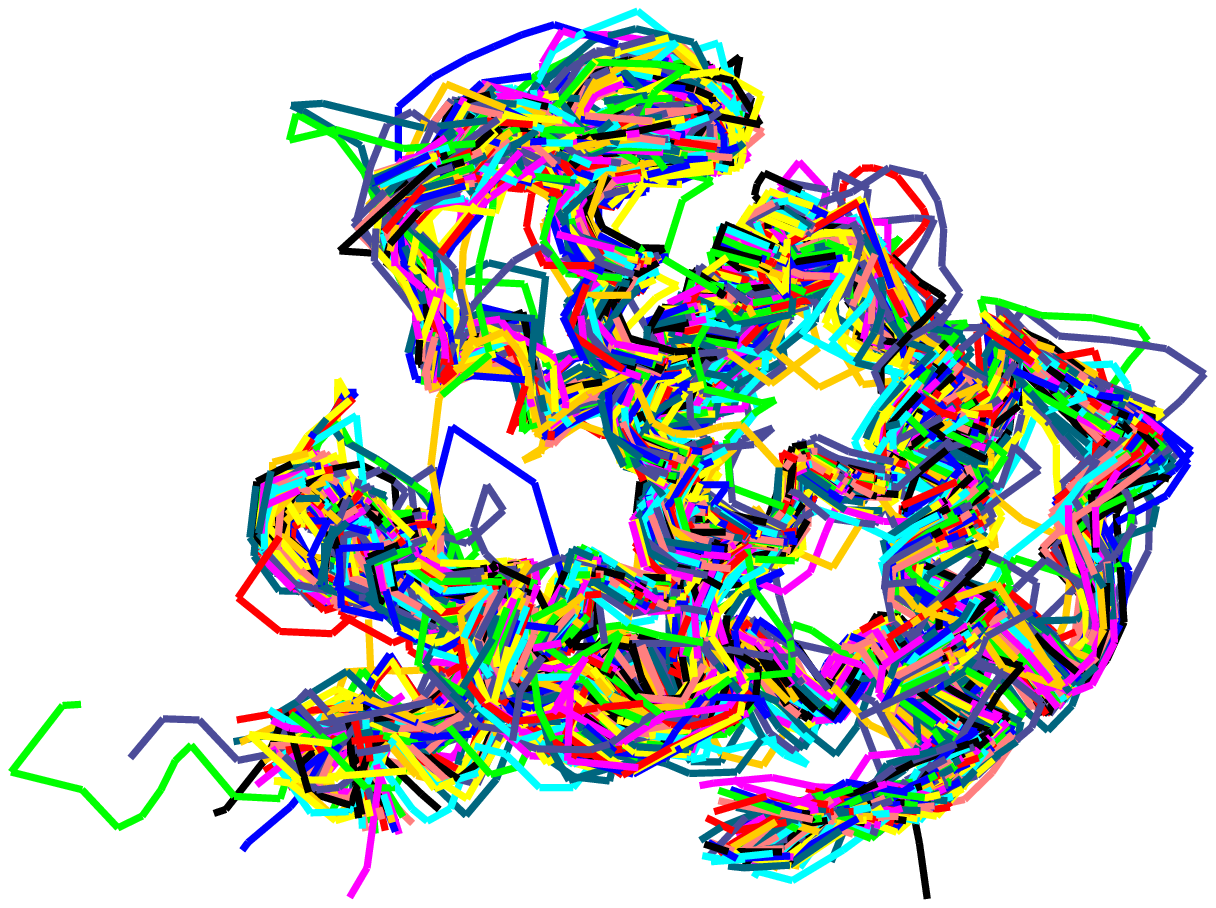}&
\includegraphics[height=0.9in]{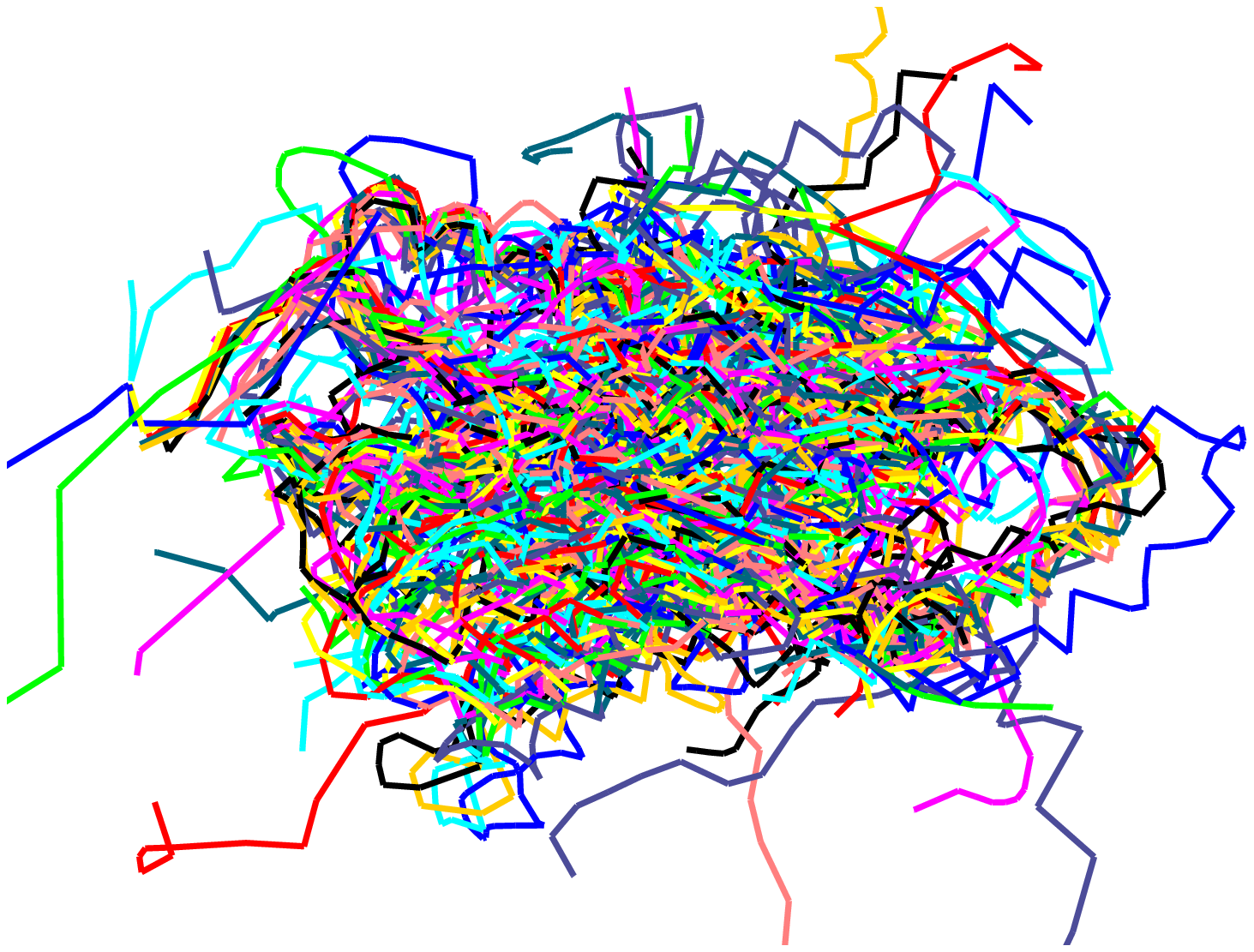}&
\includegraphics[height=0.9in]{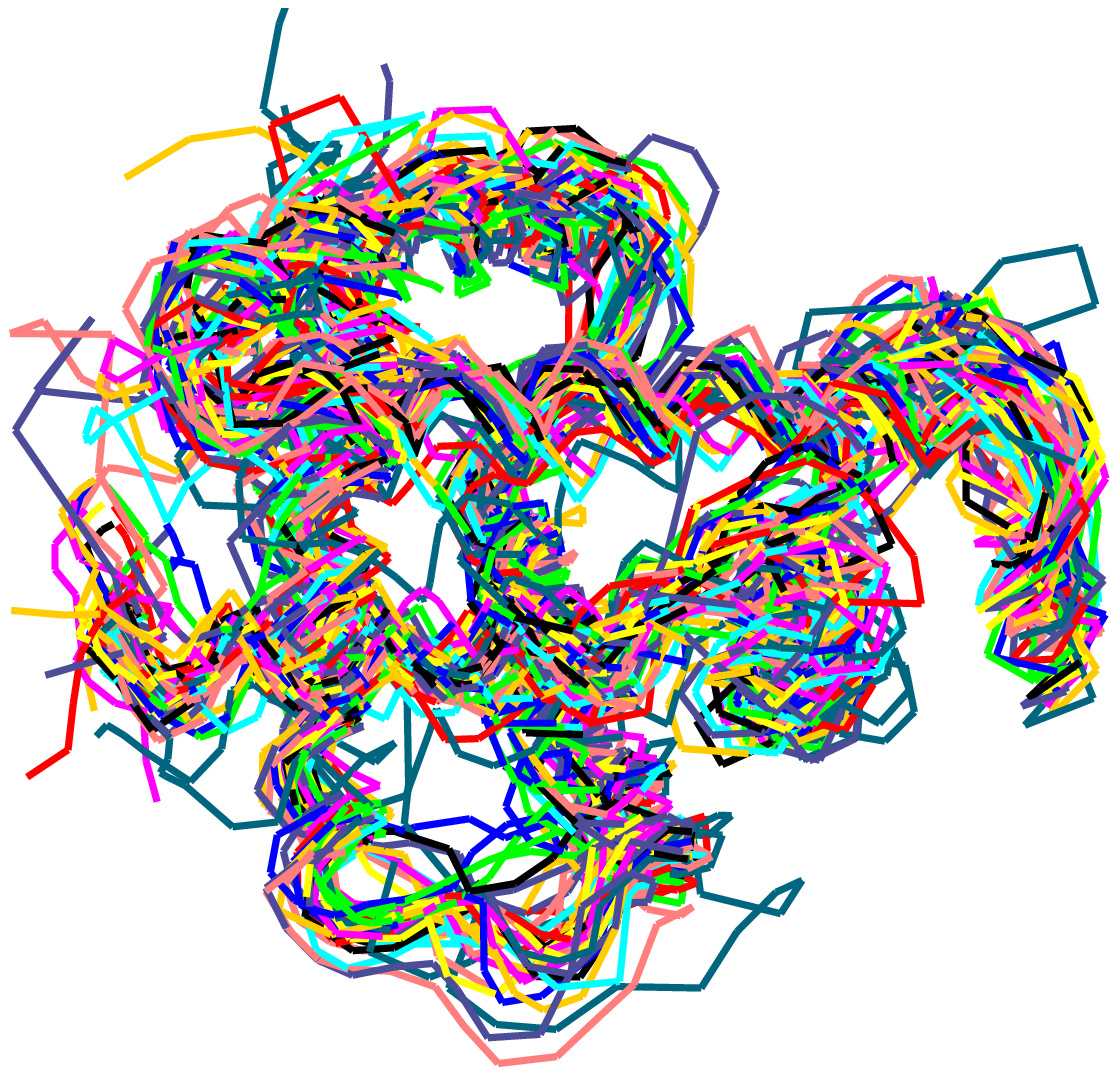}&
\includegraphics[height=0.9in]{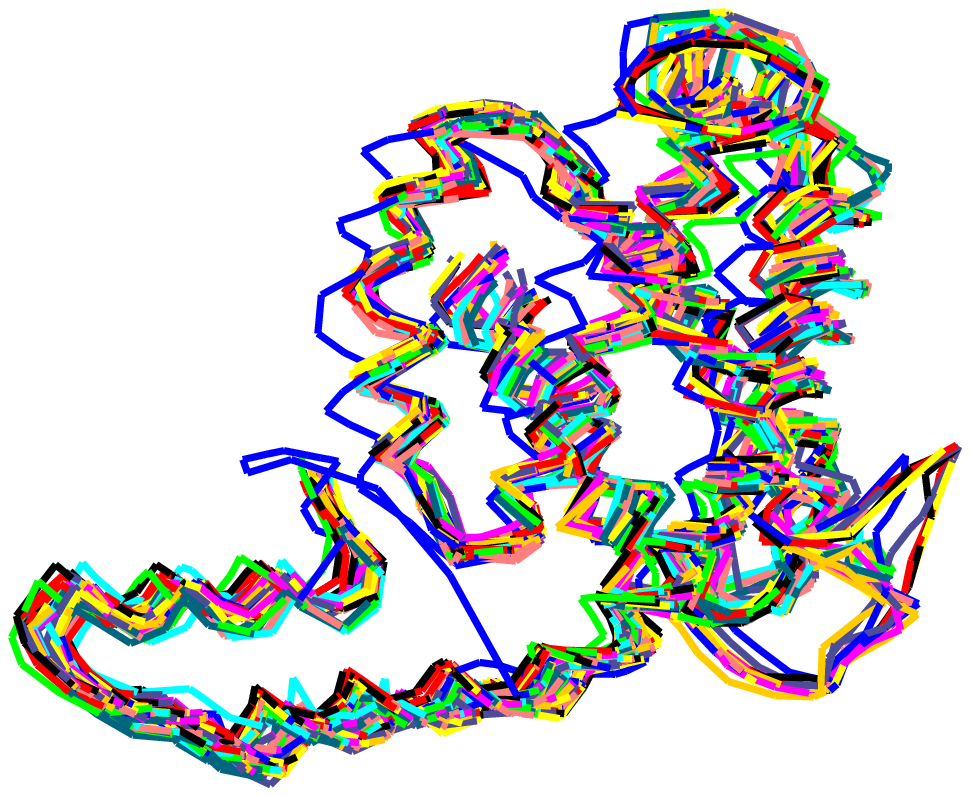}\\
Class 1 & Class 2 & Class 3 & Class 4\\
\hline
\end{tabular}
\caption{Protein structure classification on a SCOP subset of $602$ proteins. The first row shows the inner product matrix between protein structures after clustering and the corresponding inferred $B$, respectively. The second row shows the first four clusters.}
\label{fig:protein}
\end{center}
\end{figure}

In next experiment, we choose 20 classes with at least 10 elements in each class from SCOP dataset to form a subset with $602$ proteins.  
The final clustering result shows some clusters with only a few elements which we consider as {\it outliers}. In this experiment, our model identifies 17 outliers (7 small clusters).  After removing these outliers, the remaining $585$ proteins are clustered into $38$ classes. The clustering rate is $84.1\%$. The first row in Fig. \ref{fig:protein} shows the inner product matrix corresponding to the $585$ protein structures (after putting elements in the same cluster together), and the posterior estimate of the partition matrix $B$.  The second row shows first four clusters of the clustering result after the alignment (removing shape-preserving transformations). One can see that inside each cluster, the shapes of these protein structures are very similar to each other. As comparisons, we remove the outliers detected by our method, then apply ESA + PW and K-medians method to cluster the left $585$ proteins by setting $K=20$. ESA + PW gets $75.99\%$ of classification rate and K-medians gets $63.42\%$. The Rand indexes for our model, ESA + PW and K-medians are 0.95,0.93, and 0.91 respectively.  As evident, we obtain a good clustering result based on only the shape of the proteins. 

\section{Conclusion}\label{sec:disc}
We have presented a Bayesian approach for clustering of shape data that does not require the 
number of clusters a priori. Instead, it assumes a flexible prior on the space of data partitions and 
studies the resulting posterior distribution on the clustering configuration. This prior is derived
from a Dirichlet process (realized using the Chinese restaurant process) and the likelihood is 
given by the Wishart distribution. The Bayesian inference provides a reasonable solution for 
each of the simulated and real shape datasets studied in this paper. This fully automated method 
provides a way for shape-based partitioning of large datasets even in situations where visualization-based 
tools are not feasible.

\bibliographystyle{elsarticle-harv}
\bibliography{paper}

\end{document}